\documentclass[10pt,journal,compsoc]{IEEEtran}
\usepackage[pdftex]{graphicx}
\usepackage[table,xcdraw]{xcolor}
\usepackage{placeins}
\usepackage{amsfonts}
\usepackage{amssymb}
\usepackage{amsmath}
\usepackage{rotating}
\usepackage{multirow}

\begin{document}
\newcommand{\superset}[1]{\mathcal{{#1}}}
\newcommand{\set}[1]{\textit{#1}}
\newcommand{\function}[1]{\textbf{\textit{#1}}}
\newcommand{\concept}[1]{\textit{#1}}

%
\title{ITSELF: Iterative Saliency Estimation fLexible Framework}
%
%
%

\author{Leonardo~de~Melo~Joao, Felipe de Castro Belem, and Alexandre~Xavier~Falcao, ~\IEEEmembership{Member,~IEEE,}}


\IEEEtitleabstractindextext{%
\begin{abstract}
    Saliency object detection estimates the objects that most stand out in an image. The available unsupervised saliency estimators rely on a pre-determined set of assumptions of how humans perceive saliency to create discriminating features. By fixing the pre-selected assumptions as an integral part of their models, these methods cannot be easily extended for specific settings and different image domains. We then propose a superpixel-based ITerative Saliency Estimation fLexible Framework (ITSELF) that allows any user-defined assumptions to be added to the model when required. Thanks to recent advancements in superpixel segmentation algorithms, saliency-maps can be used to improve superpixel delineation. By combining a saliency-based superpixel algorithm to a superpixel-based saliency estimator, we propose a novel saliency/superpixel self-improving loop to iteratively enhance saliency maps. We compare ITSELF to two state-of-the-art saliency estimators on five metrics and six datasets, four of which are composed of natural-images, and two of biomedical-images. Experiments show that our approach is more robust than the compared methods, presenting competitive results on natural-image datasets and outperforming them on biomedical-image datasets.
\end{abstract}
}
\maketitle
\IEEEdisplaynontitleabstractindextext
\IEEEpeerreviewmaketitle

%

\IEEEraisesectionheading{\section{Introduction}\label{sec:intro}}
\IEEEPARstart{D}{etermining} visual saliency of image objects is a broadly studied subject, highly applicable to a vast number of tasks, such as image quality assessment \cite{liu2011visual}, content-based image retrieval \cite{chen2009sketch2photo}, and image compression \cite{guo2009novel}. Many algorithms have been proposed to estimate visual saliency, and they can be categorized as supervised and unsupervised approaches.

Supervised saliency estimators use pixel-wise ground-truth images to learn discriminant features of salient objects. The most accurate supervised algorithms are based on deep-learning techniques \cite{wang2019salient}, but they require large amounts of training data annotated by humans, and the generalization of the trained models across image datasets or image domains usually requires retraining and adaptations. Unsupervised saliency estimators, however, model saliency based on prior knowledge about the salient objects and local image characteristics, usually compromising accuracy in exchange for removing the requirement for intensive data annotation, but being more flexible across image domains. In this work, we are focused on unsupervised saliency estimators.

Most unsupervised methods for saliency estimation adopt a combination of bottom-up image-extracted information and top-down domain-specific assumptions. The bottom-up information is often extracted from image regions that, given modeled assumptions, have a high likelihood of being either foreground (salient) or background. These regions, namely \concept{queries}, are compared to the rest of the image, and a similarity score defines how salient the other regions are. Top-down assumptions, on the other hand, use \concept{prior} knowledge of how humans perceive saliency -- \textit{e.g.,} increase the saliency of centered, focused, and vivid-colored objects.

The available methods use a combination of pre-selected priors and query-comparison strategies to model the saliency perception of an average viewer. This pre-selection of assumptions allows for off-the-shelf methods that are easy to use and perform well in many scenarios. On the other hand, they are not extensible to applications that drift off of their pre-determined guesses.

For example, if we shift the image domain from natural images to medical images, the desired saliency is not modeled after the average viewer; Rather, it is modeled after a specialist's perception. For instance, say a physician is analyzing an x-ray image of the thorax: object centering and vivid colors cease to be salient object characteristics, causing the off-the-shelf methods to perform poorly.

To the best of our knowledge, there is no unsupervised saliency estimation algorithm that allows the user to select or incorporate a problem-specific set of assumptions. To fulfill this gap, we propose the \concept{ITerative Saliency Estimation fLexible Framework (ITSELF)}. 

\begin{figure}[t!]
        \centering
        \begin{tabular}{c}
            \includegraphics[width=0.4\textwidth]{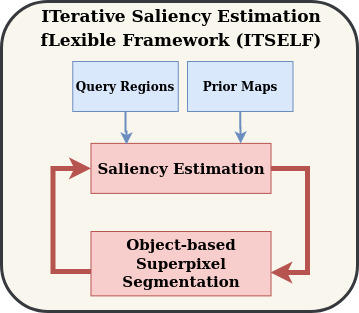}
        \end{tabular}
        \caption{ITSELF's components and structure. Note the saliency-superpixel loop depicted in red. }
        \label{fig:framework}
    \end{figure}

\begin{figure}[t!]
        \centering
        \begin{tabular}{c c}
                \includegraphics[width=0.18\textwidth]{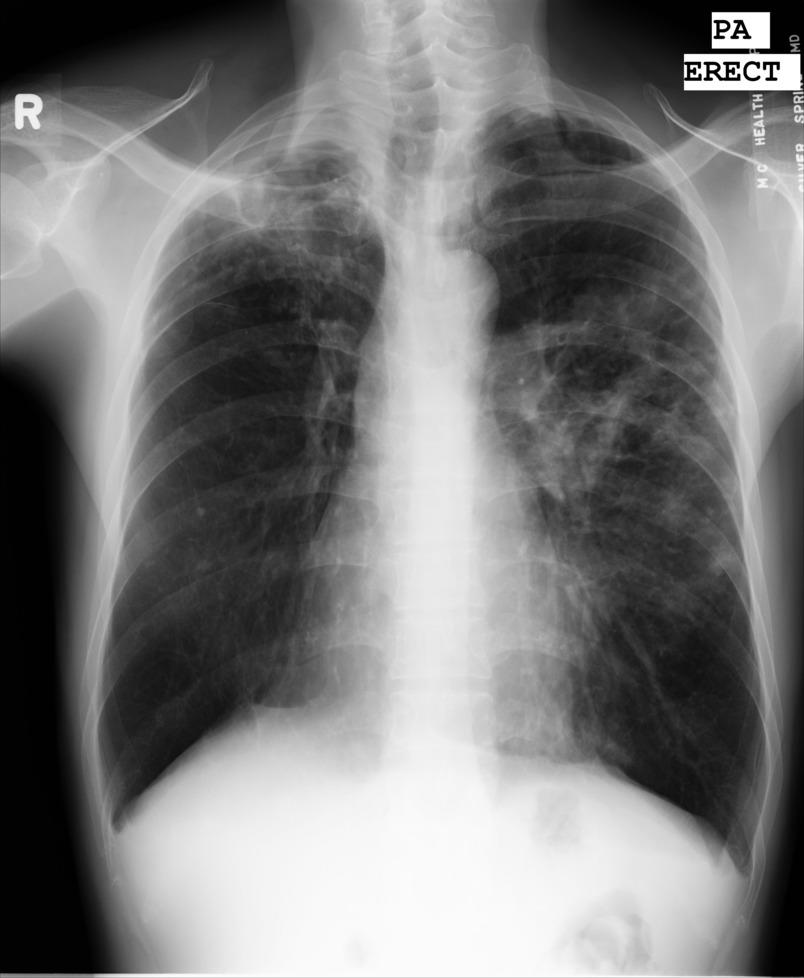} &
                \includegraphics[width=0.18\textwidth]{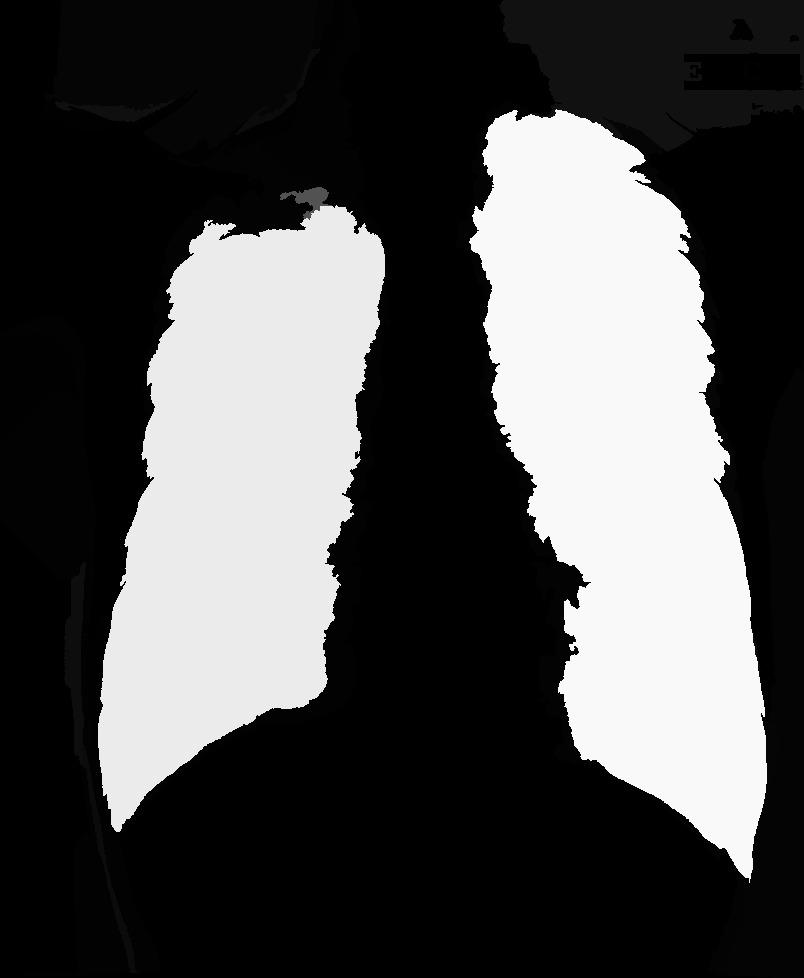} \\
                (a) & (b)
                \\
                \includegraphics[width=0.18\textwidth]{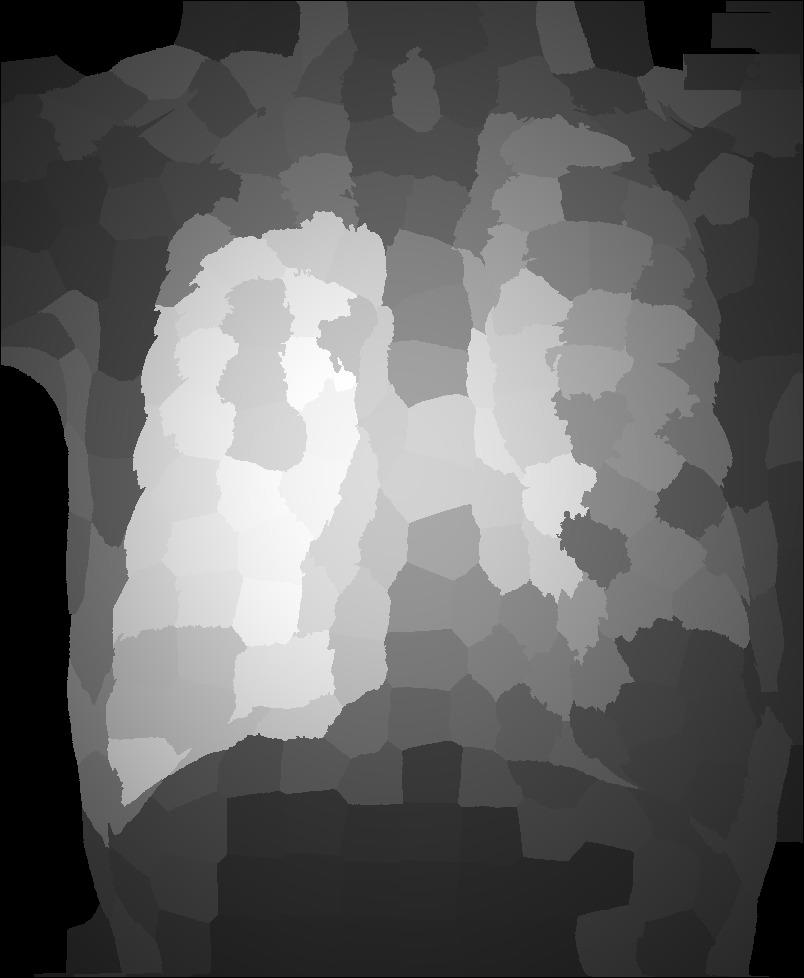} &
                \includegraphics[width=0.18\textwidth]{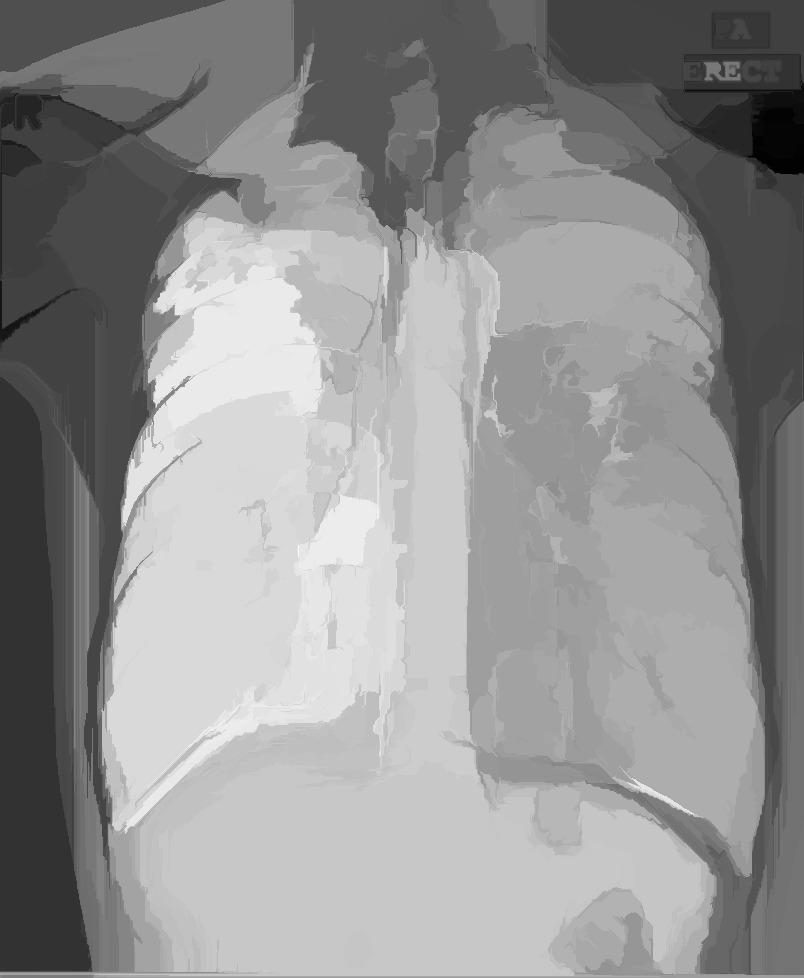} \\
                (c) & (d)
            \end{tabular}
        \caption{(a) Original image. (b-d) ITSELF, SMD and DRFI saliency maps.}
        \label{fig:inflexibility-limitations}
\end{figure}

ITSELF is a graph-based framework that allows user-defined \concept{priors} and \concept{query-region selection}, making it flexible to multiple image domains (Figures \ref{fig:framework} and \ref{fig:inflexibility-limitations}). Saliency is estimated by computing similarities on a superpixel graph, where the nodes are superpixels, and the arcs connect superpixels according to some adjacency relation based on the query regions. The saliency score is improved using multiple top-down prior information combined into a single prior map.

\begin{figure}[t!]
        \centering
        \begin{tabular}{c c c}
                 \includegraphics[width=0.13\textwidth]{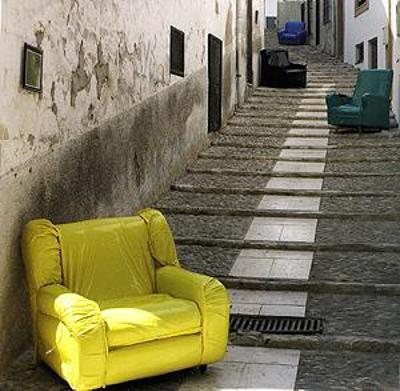} &
                 \includegraphics[width=0.13\textwidth]{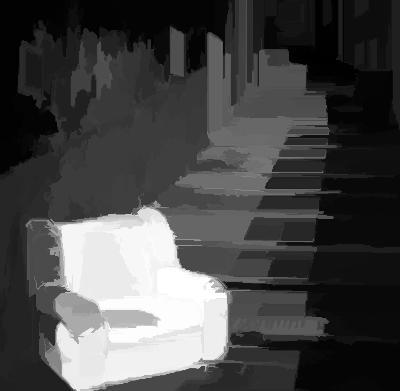} &
                 \includegraphics[width=0.13\textwidth]{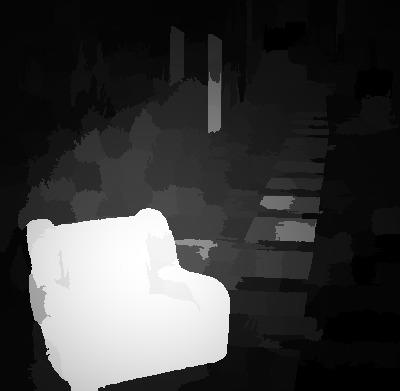} \\
                (a) & (b) & (c)
                \\
                 \includegraphics[width=0.13\textwidth]{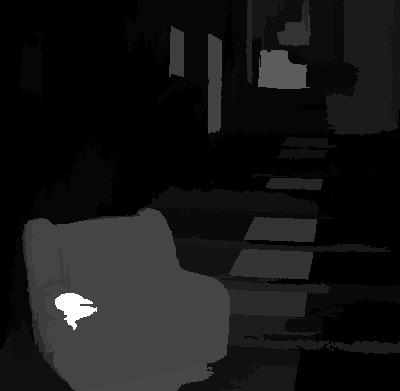} &
                 \includegraphics[width=0.13\textwidth]{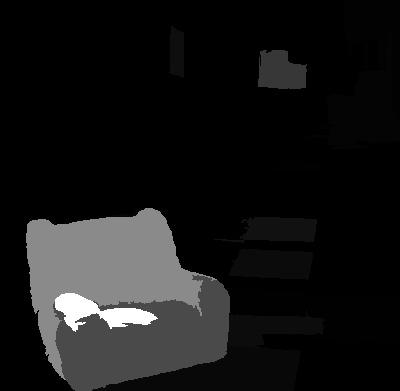} &
                 \includegraphics[width=0.13\textwidth]{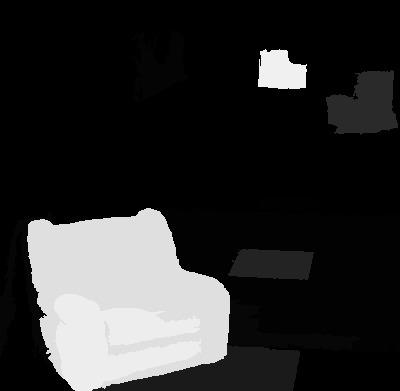} \\
                 & (d) &
                \\
                 \includegraphics[width=0.13\textwidth]{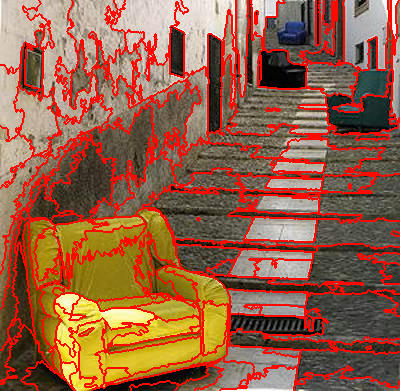} &
                 \includegraphics[width=0.13\textwidth]{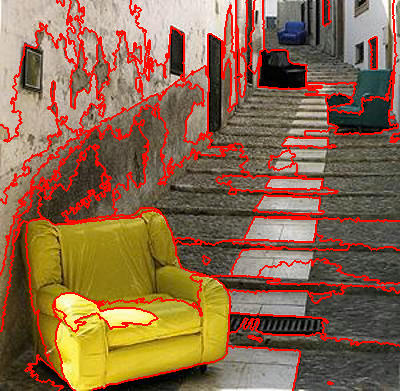} &
                 \includegraphics[width=0.13\textwidth]{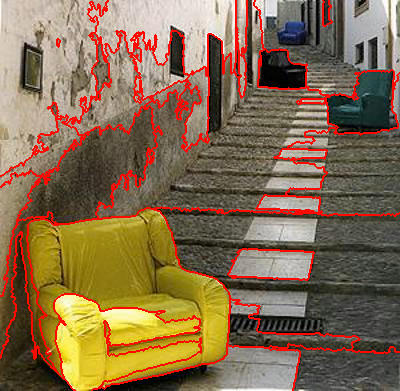} \\
                 & (e) &
            \end{tabular}
        \caption{(a) Original image. (b) DRFI saliency map; (c) SMD saliency map. (d-e) ITSELF saliency maps and superpixel segmentation at iterations 1, 5 and 8. Note how the iterations reduced the initial errors and how ITSELF also highlighted the other chairs.}
        \label{fig:itself-superpixel-semantic}
\end{figure}

Additionally, ITSELF proposes a novel approach to enhance saliency maps iteratively by using object-based superpixel delineation \cite{belem2019oisf}. This approach uses saliency information to better represent the salient objects as the union of a few superpixels. By revisiting the saliency estimation with the improved superpixels, we can also create better saliency maps and use them to re-segment the image into superpixels (Figure \ref{fig:framework}). This virtuous cycle improves saliency over time and creates more intuitive saliency maps (Figure \ref{fig:itself-superpixel-semantic}). When compared to the existing unsupervised approaches, ITSELF can leverage the object saliency map under construction to improve the process and output a superpixel segmentation as a byproduct. In this work, our focus is on the saliency estimator only.

 We propose new prior modeling for specific scenarios, including a shape-based, a saliency-based, and a user-drawn scribble-based prior. Also, we present query selection strategies based on image boundary, and previously computed saliency maps. It is worth noting that ITSELF allows for any number of prior maps and query selection strategies. Therefore, the results presented in this paper are just possible implementations of the framework.
 
We compare ITSELF to two state-of-the-art methods --- namely DRFI \cite{jiang2013drfi} and SMD \cite{peng2016salient} --- using four well-established natural-image datasets, an in-house biomedical image dataset of parasite-eggs, and an x-ray dataset of lung gray-scale images. Even though the selected datasets provide three different image domains, ITSELF provided appropriate saliency estimations for all of them. We achieved comparable results to the state-of-the-art algorithms on natural images and considerably outperformed them on non-natural images.

Thus, the contributions of this work are: (1) a saliency estimation framework that easily allows the incorporation of domain-specific information; (2) the improvement of saliency estimation by using object-information during superpixel segmentation; (3) a novel method for iteratively enhancing both saliency maps and superpixel segmentation.

The core elements of ITSELF are presented in Section \ref{sec:framework}. Sections \ref{sec:query-estimation} and \ref{sec:prior-estimation} present strategies to implement queries and priors. The results of some ITSELF create saliency estimator are presented on Section \ref{sec:experiments}, together with their setup and the databases used for comparison. Lastly, conclusions are drawn on Section \ref{sec:conclusion}.
\section{Related Works}\label{sec:related}
    \subsection{Superpixels on saliency estimation}
    Early methods used single pixels or $n \times n$-blocks of pixels to compute contrast \cite{valenti2009image, itti1998model, achanta2008salient} but they usually lack well-defined separation between object and background, overly increasing the saliency of blocks adjacent to actual salient regions. To better define these regions, most modern methods adopt the usage of superpixels.
    
    Many methods have been proposed to super-segment the image into superpixels (\textit{e.g.,} \cite{achanta2012slic, felzenszwalb2004efficient, comaniciu2002mean, vargas2019iterative}), however, choosing which superpixel segmentation to use is a somewhat overlooked task when estimating saliency. Most methods opt for using SLIC \cite{achanta2012slic}, which is a fast grid-based segmentation method that creates regular superpixels (superpixels are of similar size and shape). Despite superpixel regularity being an essential feature for many applications, there is always a trade-off between regularity and object-boundary adherence.
    
    
    Additionally, recent advances in superpixel algorithms allow the usage of object information (\textit{e.g.,} saliency maps) to improve segmentation and provide control over the behavior of superpixels \cite{belem2019oisf, belem2019importance}. To the best of our knowledge, no saliency estimator has explored a saliency-based superpixel segmentation yet.

    \subsection{Priors and Queries}
   
    Most unsupervised saliency estimators model saliency using a combination of prior domain-specific knowledge, and salient characteristics extracted from the input image. The prior knowledge is used to create global assumptions that do not depend on image-specific characteristics: On natural images, for example, the salient object is most likely centered \cite{peng2016salient, shen2012unified}, focused \cite{jiang2013salient} and composed of vivid colors \cite{peng2016salient, shen2012unified}. It is not hard to imagine scenarios where these assumptions fail, and the results are sub-par.

    On the other hand, bottom-up information can be used to model saliency based on similarities of intrinsic low-level image information: For example, one may assume that regions with high color contrast to its adjacency are likely to be salient \cite{cheng2014global}. However, the over-segmentation of the regions (superpixels) may introduce errors. In this regard, several methods propose saliency to be computed on multiple scales \cite{lin2018saliency, zhang2018hypergraph, tong2014saliency, jiang2013drfi}, and then combined later on.

 Another common approach defines global query regions as representing the background or foreground. Assuming that the object is usually centered and fully enclosed on natural images, the most common background query selection strategy is to use regions on the image borders. However, when the object touches the image border, errors may occur. Multiple strategies have then been proposed to reduce the influence of miss-selected queries. One strategy is to combine multiple saliency maps using subsets of the boundary regions, say one map for each of the four sides (top, right, bottom, and left) \cite{zhang2018hypergraph, peng2016salient, zhang2016ranking}. Another option is to assign a confidence value to boundary-regions based on its part connected to the image border \cite{zhu2014saliency}.
    
        Instead of assuming the background to be on the image borders, another set of algorithms expect the image background to be composed of highly redundant information. They solve saliency estimation using \concept{low-rank matrix recovery} (LR) theory. LR-based methods use a low-rank feature matrix to approximate the background regions, and sparse salient object regions, on the other hand, are represented by a sparse sensory matrix. One method that stands out on this approach uses a \concept{Structured Matrix Decomposition (SMD)} \cite{peng2016salient} model, which adds connectivity constraints and a regularization step used to assist images with a cluttered background.
        
    Note that all approaches have to make assumptions based on prior knowledge of the image domain. In this regard, by pre-selecting a set of query strategies and top-down priors, even bottom-up strategies are constrained to specific scenarios.
    
    \subsection{Graph-based saliency estimation}
    
    In recent years, many methods have been proposed using graphs to model saliency \cite{zhu2014saliency, tong2014saliency, zhang2018hypergraph, yang2013saliency, wu2018salient}. Each image is represented by a graph, where the vertices are image regions (superpixels), and an edge connects two related vertices. Regions are usually connected to their adjacency and query regions, and saliency is estimated in a bottom-up manner.

    Yang \textit{et.al} \cite{yang2013saliency} proposed using the four image borders as background queries to compute four saliency maps using manifold ranking. Even though the multiple maps strategy reduces the error compared to using all background regions simultaneously, the resulting combination commonly highlights only parts of the salient objects. As a further step, the method thresholds the resulting background-map combination to use it as a foreground query to estimate the final saliency map. Wu \textit{et. al.} \cite{wu2018salient} use a similar framework, but they further improve the background queries by estimating how salient each border region is amongst themselves. 
    
    Zhu \cite{zhu2014saliency}, instead of computing multiple maps, propose a weighting function to determine the confidence of a border region to be the background. They also include a smoothness term to regularize the optimization of cluttered regions of the image.
    
    Taking closer attention to the role superpixels have in the process, Tong \cite{tong2014saliency} proposed computing saliency on multiple scales of superpixels and added a filtering property to improve edge preservation on the resulting map. They compute multiple single-scale saliency maps and integrates them by proposing an integration function that optimizes a pixel-to-superpixel similarity measure. Similarly, Zhang \cite{zhang2018hypergraph} proposes using multiple scales of superpixels to compute a background and foreground-query based hypergraph saliency estimator. They present their results using both foreground and background or using a single query. Combining both queries outperforms both other options.
    
    All the aforementioned graph-based strategies use a bottom-up only approach and do not leverage top-down prior knowledge. As shown by Peng \cite{peng2016salient}, combining both top-down and bottom-up strategies may be beneficial.
\section{\textit{Iterative Saliency Estimation fLexible Framework}}\label{sec:framework}

The \emph{ITerative Saliency Estimation fLexible Framework} (ITSELF) is a graph-based algorithm that leverages domain knowledge and low-level image information to estimate and enhance object-based superpixels and saliency iteratively. The interaction between superpixel-based saliency and saliency-based superpixels allows for an iterative enhancement cycle that characterizes the core of ITSELF (Figure \ref{fig:framework}). The framework's flexibility comes from user-defined query-regions and prior maps. Queries define examples of foreground/background regions to be compared to, while the priors enhance the initial estimation. 

On this work we represent an image as a pair $I = (\set{P},\function{I})$, in which $\set{P} \subseteq \mathbb{Z}^2$ is the set of pixels, and $\function{I} : \set{P}\rightarrow \mathbb{R}^{m}$. In this work, we consider colored and grayscale images --- \textit{i.e.,} $m \geq 1$. Similarly, we consider a saliency map to be a pair $SM = (\set{P},\function{S})$, in which $\function{S} : \set{P}\rightarrow \mathbb{R}^{1}$.

\subsection{Object-based Superpixel Segmentation}\label{subsec:superpixel-segmentation}
    Superpixel segmentation is partitioning an image into $n$ connected regions of pixels (\textit{i.e.,} superpixels) that share similar characteristics. Superpixels are extensively used in saliency estimation for reducing the algorithm's input, and provide better object boundary definition. However, when it comes to saliency estimators, the choice of which superpixel segmentation algorithm to use is somewhat overlooked. 
    
    Recent studies have allowed the incorporation of object information (\textit{e.g.}, saliency) into superpixel segmentation \cite{belem2019oisf, belem2019importance}. To the best of our knowledge, the only method that leverages saliency maps for superpixel segmentation is the Object-based Iterative Spanning Forest (\textit{OISF}) \cite{belem2019oisf}, an extension of the \textit{Iterative Spanning Forest} \cite{vargas2019iterative} framework. \textit{OISF} is composed of three main steps: (i) estimate a representative set with one pixel for each superpixel (namely seed set); (ii) form pixel groups according to how strongly connected they are to the seeds; (iii) recompute the seeds. Through $n_r$ iterations, the segmentation result is improved through subsequent executions of steps (ii) and (iii).

\begin{figure}[t!]
        \centering
        \begin{tabular}{c c}
                 \includegraphics[width=0.18\textwidth]{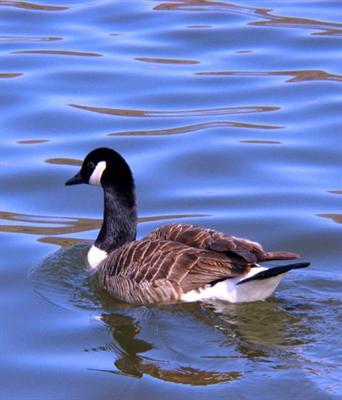} &
                 \includegraphics[width=0.18\textwidth]{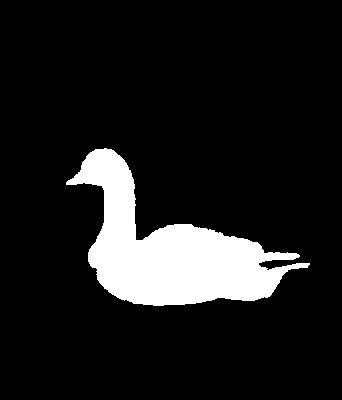} \\
                (a) & (b)
            \end{tabular}
            \begin{tabular}{c c}
                 \includegraphics[width=0.18\textwidth]{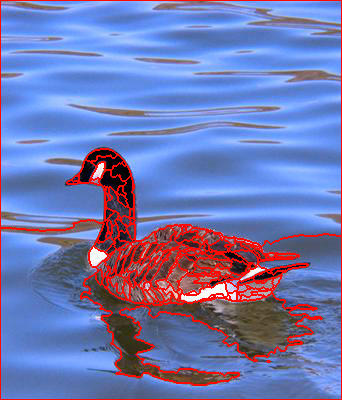} &
                 \includegraphics[width=0.18\textwidth]{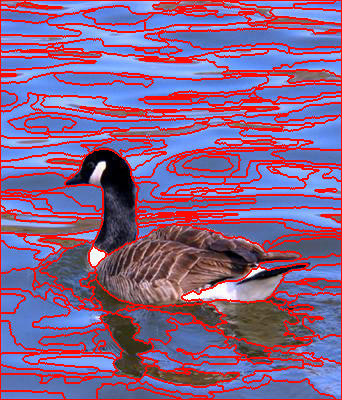} \\
                (c) & (d)
            \end{tabular}
        \caption{(a) Input image. (b) Ground-truth; (c) Super-segmented salient object \cite{belem2019oisf}; (d) Sub-segmented salient object. Both segmentation images were computed using OISF \cite{belem2019oisf} with 200 superpixels and the ground-truth as the input saliency map}
        \label{fig:oisf-example}
\end{figure}

     For seed estimation, \textit{Belem et.al} \cite{belem2019importance} proposed two approaches that takes object information into consideration, Object-based grid (OGRID) and Object Saliency  Map  sampling  by  Ordered  Extraction  (OSMOX). On both strategies, the user can control the number of superpixels and the percentage of object seeds $\rho \in [0..1]$ --- \textit{i.e.,} how many seeds will fall into salient regions (Figure \ref{fig:oisf-example}). However, OGRID loses saliency information by requiring a thresholded map when determining the salient regions. On the other hand, OSMOX is faster, has equivalent results, and takes advantage of the saliency map's nuances. 
    
    Let $n_o = n \rho$ to denote the number of object seeds. Briefly explaining OSMOX, $n_o$ seeds are selected from a priority queue of pixels, where the priority is defined according to the saliency of the pixel's neighbors. To ensure better seed distribution, for each pixel selected as object seed, the priority of adjacent pixels is reduced, and the priority queue is rearranged. The previous steps are repeated until the number of seeds is obtained --- it is analogous for $n - n_o$ background seeds. 
    
    With the seeds selected, \textit{OISF} runs the \textit{Image Foresting Transform} (\textit{IFT}) algorithm \cite{falcao2004ift} for delineating the superpixels. The \textit{IFT} computes an optimum-path forest, where the seeds are the roots of the trees, and optimality is defined in terms of a path-cost function. Non-seed pixels are aggregated to the tree that provides the minimum path-cost to it. Each tree of the resulting forest is taken as a superpixel.  
    
    Let $p, q \in \set{P}$, $\pi_p$ be a path with terminus $p$, $\pi_p \cdot \langle p, q\rangle$ be the extension of $\pi_p$ by $q$, and $r_p$ be the root of the tree $p$ belongs to. \textit{OISF} proposes the path cost to be additive, where the added value is derived from the color and the saliency difference between pixels:
    
    \begin{eqnarray} 
    &f(\pi_p \cdot \langle p, q\rangle) = f(\pi_p) + \|q - p\| +  \\  &  \left[\alpha\|\function{I}(r_p) - \function{I}(q)\|\gamma^{|\function{S}(r_p)-\function{S}(q)|} + \gamma|\function{S}(r_p)-\function{S}(q)|\right]^\beta, \nonumber, 
    \end{eqnarray}

    where $ \alpha $ controls the size regularity of the superpixels, $ \beta $ the border adherence, and $ \gamma $ indirectly controls the saliency score's influence when defining superpixel borders. Even though OISF's path-cost function may not satisfy some conditions to achieve optimality \cite{ciesielski2018path}, the resulting trees are suitable for image representation.
    
    After the first segmentation is finished, the results can be improved by recomputing the seeds and running another iteration of the pipeline. In this work, at each \textit{OISF} iteration, the seed is re-positioned to be the pixel whose color is closer to the mean color of the superpixel (\textit{i.e.,} the superpixel medoid on the feature space). Additionally, by enhancing the saliency map over iterations of \textit{ITSELF}, the next \textit{OISF} segmentation is being performed on an improved initial set of seeds.
    
    At each ITSELF iteration $t$, the number of superpixels may change in order to compute saliency on multiple scales. For that, we added a parameter $\kappa \in (0..1]$ that redefines the number of superpixels on each iteration: $n^{t+1} = n^t \kappa$; in which $n^1 = n$.
    
\subsection{Superpixel-based Saliency Estimation}\label{subsec:saliency-estimation}

    Let $\superset{S}$ be the set of all superpixels, $\set{S}, \set{R} \in \superset{S}$, and $\set{Q} \subset \superset{S}$ be the subset of query superpixels. We start by representing the image as a superpixel weighted graph $\mathcal{G} = (\superset{S}, \set{E})$ where the vertices are the superpixels and $\set{E} = \set{E}_\mathcal{A} \cup \set{E}_T \cup \set{E}_Q$, where $\set{E}_\mathcal{A}$ is the set of \emph{adjacency edges}, $\set{E}_T$ is the set of \emph{transitively extended edges}, and $\set{E}_Q$ of \emph{query edges}. The query edges connect every superpixel to every query, \textit{i.e.,} $\set{E}_Q = \{(\set{S},\set{R}) \in \superset{S}\times\set{Q} \; | \; \set{S} \neq \set{R} \}$. Let $\mathcal{A}_8 \subset \set{P}^2$ denote a 8-adjacency neighborhood; then, the adjacency edge set connect every vertex to its adjacents in the image domain, \textit{i.e.,} $E_{\mathcal{A}} = \{(\set{S},\set{R}) \in \superset{S}^2 \; | \; \exists \; (p,q) \in \mathcal{A}_8 \text{ for } p \in \set{S}, q\in \set{R}, \text{ and } \set{S} \neq \set{R}\}$. Lastly, the transitively edges extend the image adjacency by one level, $E_T = \{(\set{S},\set{R}) \in \superset{S}^2 \; | \; \exists \; \set{W} \in \superset{S} \text{ that } (\set{S}, \set{W}),(\set{W}, \set{R}) \in E_{\mathcal{A}}\}$.
    
    To allow for the user to control the importance of query over non-query edges, the edges start with parametrically defined weight $\function{e}(S,R)$, where query edges have weight $\psi \in [0..1]$ and adjacency edges have weight $1 - \psi$.
    
    We define dissimilarity in terms of color differences between superpixels. Let $\set{C}_I$ be the set of all unique colors that compose $I$, $\set{C}_S \subseteq \set{C}_I$ be the existing colors in a superpixel $\set{S}$, and $p(c, \set{S}), c \in \set{C}_S$ be the percentage of $c$ colored pixels on $\set{S}$. We incorporate the dissimilarity measure to the graph by updating the edge weights using a Gaussian function:
    
    \begin{equation}
        \function{e'}(\set{S},\set{R}) = \function{e}(\set{S},\set{R}) \sum\limits_{\forall c_i \in \set{C}_\set{S}}{\sum\limits_{\forall c_j \in \set{C}_\set{R}}{\exp^{-\frac{\|c_i - c_j\|}{\sigma_s}}  p(c_i, \set{S})  p(c_j, \set{R})}} ,
    \end{equation}
    
    where $\sigma_s \in (0, 1]$ is the variance and controls the rate in which the distance function increases, and $(\set{S}, \set{R}) \in \set{E}$.
    
    Then, let $\set{E}_F \subset \set{E}_Q$ be the subset of foreground-query edges. We invert the foreground query weights to account for similarity instead dissimilarity, defining \textit{vertex saliency} to be:
    
    \begin{equation}
        \function{VS}(\set{S}) = \sum_{\forall \set{R} \in \set{E}\setminus\set{E}_F}{\function{e'}(\set{S},\set{R})} + \sum_{\forall \set{F} \in \set{E}_F}{1 - \function{e'}(\set{S},\set{F})}.
    \end{equation}
    
    Finally, we incorporate the prior domain information simply by multiplying the saliency score of each vertex by the \emph{normalized combined prior map} $\function{PS}$ (detailed in Subsection \ref{subsec:prior-integration}) to get the final saliency score:
    
    \begin{equation}
        \function{S}(\set{S}) =  \function{VS}(\set{S}) \function{PS}(\set{S})
        \label{eq:saliency-computation}
    \end{equation}
    
    The resulting saliency image maps to each pixel $p$ the saliency value of its corresponding superpixel.
    
    Additionally, instead of taking the last iteration result as final, we used the prior integration step to combine the multiple computed maps. We only discard the first estimated map as it often highlights a big part of the background.
    
\subsection{Prior map integration}\label{subsec:prior-integration}
    
\begin{figure}[t!]
        \centering
        \begin{tabular}{c c}
                 \includegraphics[width=0.18\textwidth]{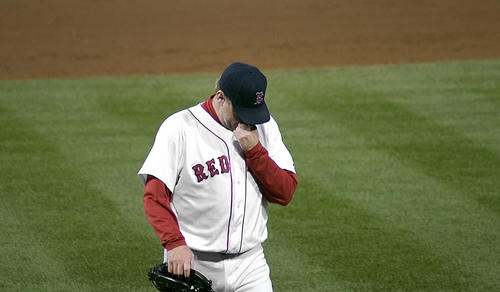} &
                 \includegraphics[width=0.18\textwidth]{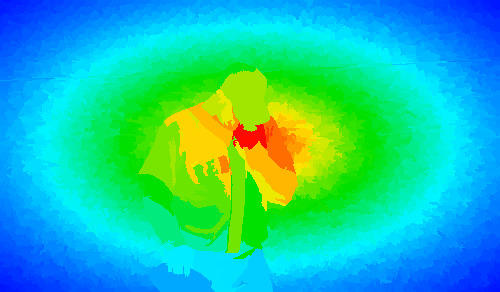} \\
                (a) & (b)
            \end{tabular}
            \begin{tabular}{c c}
                 \includegraphics[width=0.18\textwidth]{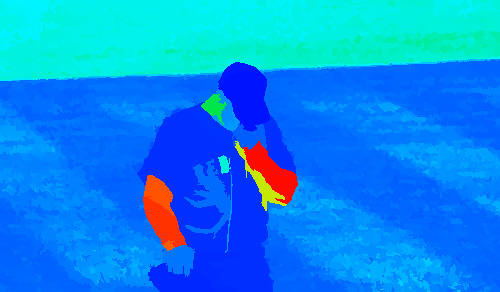} &
                 \includegraphics[width=0.18\textwidth]{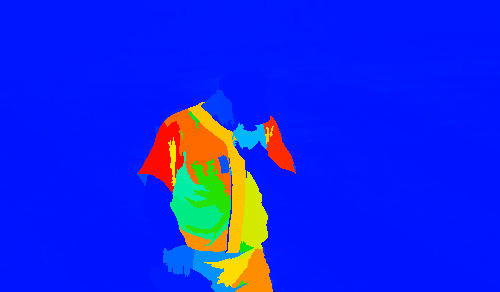} \\
                (c) & (d)
            \end{tabular}
            \begin{tabular}{c c}
                 \includegraphics[width=0.18\textwidth]{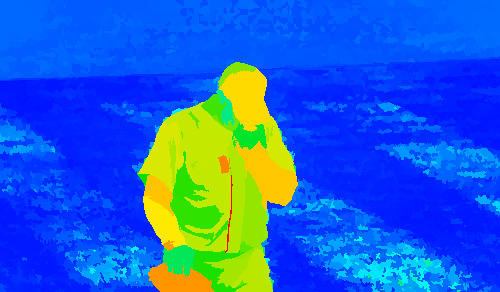} &
                 \includegraphics[width=0.18\textwidth]{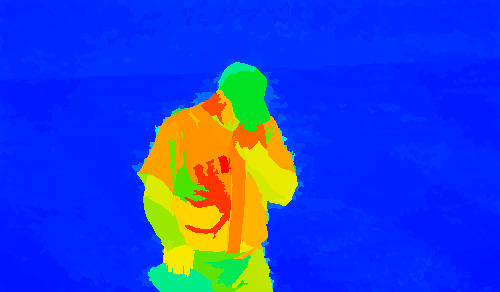} \\
                (e) & (f)
            \end{tabular}
        \caption{(a) Input image. (b) Center-surround prior; (c) Red-yellow prior; (d) White prior; (e) Global color contrast prior; (f) Combined priors.}
        \label{fig:priors_example}
\end{figure}

    Prior knowledge of how humans perceive saliency allows for assumptions to be drawn on which characteristics are determinant when defining saliency. These assumptions alone are often insufficient to accurately identify salient regions. However, by combining multiple priors, it is possible to create more accurate models (Figure \ref{fig:priors_example}).
    
     We propose ITSELF to be flexible and allow any number of priors to be incorporated into the model. For such, we provide a combination strategy that allows multiple prior maps to be combined into a single map. The resulting prior map is then used during the saliency estimation step (Equation \ref{eq:saliency-computation}).
    
    Yao Qin \textit{et.al} \cite{qin2015saliency} has proposed an iterative saliency-estimation algorithm that uses Cellular-Automata and a Bayesian framework to combine multiple saliency maps. A cellular automaton is a collection of grid-disposed cells that evolve iteratively according to their neighbors' state and a set of rules. The automaton cells are the pixels of all saliency maps organized so that they form a 3D grid $G = (\mathcal{C}, \function{S}^t_\oplus)$, lastly updated on iteration $t$. For an ordered list of saliency maps $\langle SM_1 .. SM_m\rangle$, $\mathcal{C} = \{(x_p, y_p, i) \in \set{P}^2\times \mathbb{N}^* \; | \; p = (x_p, y_p) \in \set{P}_i\}$ and $\function{S}_{\oplus}^1(\mathtt{c}_i) = \log(\function{S}_i(p))$, for $SM_i = (\set{P}_i, \function{S}_i)$, and $1 \leq i \leq m$.
    
    Let $\mathtt{c}_i = (x_p, y_p, i)$, $\mathtt{c}_j = (x_q, y_q, j) \in \mathcal{C}$, we consider that a cell impacts the evolution of another if they are connected by a \emph{cuboid adjacency relation} $\mathcal{A}_\square$. We formally define the cuboid adjacency to be $\mathcal{A}_{\square} = \{(\mathtt{c}_i,\mathtt{c}_j) \in \mathcal{C} \; | \; \exists \; (p,q) \in \mathcal{A}_4 \text{ which } p = (x_p, y_p) \text{ and } q = (x_q, y_q)\}$.
    
\begin{figure}[t!]
        \centering
        \begin{tabular}{c c c}
                 \includegraphics[width=0.13\textwidth]{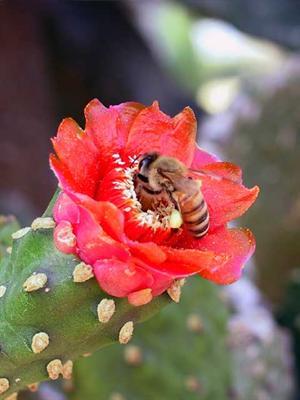} &
                 \includegraphics[width=0.13\textwidth]{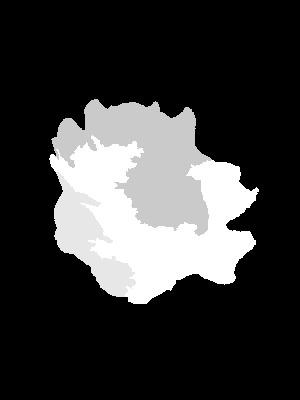} &
                 \includegraphics[width=0.13\textwidth]{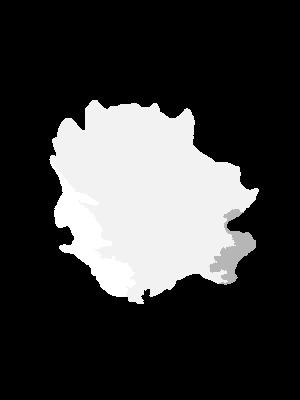} \\
                 \includegraphics[width=0.13\textwidth]{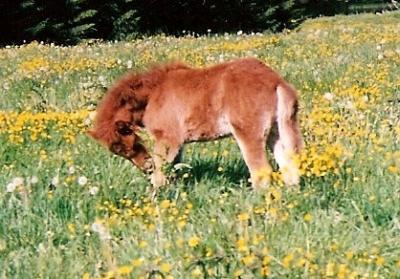} &
                 \includegraphics[width=0.13\textwidth]{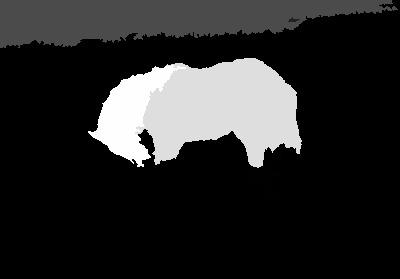} &
                 \includegraphics[width=0.13\textwidth]{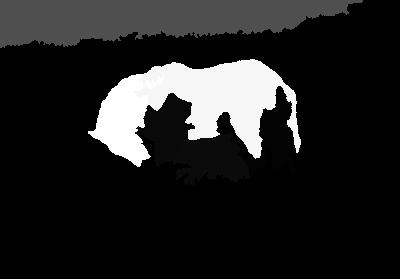} \\
                (a) & (b) & (c) \\
            \end{tabular}
        \caption{(a) Input image. (b-c) Result of saliency map integration using $\lambda \in \{0.01, 0.1\}$, respectively. Note that although a higher $\lambda$ creates more homogeneous salient objects, less salient object parts may be lost.}
        \label{fig:integration-lambda}
\end{figure}

    We consider a cell $\mathtt{c}_i$ to be salient if its saliency score is higher than the mean saliency value ($\mu_i$) of its originating map $SM_i$. The saliency of each cell is increased over time ($t$) based on the saliency of its neighborhood:
    
    \begin{equation}
        \function{S}_{\oplus}^t(\mathtt{c}_i) = \function{S}_{\oplus}^{t-1}(\mathtt{c}_i) + \Lambda \sum\limits_{\forall \mathtt{c}_j \; | \; (\mathtt{c}_i,\mathtt{c}_j) \in \mathcal{A}_{\square}}{sign(\function{S}_{\oplus}^{t-1}(\mathtt{c}_j) - \mu_j)}
    \end{equation}{}
    
    where $\Lambda \in (0..1]$ is a constant that controls the strength of the update (Figure \ref{fig:integration-lambda}). The final updated saliency score of each pixel on each map is than normalized: $\function{S}^t_{\oplus}(\mathtt{c}_i) \leftarrow \frac{\exp^{\function{S}^t_{\oplus}(\mathtt{c}_i)}}{(1 + \exp^{\function{S}^t_{\oplus}(\mathtt{c}_i)})}$.
    
    After $t$ iterations, the final saliency map is the cell combination on the $z$ coordinate:
    
    \begin{equation}
        \function{PS}(p) = \frac{1}{m}\sum\limits_{i = 1}^{m}{\function{S}^t_{\oplus}(\mathtt{c}_i)},
    \end{equation}
    
    where $\mathtt{c}_i = (x_p, y_p, i)$.
    
    The same automaton is used on every iteration of ITSELF; however, the number of cells may change if more priors are added on later iterations. We also use this prior integration step to combine the output saliency maps from each ITSELF's iterations. By changing the number of superpixels used to compute each saliency map, the automaton is considering multiple scales.
\section{Query Selection}\label{sec:query-estimation}
    We propose three different approaches to estimate queries: (A) border-based query, assuming most of the image boundary regions are background on natural images; (B) saliency-based, used to incorporates any pre-computed saliency map into the framework.
    
\subsection{Border-based Query Selection}\label{subsec:border-based-query}
    
    We propose combining both boundary connectivity \cite{zhu2014saliency} and multi-map estimation \cite{zhang2018hypergraph, peng2016salient, zhang2016ranking} to further reduce the miss-selection of background regions. For such, instead of using the four sides of the image as the multiple maps, we propose clustering the superpixels based on their color similarity. We then compute a saliency map for each of the clusters that contain at least one superpixel on the image boundary. During this computation, we only use the cluster's superpixels touching the image boundaries as queries.
    
    Any clustering algorithm could be used; however, we opted on using the \textit{Unsupervised Optimum-Path Forest (OPF)} \cite{rocha2009data}: A graph-based algorithm that performs clustering by solving an optimum-path forest on a graph of samples. OPF finds an adequate number of clusters $g$ automatically; therefore, different images may have a different number of clusters. 

    Let $\superset{S}_g \ \subset \superset{S}$ be the set of superpixels contained in cluster $g$, and $\superset{B}_g \subset \superset{S}_g$ be the set of superpixels in $\superset{S}_g$ that touches the image borders. For each cluster, we compute a saliency map $\function{SM}_g$ using Equation \ref{eq:saliency-computation} and a boundary-connectivity score $w_g$. In this work, the \concept{boundary connectivity score} of cluster measures how many of its superpixels touch the image border, and is defined as $w_g = |\superset{B}_g| / |\superset{S}_g|$.
    
\begin{figure}[t!]
        \centering
        \begin{tabular}{c c}
                 \includegraphics[width=0.18\textwidth]{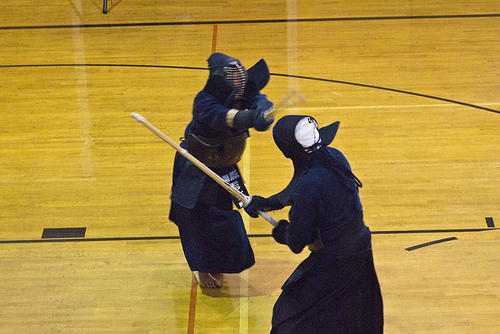} &
                 \includegraphics[width=0.18\textwidth]{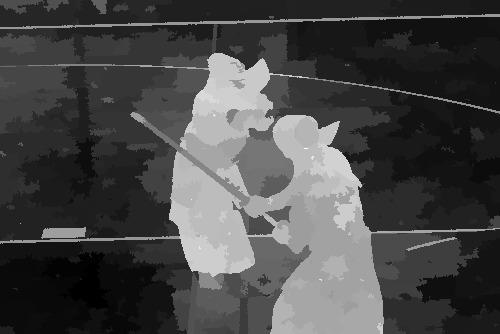} \\
                (a) & (b)
            \end{tabular}
            \begin{tabular}{c c}
                 \includegraphics[width=0.18\textwidth]{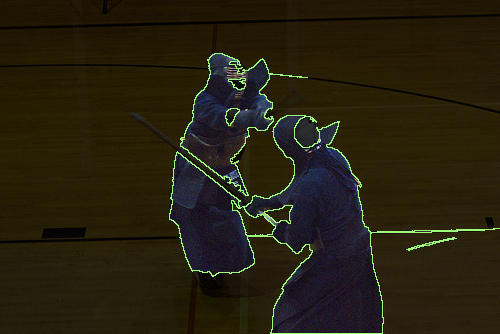} &
                 \includegraphics[width=0.18\textwidth]{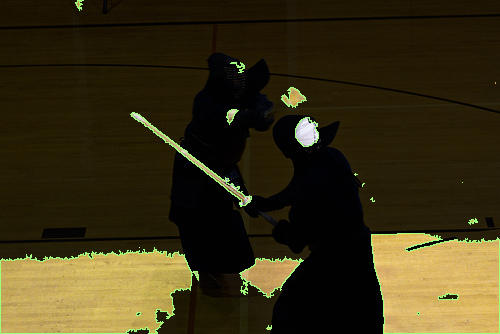} \\
                (c) & (d)
            \end{tabular}
        \caption{(a) Input image. (b) The result combination of the boundary clusters; (c) The highest boundary-connectivity score cluster with $w_c = 0.453$; (d) A boundary cluster containing most of the object with boundary-connectivity score $w_c = 0.142$. Note that the combined saliency map is not the final result of ITSELF, rather it is the simple combination of the boundary clusters.}
        
        \label{fig:boundary-clustering}
\end{figure}

    We then perform a weighted average to attribute to each superpixel a single saliency score:
    
    \begin{equation}
        \function{CS}(\set{S}) =\frac{1}{W}\sum_{g}^{n_g}  w_g SM_g(S)
    \end{equation}
    
    where $W = \sum_{g}^{n_g} w_g$. With the saliency score of each superpixel, the final saliency map is the propagation of the saliency scores of each superpixel to every pixel that composes it. A visual representation of the combination of clustering and the boundary-connectivity score is depicted in Figure \ref{fig:boundary-clustering}.
    
    \subsection{Saliency-based Query Selection}\label{subsec:saliency-based-query}
    
    Queries are subsets of image regions that are representative when estimating saliency, given a set of predicates. Whether the queries are good representations of foreground or background, the importance of such regions can be encoded by a saliency map. Thus, given a saliency map $\function{SS}$ and a threshold $\mu$, a superpixel is selected as a query if its saliency value exceeds the threshold.
\section{Prior Modeling}\label{sec:prior-estimation} 
    In this section, we present the models we implemented for each prior used in our experiments. The prior maps are represented the same way as a saliency map, however, to easily differentiate between both, the prior maps on this paper are represented as heat maps where lower values are represented on cold colors(blue$\rightarrow$green) and higher values on hot colors(yellow$\rightarrow$red).

\subsection{Center-surround prior}\label{subsec:center-surround-prior}
    
    A widespread assumption for natural images is that the salient object will be near the center of the image \cite{cheng2014global, shen2012unified}. In this regard, let $p_c \in P$ be the center pixel of the image. The center distance of a superpixel to the image center is defined as: $\function{CD}(\set{S}) = \frac{1}{|S|}\sum\limits_{\forall q \in \set{S}}{\|q - p_c\|}$.
    
    However, to change the increase rate of the distance function, the values are put into a Gaussian centered on $p_c$, thus, the center prior score is defined as follows:
    
    \begin{equation}
        \function{CP}(\set{S}) = \exp^{-\frac{\function{CD(S)}}{\sigma_{1}^2}}
    \end{equation} 
    
\begin{figure}[t!]
    \centering
    \begin{tabular}{c c}
             \includegraphics[width=0.18\textwidth]{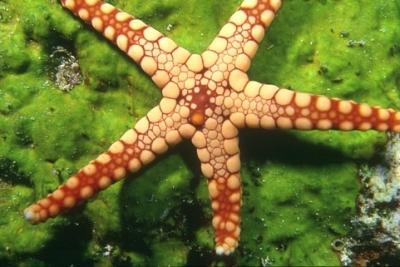} &
             \includegraphics[width=0.18\textwidth]{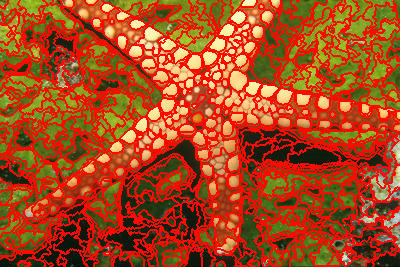} \\
            (a) & (b)
        \end{tabular}
        \begin{tabular}{c c}
             \includegraphics[width=0.18\textwidth]{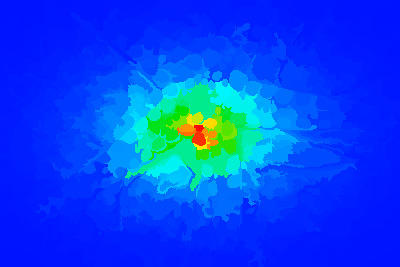} &
             \includegraphics[width=0.18\textwidth]{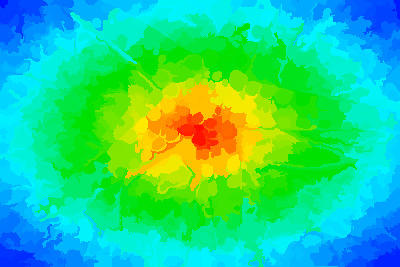} \\
            (c) & (d)
        \end{tabular}
    \caption{(a) Original image. (b) Superpixel Segmentation; (c) and (d) Center prior maps with $\sigma_1=0.1$ and $\sigma_1=0.9$, respectively.}
    
    \label{fig:center-sigma-showcase}
\end{figure}

    Smaller values of $\sigma_{1} \in (0..1)$ causes superpixel farthest to center to be less relevant (Figure \ref{fig:center-sigma-showcase}).
    
\subsection{Global color uniqueness prior}\label{subsec:global-color-prior}
    
    The saliency score represents how much an object stands-out in a scene. A defining characteristic when estimating said score is color uniqueness. Colors that appear the least are rarer in the image and may stand out \cite{cheng2014global, jiang2013salient}.
    
    Let $\set{C}_I$ be the set of all unique colors that compose $I$, $\set{P}_c \subset \set{P}$ be the set of all pixels of color $c \ in \set{C}_I$, and $p(c) = \frac{|\set{P}_c|}{|\set{P}|}$. Similar to the center prior, we use a Gaussian function to control the increasing rate of the the distance measure. In this regard, we define the \textit{Color Uniqueness Score} as $\function{US}(c) = \exp^{\frac{p(c)}{\sigma_2^2}}$. 
    
    However, even after quantization, there are several similar colors (\textit{e.g.,} slightly different tones of the same color), creating artifacts counter-intuitive to the human perspective. To reduce the impact of similar color uniqueness, similar to Cheng \textit{et. al.} \cite{cheng2014global}, we smooth the uniqueness score based on the average uniqueness of similar colors. For every pair $(c_i, c_j), c_i \neq c_j$, we define the color similarity weight to be $\function{ws}(c_i, c_j) = \exp^{-\frac{\|c_i,c_j\|}{\sigma_2^2}}$, and prose the final global color-uniqueness score to be:
    
    \begin{equation}
        \function{US}'(c_i) = \sum\limits_{\forall c_j \in \set{C}_I}{\function{US}(c_j) \function{ws}(c_i, c_j)}
    \end{equation} 
    
\begin{figure}[t!]
    \centering
    \begin{tabular}{c c}
             \includegraphics[width=0.18\textwidth]{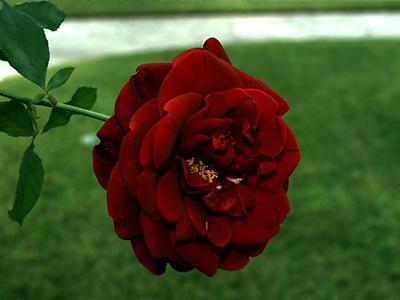} &
             \includegraphics[width=0.18\textwidth]{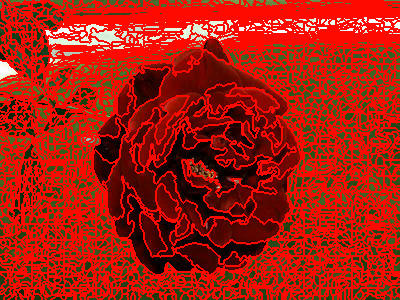} \\
            (a) & (b)
        \end{tabular}
        \begin{tabular}{c c}
             \includegraphics[width=0.18\textwidth]{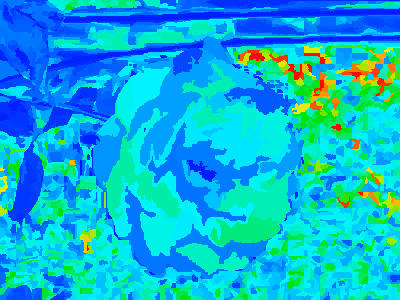} &
             \includegraphics[width=0.18\textwidth]{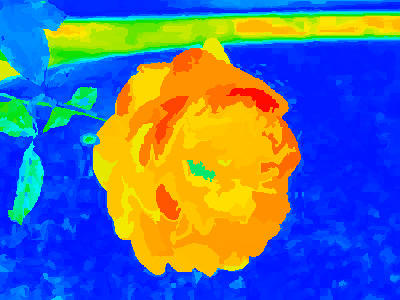} \\
            (c) & (d)
        \end{tabular}
    \caption{(a) Original image. (b) Superpixel Segmentation; (c) and (d) Global color-contrast prior maps without and with the smoothness step, respectively. Note how slight changes on tones of green impact negatively the method without smoothness.}
    
    \label{fig:global-contrast-showcase}
\end{figure}

    Figure \ref{fig:global-contrast-showcase} depicts the improvement when smoothness is applied.
    
    Each superpixel is then assigned a value according to the colors of the pixels that composes it: $\function{GP}(\set{S}) = \frac{1}{\|\set{C}_S\|} \sum\limits_{\forall c \in \set{C}_S}{p(c, \set{S}) \function{US}'(c)}$.
    
\subsection{Color-based priors}\label{subsec:color-prior}
    Based on observation of the Human Visual System, a common assumption is that red and yellow tones are naturally salient.
        
    Identifying red and yellow colors is straightforward in the \textbf{L*a*b*} colorspace: higher values on the \textbf{a} channel describe red tones,while high values on the \textit{b} channel, yellow. Therefore, we define a red/yellow score $\function{RY}(c)$ to be the sum of its \textit{a} and \textit{b} channels. As in the previous priors, we use a Guassian function to exert control over the functions increase rate, redefining the score to be $RY'(c) = \exp^{\function{RY}(c)/ \sigma_3^2}$.
    
    Lastly, we propagate the color values to every superpixel using a weighted average:
    
    \begin{equation}
        \function{RP}'(\set{S}) = \sum\limits_{\forall c \in \set{C}_S}{p(c, \set{S}) \function{RY}'(c))}
        \label{eq:color}
    \end{equation} 
    
\begin{figure}[t!]
    \centering
    \begin{tabular}{c c}
             \includegraphics[width=0.18\textwidth]{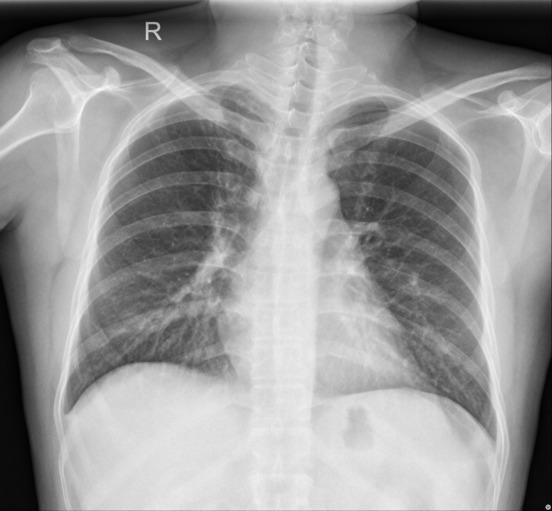} &
             \includegraphics[width=0.18\textwidth]{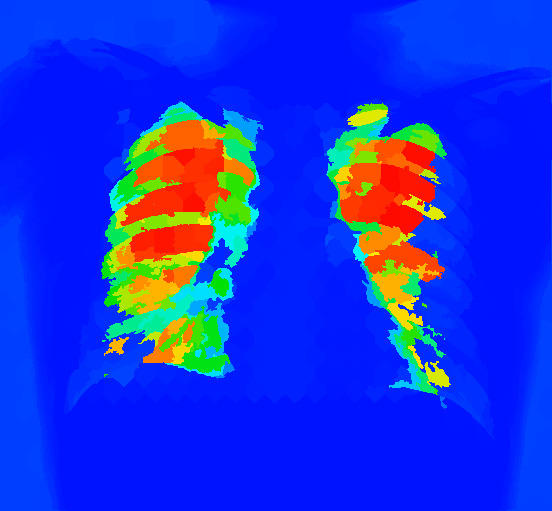} \\
            (a) & (b)
        \end{tabular}
    \caption{(a) Original image with object scribbles in light-blue. (b) Black prior map with $\sigma_3=0.5$. Additionally, we reduced the saliency of black regions connected to the image boundaries because of the natural color of the xray plate.}
    
    \label{fig:black-prior-showcase}
\end{figure}

    Although red and yellow are the most naturally salient colors, the same algorithm can be applied to other colors when required for specific objects. As an example, we know that on x-ray images of the thorax, the lungs are often darker than the other structures. So, we implemented a color prior that highlights black regions (Figure \ref{fig:black-prior-showcase}).
    
    All color-based priors follow the same principle of adding or subtracting the \textbf{L*a*b*} channels: white and black requires \textbf{a} and \textbf{b} to be closer to 0, with white having higher values on the \textbf{L} channel; the more negative the value of the \textbf{a} channel, the greener the color tone is, and the same goes for blue on the \textbf{b} channel. Changing the operations done on the channels yields new color priors.
    
\subsection{Saliency-based priors}\label{subsec:color-smoothing-prior}
    Due to the iterative nature of \textit{ITSELF}, we created a method that extrapolates a prior map from a previously computed saliency map, trying to reduce spurious saliency values of background regions sharing non-salient colors. 
    
\begin{figure}[t!]
    \centering
    \begin{tabular}{c c}
             \includegraphics[width=0.2\textwidth]{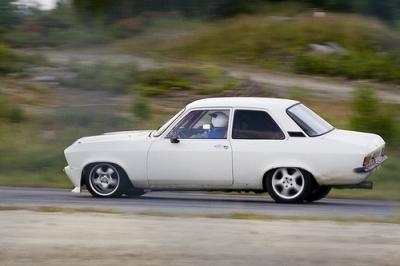} &
             \includegraphics[width=0.2\textwidth]{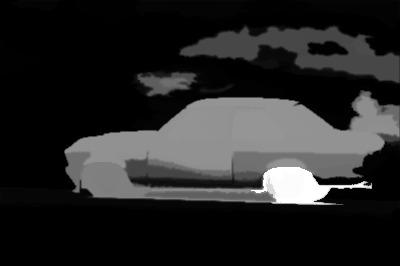} \\
            (a) & (b)
        \end{tabular}
        \begin{tabular}{c c}
             \includegraphics[width=0.2\textwidth]{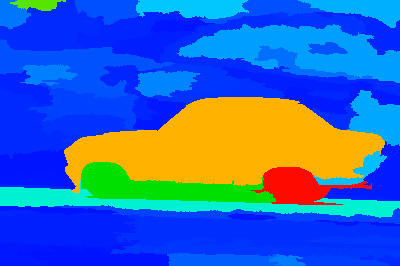} &
             \includegraphics[width=0.2\textwidth]{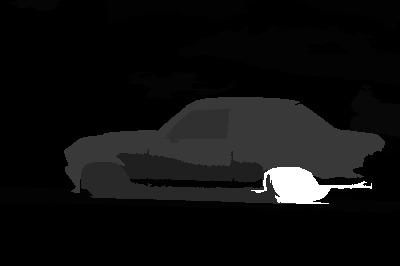} \\
            (c) & (d)
        \end{tabular}
    \caption{(a) Original image. (b) Saliency map before combining the proposed color-saliency based prior; (c) the color-saliency based prior derived from (b); (d) the result of combining the proposed prior.}
    
    \label{fig:saliency-based-prior-showcase}
\end{figure}

    We propose a color-saliency prior that attributes a saliency value to a superpixel depending on how globally salient its colors are (Figure \ref{fig:saliency-based-prior-showcase}).
    
    The global saliency of a color is defined as:
    
    \begin{equation}
        \function{SC}(c) = \frac{1}{|\set{P}_c|}\sum\limits_{\forall p \in \set{P}_c}{\function{S}(p)}
    \end{equation}
    
    Let $\set{C}_I$ be the set of all unique colors that compose $I$, $\set{C}_S \subset \set{C}_I$ be the subset of colors that compose a superpixel $\set{S}$. We then compute the saliency prior score to each superpixel, combining their color scores:
    
    \begin{equation}
        \function{CS}(\set{S}) =\frac{1}{|\set{C}_S|} \sum\limits_{\forall c \in \set{C}_S}{\function{SC}(c) }
    \end{equation}
    
    Despite only presenting a color-based saliency prior, other features (\textit{e.g.,} texture, shape, or size) could be used to create new priors similarly.

\subsection{Focus prior}\label{subsec:focus-prior}
    One could draw a natural correlation between focus and saliency. When observing a picture with different focal points, our eyes naturally ignore blurred regions prioritizing focused ones. Accordingly, identifying focused regions can improve the saliency estimation task \cite{jiang2013salient}.
    
    As presented by Jiang \textit{et.al} \cite{jiang2013salient}, the focus of a region is closely related to its degree of blur. With blurriness being the lack of sharply defined edges, \textit{focusness} is more easily quantifiable by looking at the edges of objects rather than their interior. Consequently, the first step when computing focusness is to identify object edges on the image. There are several edge detection algorithms available in the literature, however, we opted on using a simple thresholded gradient image. Let the gradient $\nabla(p) = \|\function{I}(p), \function{I}(q)\| \forall q \in \mathcal{A}_4$, and $\set{P}_e \subset P$ be the set of edge pixels. We consider that $p \in \set{P}_e \leftrightarrow \nabla(p) > \omega$, where $\omega$ is the Otsu threshold \cite{otsu1979threshold} of $\mathcal{I}$.
    
\begin{figure}[t!]
    \centering
    \begin{tabular}{c c}
             \includegraphics[width=0.18\textwidth]{figs/focus_flower.jpg} &
             \includegraphics[width=0.18\textwidth]{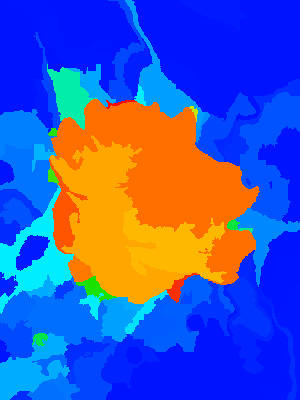} \\
            (a) & (b) \\
             \includegraphics[width=0.18\textwidth]{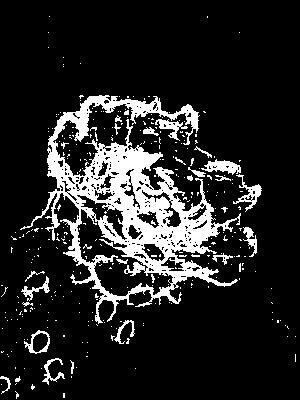} &
             \includegraphics[width=0.18\textwidth]{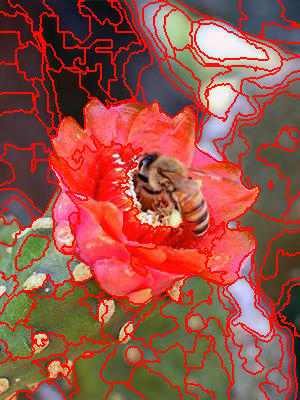} \\
            (c) & (d)\\
        \end{tabular}
    \caption{(a) Original image. (b) Result of the focus prior. (c) Estimated object edges; (d) Object-based superpixel segmentation. }
    
    \label{fig:focus-prior-showcase}
\end{figure}

    Within \textit{ITSELF}, the regions are delimited by superpixels and, thus, the focusness score can be defined by correlating superpixels to the detected edges. Superpixel segmentation also uses gradient information, and the created superpixel boundaries are frequently located in regions with a higher gradient. However, in blurred regions, the natural image edges will not exceed the threshold and will not be present on the estimated edges. In this regard, focused regions should have a higher match between superpixel boundaries and sharp image edges (Figure \ref{fig:focus-prior-showcase}).
    
    Let $\set{P}_b \subset{S}$, $p_b \in \set{P}_b$ be a boundary pixels of $\set{S}$ --- \textit{i.e.,} $\exists q \in \{\mathcal{A}_8(p_b), \set{R}\}, \set{R} \neq \set{S}$. We define the focusness score of $\set{S}$ by:
    
    \begin{equation}
        \function{FS}(\set{S}) = \frac{|\set{P}_b \cap \set{P}_e|}{|\set{P}_b|}
    \end{equation}
    
   Like the other priors, we map the focusness score to its location within a Gaussian function:
    
     \begin{equation}
        \function{FP}(\set{S}) = 1 - \exp^{-\frac{\function{FS}(\set{S})}{\sigma_4^2}}
    \end{equation}

\subsection{Ellipse-matching prior}\label{subsec:ellipse-matching-prior}
    
    Thanks to OISF's capability of representing objects with few superpixels, shape-based priors are viable. As an example, we created a prior that highlights elliptical objects to increase ITSELF's precision on an in-house dataset of intestinal parasite eggs (Section \ref{sec:datasets}). 
    
    To score how elliptical the superpixels are, we first compute a Tensor Scale Representation (\textit{TSR}) of each of them. The \textit{TSR} of a homogeneous region is a parametric representation of the best fit ellipse enclosed inside the region. Each ellipse is defined through its orientation (the angle between the ellipse's major axis and the image's y-axis), its anisotropy (the ratio between its major and minor axis), and thickness (size of the minor-axis). To compute the \textit{TSR} for every superpixel, we use a slightly modified version of Miranda's optimized algorithm \cite{miranda2005tsd}. The algorithm consists of identifying the edges of the homogeneous regions, finding the orientation of the best-fit ellipse, and computing the length of the ellipse's semi-axis.
    
    The main difference of our implementation when compared to Miranda's is when defining the region edges. Let $p \in \set{S}, q \in \set{R}, \set{S} \neq \set{R}$, and $\set{P}_e \subset{P}$ be the set of all region edges. We consider that $p \in \set{P}_e \leftrightarrow \exists q \in \mathcal{A}_8(p)$.
    
    The last two stages are implemented as described in \cite{miranda2005tsd}, taking the superpixels as the homogeneous regions and the center pixel of the superpixel as the center of the ellipse.
    
\begin{figure}[t!]
    \centering
    \begin{tabular}{c c}
             \includegraphics[width=0.18\textwidth]{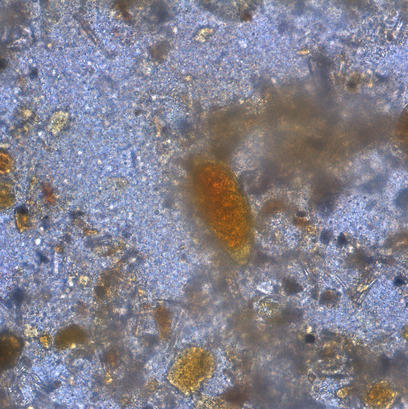} &
             \includegraphics[width=0.18\textwidth]{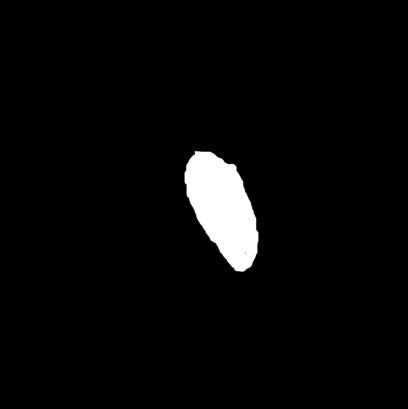} \\
            (a) & (b) \\
             \includegraphics[width=0.18\textwidth]{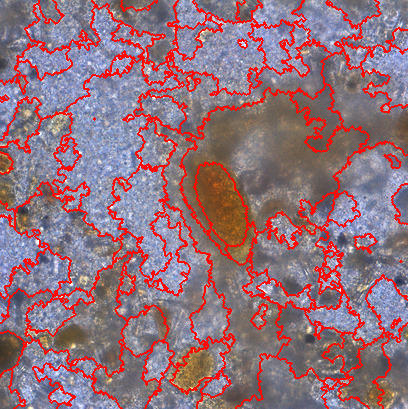} &
             \includegraphics[width=0.18\textwidth]{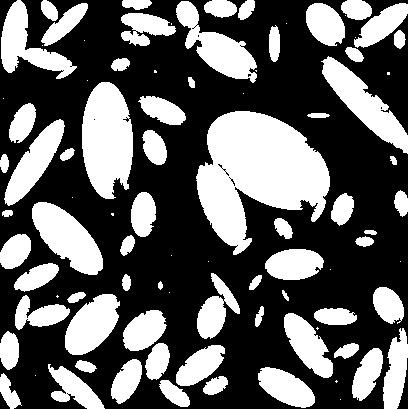} \\
            (c) & (d)\\
        \end{tabular}
    \caption{(a) Original image. (b) Object mask of the parasite. (c) Superpixel segmentation; (d) Ellipse Matching of each superpixel }
    
    \label{fig:ellipse_matching}
\end{figure}

    Afterwards, we estimate an ellipse matching for each superpixel (Figure \ref{fig:ellipse_matching}):
    
    \begin{equation}
        \function{EM}(\set{S}) = \frac{1}{|\set{S}|}\sum\limits_{\forall p \in \set{S}}{\delta_e([\|p, f_1\| + \|p, f_2\|] < 2l)},
    \end{equation}
    
\begin{figure}[t!]
    \centering
    \begin{tabular}{c c}
             \includegraphics[width=0.2\textwidth]{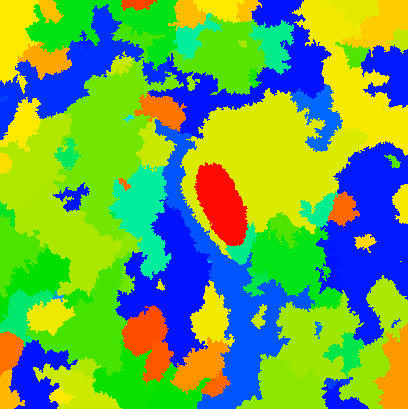} &
             \includegraphics[width=0.2\textwidth]{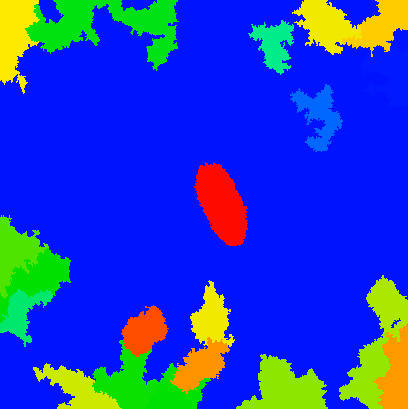} \\
            (a) & (b)
        \end{tabular}
    \caption{(a) Ellipse-based prior without size filtering (b) Result of reducing region saliency by size. }
    
    \label{fig:ellipse_prior}
\end{figure}
    
    where $\delta_e(\cdot) = \{0,1\}$ determines whether a pixel is positioned inside its respective ellipse, and $f_i$ are the estimated \textit{foci}.
    
    The final Ellipse-matching prior is also weighted by a Gaussian and is computed as follows:
    
      \begin{equation}
        \function{EP}(\set{S}) = 1 - \exp^{-\frac{\function{ES}(\set{S})}{\sigma_5^2}}
    \end{equation}
    
    Specific to the parasite dataset, we improve the ellipse prior result by adding a size constraint:
    
    \begin{eqnarray}
        \function{EP}'(\set{S}) & = & \left\{\begin{array}{ll}
        \function{EP}(\set{S}) & \mbox{ if $|S| \in (s_0, s_1) $,}\\
        min(\function{EP}(\superset{S})) & \mbox{otherwise,}
        \end{array}\right. \label{eq.ellipse-size}  
    \end{eqnarray}
    
    where $s_0$ and $s_1$ are, respectively, the lower and upper limits of the size range defined empirically. Figure \ref{fig:ellipse_prior} shows the improvement achieved by size filtering.

\subsection{Scribbles based priors}\label{subsec:scribble-based-prior}
    Object saliency maps have been frequently used to assist interactive object segmentation \cite{falcao2019role, yang2010user, tang2013grabcut}. In this scenario, the objects' locations are given by the user who interactively places scribbles in the object and background.
    
\begin{figure}[t]
    \centering
    \begin{tabular}{c c}
             \includegraphics[width=0.18\textwidth]{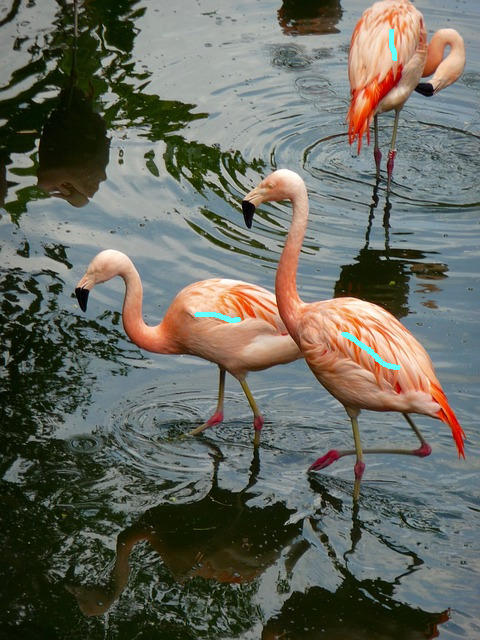} &
             \includegraphics[width=0.18\textwidth]{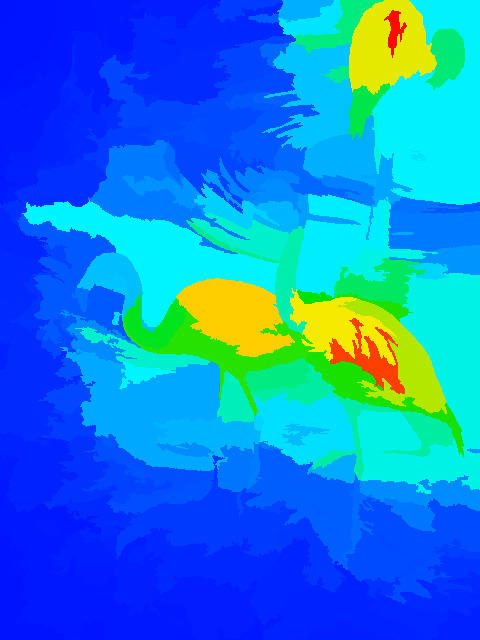} \\ 
             (a) & (b) \\
             \includegraphics[width=0.18\textwidth]{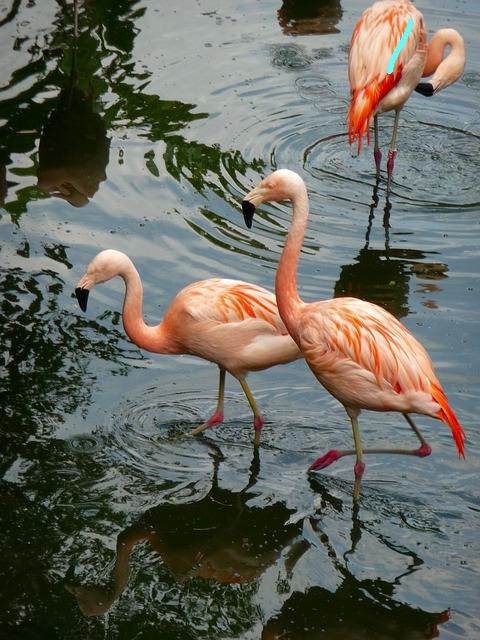} &
             \includegraphics[width=0.18\textwidth]{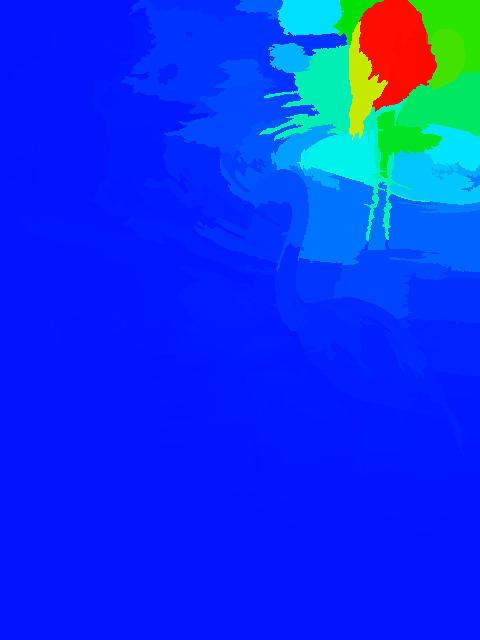} \\
             (c) & (d)
        \end{tabular}
    \caption{(a) Original image with three object scribbles in light-blue. (b) Scribble-based location prior map of (a); (c) Original image with a single object scribble in light-blue. (d) Scribble-based location prior map of (c).}
    
    \label{fig:scribbles_prior_showcase}
\end{figure}

    These user placed scribbles can be used as precise object detection, allowing the creation of several new priors with high accuracy. As a simple example, we can create location priors (similar to \concept{center-surround}) regarding the detected objects: A point has brighter values when they are close to object scribbles or far from background ones. Figure \ref{fig:scribbles_prior_showcase} shows the result of using object scribbles as a location prior. 
    
    It is worth noting that scribble based location priors can be used, for instance, segmentation. The challenge in instance segmentation is to individually segment objects of the same class in a picture with multiple objects. Take the single scribble example in Figure \ref{fig:scribbles_prior_showcase}: There are multiple flamingos on the image, but one may only interested in the top right flamingo. To fulfill such needs, we use the location scribble-based prior, reducing the saliency of all other objects that are not close to the user-provided marker.
    
    Note that other highly accurate priors could be created by using scribbles. They could be based on color, texture, or even shape and size by exploring object-based superpixels. We strongly advise exploring these possibilities in further works.
\section{Experiments and Results}\label{sec:experiments}
    We compare ITSELF to two other popular saliency estimators, namely the \textit{DiscriminativeRegional Feature Integration Approach} (DRFI) and the \textit{Structured Matrix Decomposition} (SMD). Assuming the background to be more homogeneous than the foreground, SMD uses the low-rank (LR) matrix theory to approximate the redundant background regions on a low-rank feature matrix, while a sparse sensory matrix represents sparse salient object regions. They use connectivity constraints, and a regularization step used to assist images with a cluttered background. Additionally, SMD incorporates location, color, and background priors to improve its results further.
    
    Despite DRFI being a supervised algorithm, it uses only hand-crafted features extracted from the input image, using a Random Forest to combine them and form the saliency score. The features are similar to other unsupervised methods (color, texture, and guess location). They compute multiple saliency scores on multiple scales and combine them at a fusion step. By learning the importance of each feature for different datasets, DRFI can be more easily extensible to other image domains.
    
    We did not include comparisons to the state-of-the-art graph-based algorithms because there was no code available, and we could not run the same experiments and evaluate using the same metrics as we did the others. However, they have the same inflexibility of the other methods, incorporating pre-selected assumptions into their models.
    
\subsection{Datasets}\label{sec:datasets}
    To validate our method, we used four popular natural image datasets: the \textbf{MSRA10K} \cite{liu2010learning}, which is the largest dataset selected (10000 images) and is composed of images with a singular salient object and a somewhat simple background; the \textbf{ECSSD} dataset \cite{yan2013hierarchical}, containing 1000 images of a singular salient object in a complex background; the \textbf{DUT-OMRON} dataset \cite{yang2013saliency}, which was proposed to n saliency detection dataset, composed of 5,168 complex images containing one or more salient objects; and \textbf{ICoSeg} \cite{batra2011interactively}, which is composed of 643 images, most of them containing multiple salient objects.
    
\begin{figure}[t!]
    \centering
    \begin{tabular}{c c}
             \includegraphics[width=0.18\textwidth]{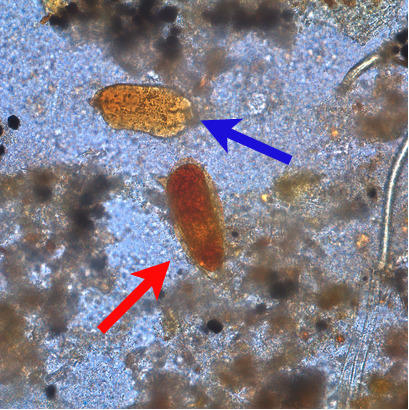} &
             \includegraphics[width=0.18\textwidth]{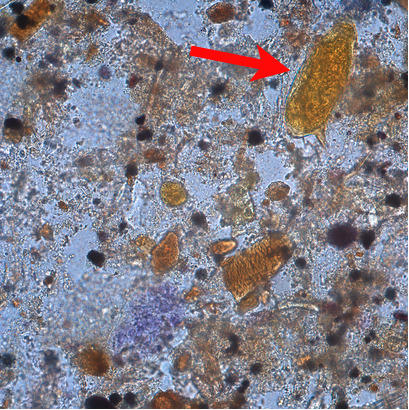} \\
             (a) & (b)
        \end{tabular}
    \caption{(a) A parasite egg (red arrow) and a fecal impurity (blue arrow) that shares similar characteristics to the eggs; (b) A heavily cluttered image with one parasite egg (red arrow)}
    
    \label{fig:parasite-example}
\end{figure}

    Additionally, we ran experiments on an in-house biomedical image dataset of intestinal \textbf{parasite eggs}. The dataset is composed of 630 images of \textit{schistosoma-mansoni} eggs obtained via TF-test \cite{}, with. The background is overloaded with fecal impurities that share similar characteristics to the eggs, posing a challenge to highlight the wanted objects alone (Figure \ref{fig:parasite-example}). Differently than the impurities, the eggs are elliptical and fall into a specific size range.
    
    Lastly, we used a \textbf{lung x-ray} dataset proposed in a Kaggle segmentation challenge to showcase that ITSELF can be extended to grayscale medical images. This dataset is composed of 704 images and contains normal and abnormal x-rays with manifestations of tuberculosis. It is required to attribute the data source to the National Library of Medicine, National Institutes of Health, Bethesda, MD, USA, and to the Shenzhen No.3 People's Hospital, Guangdong Medical College, Shenzhen, China. This dataset was made viable thanks to \cite{jaeger2013automatic, candemir2013lung}.

\subsection{Parameter tuning and Experimental setup}\label{subsec:parameter-tunning}
    For optimizing the methods' parameters we created subsets of size $N = min(\frac{\|D\|}{l10}, 100)$ where $|D|$ is the dataset size. We want to make sure that the user does not need many images to achieve satisfactory results, so $N$ limits the MSRA10K and DUT-OMRON training set size.
    
    Regarding the parameters, some parameters were fixed, and others changed for each dataset. The dataset-specific parameters values are grouped on Table \ref{tab:parameters}. The fixed parameter values will be presented as we list them.
    
    For superpixel segmentation, there are six parameters: the number of superpixels $n$; the number of foreground seeds $n_o = 3$; the number of OISF iterations over recomputed seeds $\hat{t}$, the superpixel size regularity ($\alpha = 0.8$), the border adherence weight ($\beta = 12$) and the saliency weight ($\gamma'$). On the saliency computation, there are two parameters: the query region importance $\psi$ and the saliency variance ($\sigma_s = 0.4$). For the prior integration step, there are also two parameters: the number of iterations $t' = 1$; and the updated strength $\Lambda$. Despite $t' = 1$, the automaton updates over the number of ITSELF iterations.
    
    We used two different color priors, one that highlights red/yellow colors and another to highlight color intensities. The intensity prior was used on all natural-image datasets and the x-ray dataset; however, on the x-ray dataset, we highlight darker intensities.  
    
    For prior estimation, there are eight parameters: the variance of each prior variance ($\sigma_i \| i \in (1..6)$ --- where $\sigma_3$ is related to the red/yellow prior and $\sigma_3'$ to intensities; and the size constrains for the ellipse prior $s_0 = 1500$, $s_1 = 5000$. Lastly, we run all the experiments using $i = 8$ full ITSELF iterations.
    
    Regarding SMD, they proposed a method to be used without any training step. In this regard, we used their available code without any modifications or parameter tuning. Note that SMD did a pre-training on the used datasets but are not clear regarding the size of their training split. As for DRFI, we used their available implementation and the same splits as we did for ITSELF.
    
    \begin{table}[ht]
    \resizebox{\columnwidth}{!}{%
    \begin{tabular}{|c|c|c|c|c|c|c|}
    \hline
       & ECSSD & DUT\_OMRON & ICOSEG & MSRA10K & Lungs & Parasites \\ \hline
    $\sigma_{1}$  & 0.2 & 0.2 & 0.2 & 0.2 & --- & --- \\ \hline 
    $\sigma_{2}$  & 0.2 & 0.5 & 0.5 & 0.5 & --- & 0.2 \\ \hline 
    $\sigma_{3}$  & 0.2 & --- & 0.2 & 0.8 & 0.5 & --- \\ \hline 
    $\sigma_{3}'$ & 0.2 & 0.5 & 0.8 & 0.8 & --- & 0.2 \\ \hline 
    $\sigma_{4}$  & 0.5 & 0.5 & 0.5 & 0.8 & 0.8 & --- \\ \hline 
    $\sigma_{5}$  & --- & --- & --- & --- & --- & 1.0 \\ \hline 
    $n$ & 200  & 200 & 200 & 200 & 200 & 500\\ \hline 
    $\gamma$ & 2.0 & 2.0 & 2.0 & 1.0 & 2.0 & 0.5\\ \hline 
    $\Lambda$ & 0.01 & 0.008 & 0.01 & 0.01 & 0.01 & 0.05 \\ \hline 
    $\psi'$ & 0.5 & 0.3 & 0.8 & 0.3 & 0.3 & 0.5\\ \hline 
    $\hat{t}'$ & 0.5 & 0.3 & 0.8 & 0.3 & 0.3 & 0.5\\ \hline 
    \end{tabular}%
    }
    \caption{A list of all parameters values that changed over the datasets.}
    \label{tab:parameters}
\end{table}

    The query selection strategies employed for each dataset were different from the first iteration of the framework. On further iterations, every dataset used the result of the past iteration to estimate foreground queries. For the natural-image datasets, the first framework iteration uses image-borders background queries to estimate a first saliency map and then uses the result to estimate foreground queries. For the parasites and x-ray datasets, the first iteration uses the combined prior map to estimate foreground queries.
    
    Lastly, when combining the multiple iteration's outputs, we observed that the first saliency estimation is often noisy; thus, we discard it.

\subsection{Evaluation Metrics}
    We used four traditional saliency metrics: weighted F-Measure (WF-Measure); weighted Precision (PRE$^\omega$); weighted Recall (REC$^\omega$); the mean-average error. Moreover, we propose using boundary recall to quantify this characteristic of the over-salient values of regions close to the object. By increasing the saliency of regions close to the object, the estimated objects, boundaries are moved away from the real object boundaries, reducing the BR (Figure \ref{fig:boundary-recall}). 
    The weighted F-Measure is the harmonic mean of PRE$^\omega$ and REC$^\omega$. The PRE$^\omega$ measures the exactness (\textit{i.e.}, whether non-salient regions were defined as salient) and REC$^\omega$ measures completeness (\textit{i.e.}, whether salient regions were defined as non-salient). These metrics were proposed to substitute the traditional binary-image-based precision and recall metrics, removing the need for computing the results on multiple threshold-segmented maps \cite{margolin2014evaluate}. Rather, the positive and negative ratios are computed based on the difference between a binary map and the saliency probability.
    
    The \emph{mean absolute error} (MAE) is the mean difference between the saliency map and the ground-truth. Even though the MAE compares the saliency map to the
    
    Having well-defined boundaries between object and background is particularly useful on tasks such as weakly-supervised semantic segmentation, where saliency maps may be used as estimates of a pixel-wise mask from an image-level annotation \cite{wei2016stc} to train more robust algorithms. We use the boundary recall (BR) over saliency maps thresholded by the mean saliency value. BR measures the percentage of match between the estimated object boundaries to the object boundaries in the ground-truth. We consider a boundary tolerance distance of two pixels, as proposed by Achanta \textit{et.al} \cite{achanta2012slic}.

\subsection{Natural-image dataset comparisons}
    As shown in Table \ref{tab:quantitative-results}, regarding the four traditional saliency metrics, ITSELF was ranked second on all but one dataset, with SMD getting the best scores. However, not only was SMD pre-trained with an unknown number of images, SMD often highlights non-salient regions close to salient ones.
    
    
   Comparing the boundary recall of the three methods, ITSELF and DRFI often alternate between first and second place, with SMD always on the bottom.

\begin{figure}[t!]
        \centering
        \begin{tabular}{c c}
                 \includegraphics[width=0.18\textwidth]{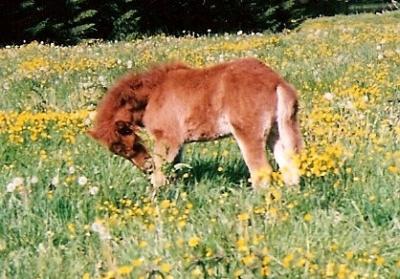} &
                 \includegraphics[width=0.18\textwidth]{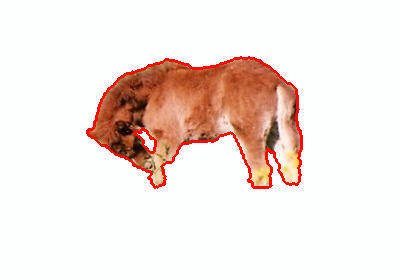} 
                 \\
                 (a) & (b) \\
                 \includegraphics[width=0.18\textwidth]{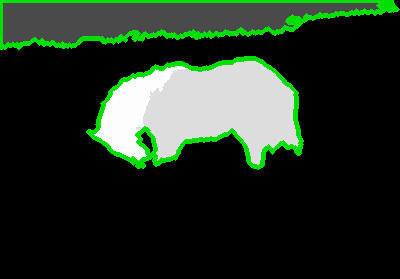} &
                 \includegraphics[width=0.18\textwidth]{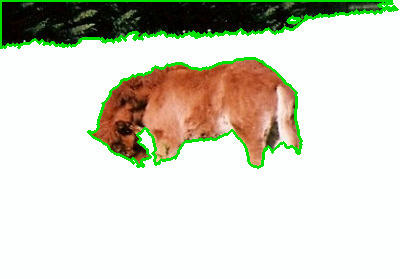} 
                 \\
                 (c) & (d) \\
                 \includegraphics[width=0.18\textwidth]{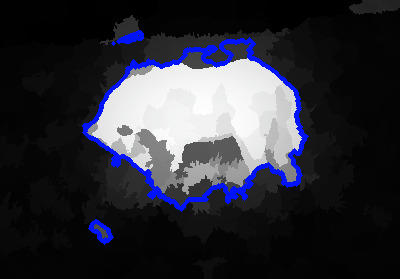} &
                 \includegraphics[width=0.18\textwidth]{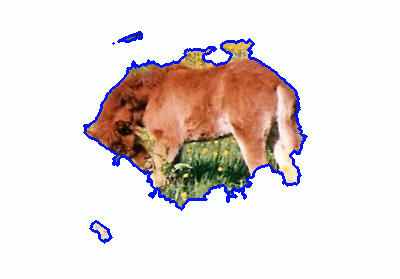} 
                 \\
                 (e) & (f) 
                 \end{tabular}
        \caption{(a) Input image. (b) Ground-truth segmentation; (c-f) ITSELF/SMD saliency maps with mean-saliency threshold  segmentation boundaries depicted on green/blue, respectively.}
        
        \label{fig:boundary-recall}
\end{figure}

\begin{figure*}[t!]
        \centering
        \begin{tabular}{c c c c c c c c c}
                 \includegraphics[width=0.09\textwidth]{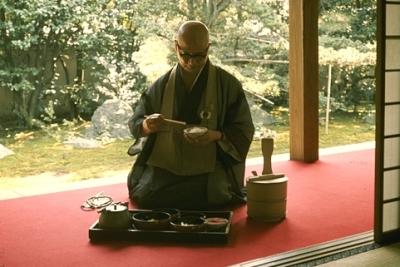} &
                 \includegraphics[width=0.09\textwidth]{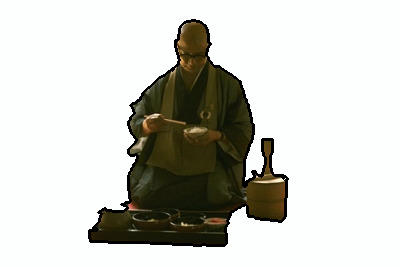} &
                 \includegraphics[width=0.09\textwidth]{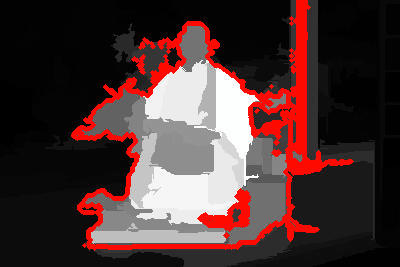} &
                 \includegraphics[width=0.09\textwidth]{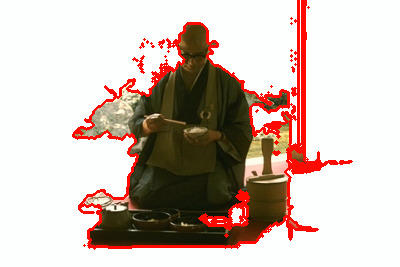} &
                 \includegraphics[width=0.09\textwidth]{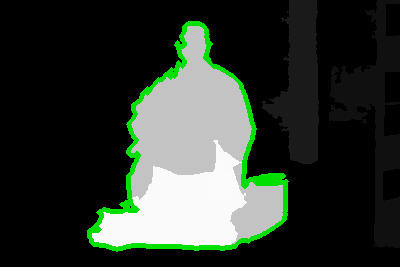} &
                 \includegraphics[width=0.09\textwidth]{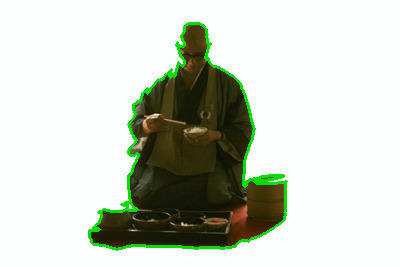} &
                 \includegraphics[width=0.09\textwidth]{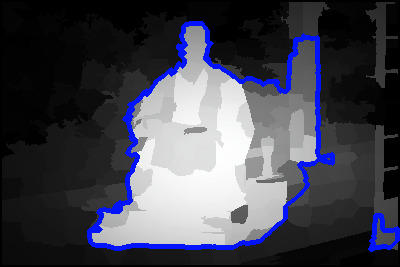} &
                 \includegraphics[width=0.09\textwidth]{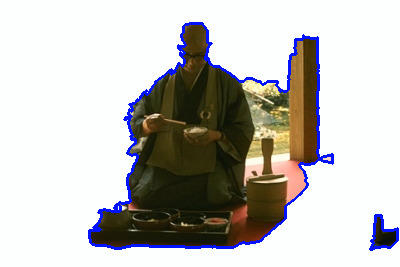} &
                 \\
                 \includegraphics[width=0.09\textwidth]{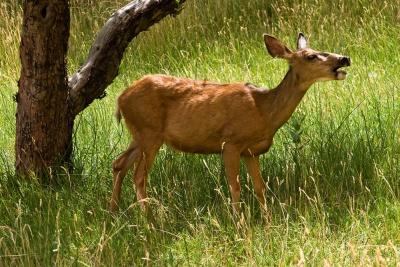} &
                 \includegraphics[width=0.09\textwidth]{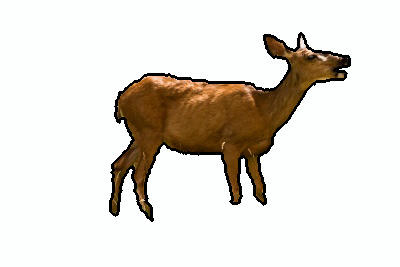} &
                 \includegraphics[width=0.09\textwidth]{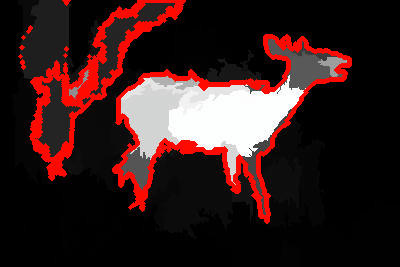} &
                 \includegraphics[width=0.09\textwidth]{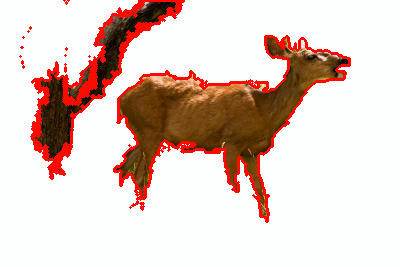} &
                 \includegraphics[width=0.09\textwidth]{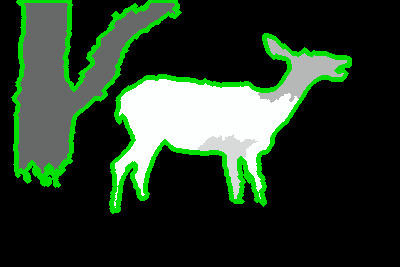} &
                 \includegraphics[width=0.09\textwidth]{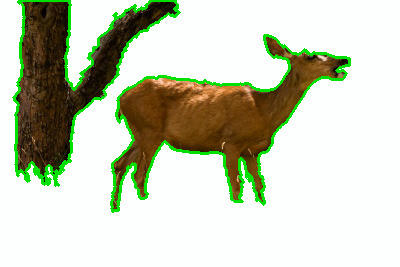} &
                 \includegraphics[width=0.09\textwidth]{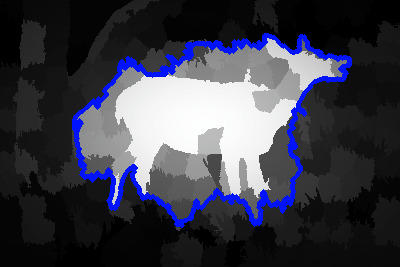} &
                 \includegraphics[width=0.09\textwidth]{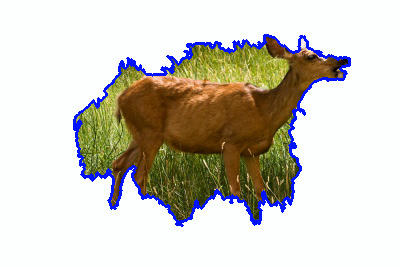} &
                 \\
                 \includegraphics[width=0.09\textwidth]{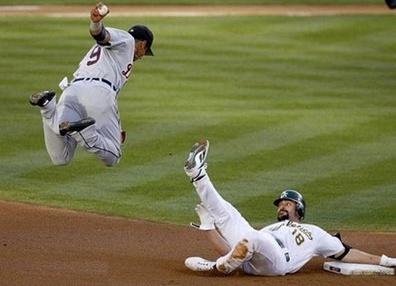} &
                 \includegraphics[width=0.09\textwidth]{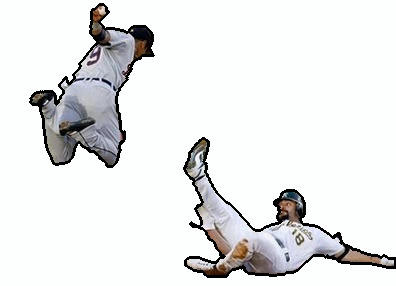} &
                 \includegraphics[width=0.09\textwidth]{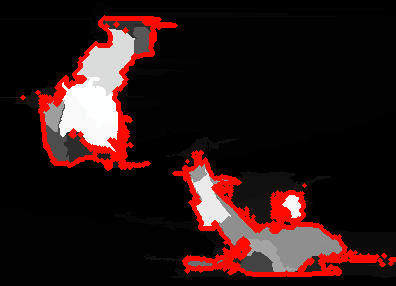} &
                 \includegraphics[width=0.09\textwidth]{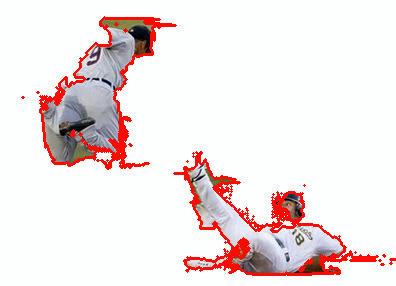} &
                 \includegraphics[width=0.09\textwidth]{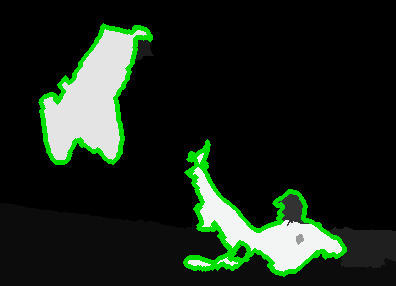} &
                 \includegraphics[width=0.09\textwidth]{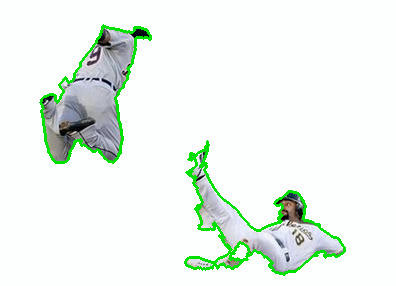} &
                 \includegraphics[width=0.09\textwidth]{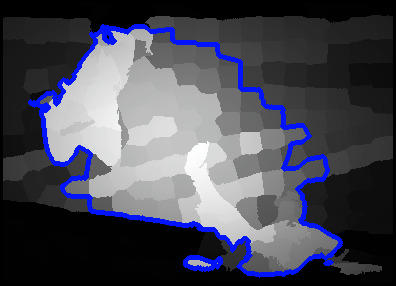} &
                 \includegraphics[width=0.09\textwidth]{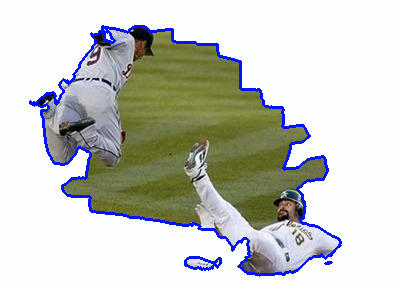} &
                 \\
                 \includegraphics[width=0.09\textwidth]{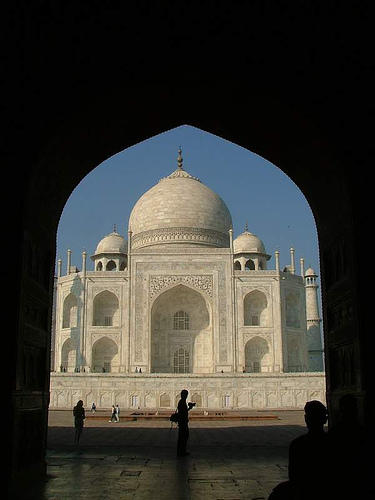} &
                 \includegraphics[width=0.09\textwidth]{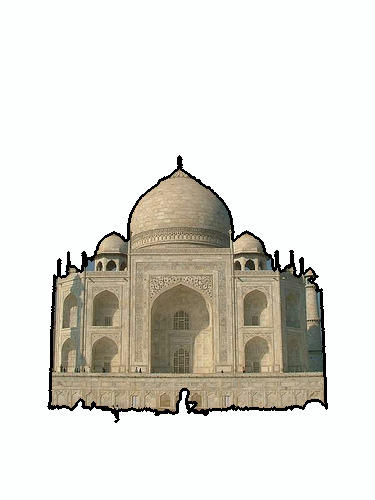} &
                 \includegraphics[width=0.09\textwidth]{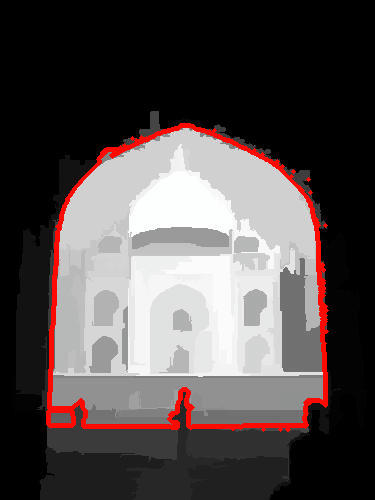} &
                 \includegraphics[width=0.09\textwidth]{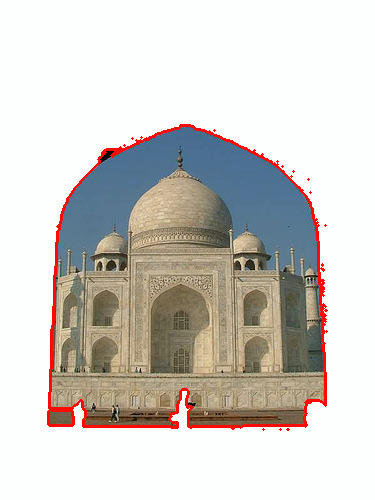} &
                 \includegraphics[width=0.09\textwidth]{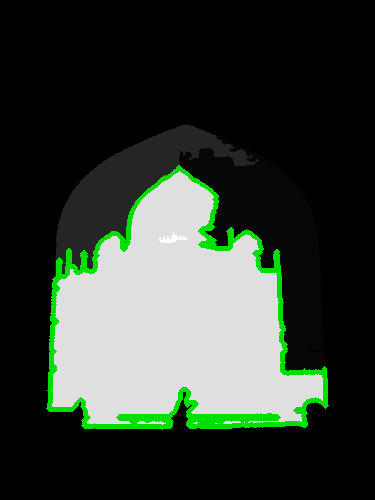} &
                 \includegraphics[width=0.09\textwidth]{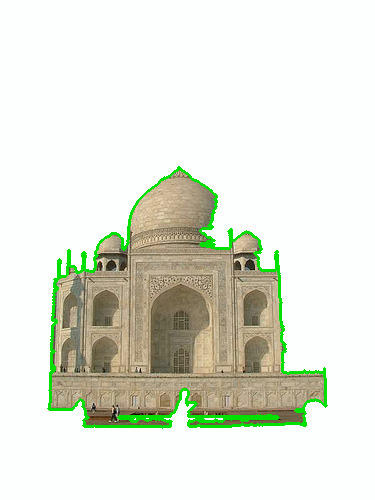} &
                 \includegraphics[width=0.09\textwidth]{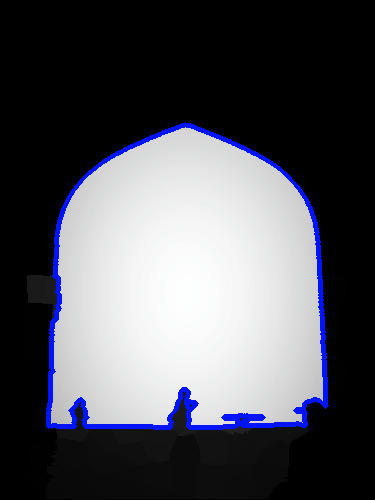} &
                 \includegraphics[width=0.09\textwidth]{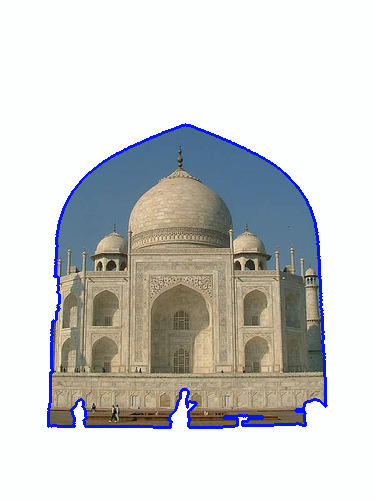} &
                 \\
                 \includegraphics[width=0.09\textwidth]{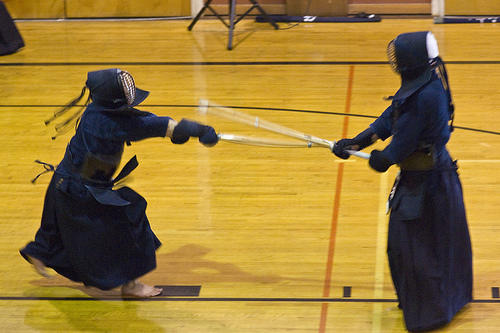} &
                 \includegraphics[width=0.09\textwidth]{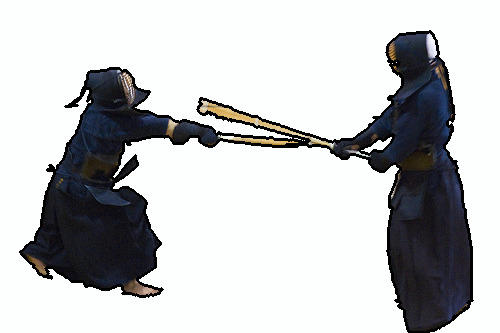} &
                 \includegraphics[width=0.09\textwidth]{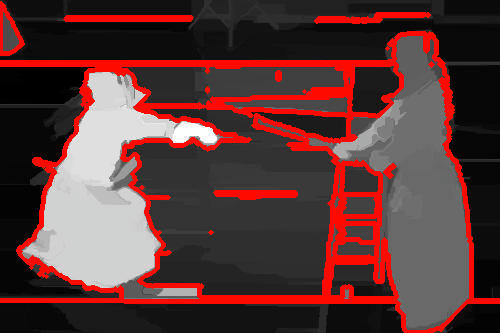} &
                 \includegraphics[width=0.09\textwidth]{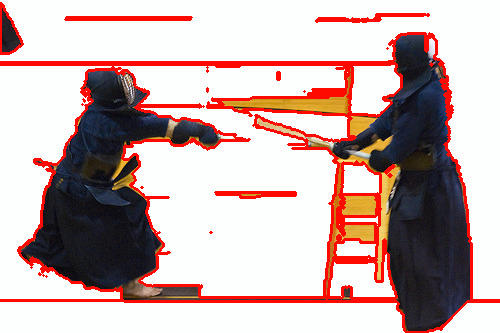} &
                 \includegraphics[width=0.09\textwidth]{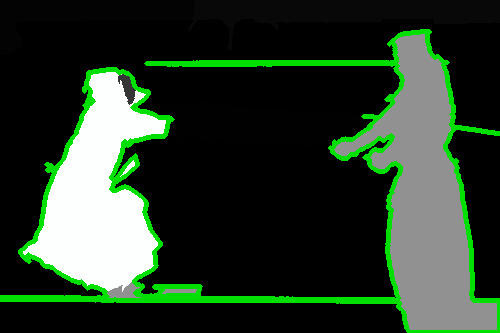} &
                 \includegraphics[width=0.09\textwidth]{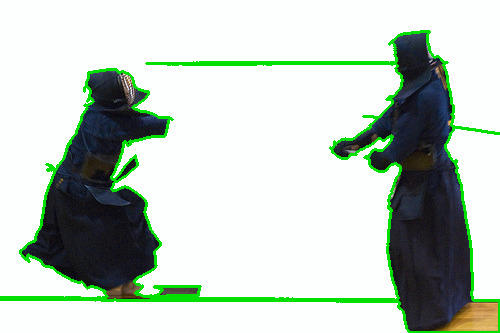} &
                 \includegraphics[width=0.09\textwidth]{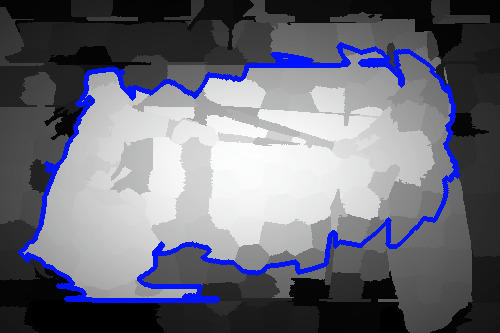} &
                 \includegraphics[width=0.09\textwidth]{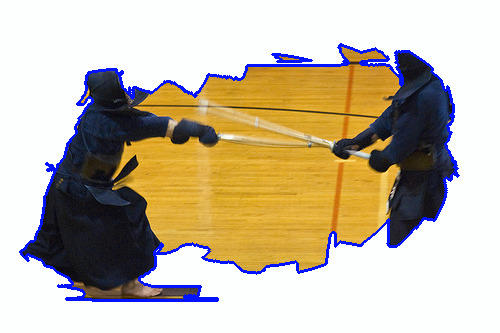} &
                 \\
                 \includegraphics[width=0.09\textwidth]{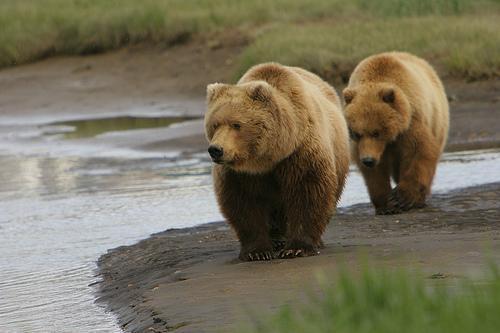} &
                 \includegraphics[width=0.09\textwidth]{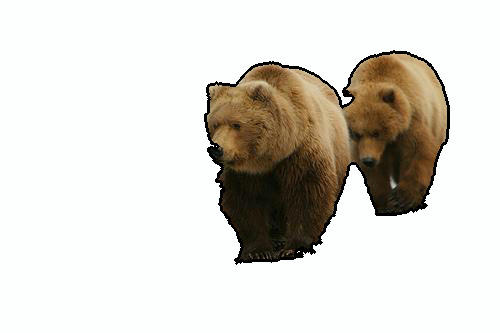} &
                 \includegraphics[width=0.09\textwidth]{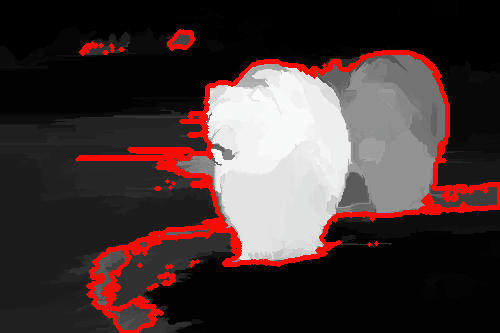} &
                 \includegraphics[width=0.09\textwidth]{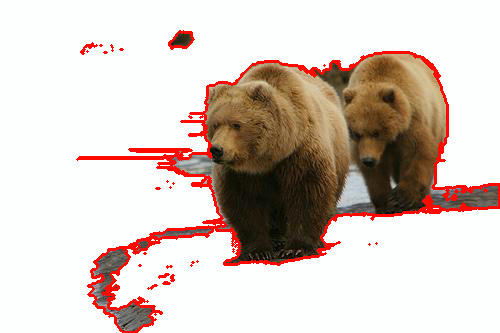} &
                 \includegraphics[width=0.09\textwidth]{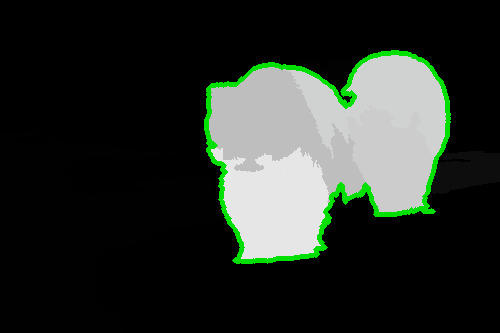} &
                 \includegraphics[width=0.09\textwidth]{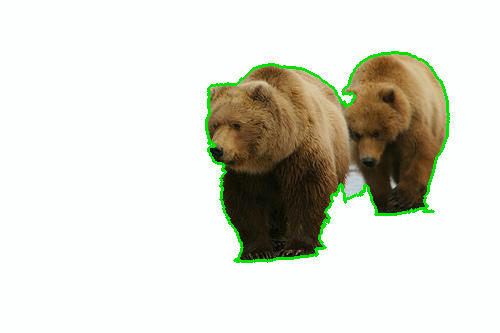} &
                 \includegraphics[width=0.09\textwidth]{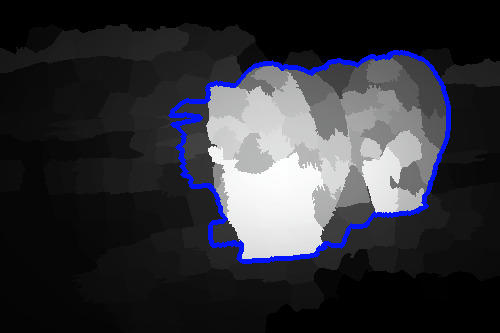} &
                 \includegraphics[width=0.09\textwidth]{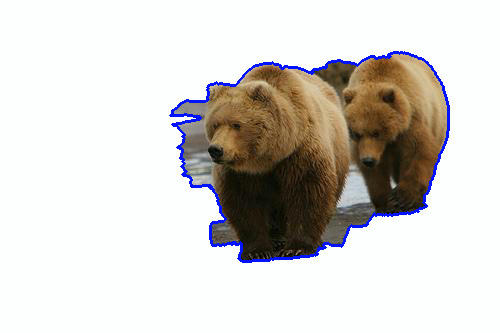} &
                 \\
                 \includegraphics[width=0.09\textwidth]{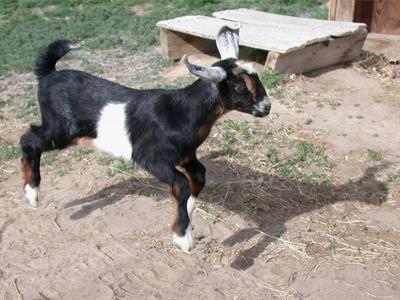} &
                 \includegraphics[width=0.09\textwidth]{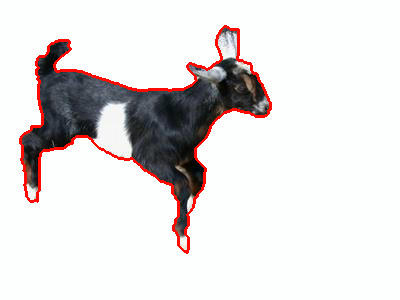} &
                 \includegraphics[width=0.09\textwidth]{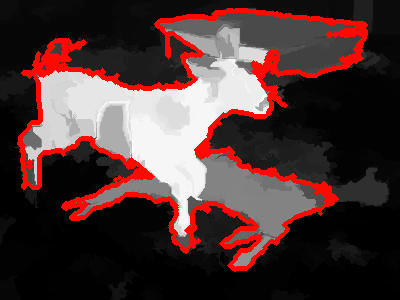} &
                 \includegraphics[width=0.09\textwidth]{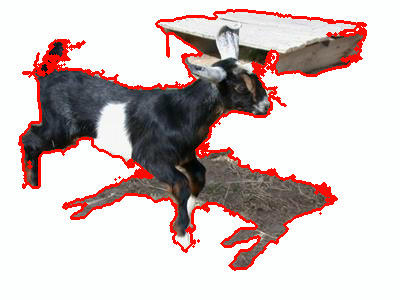} &
                 \includegraphics[width=0.09\textwidth]{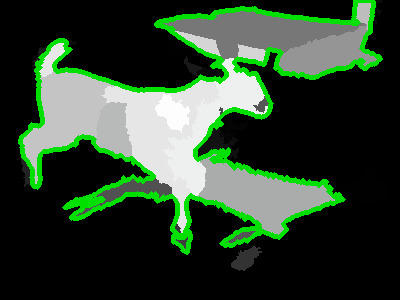} &
                 \includegraphics[width=0.09\textwidth]{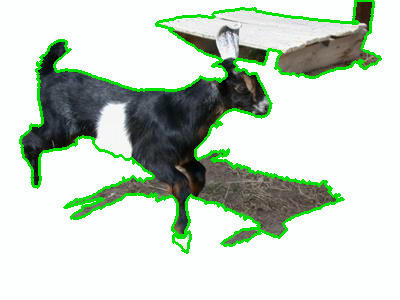} &
                 \includegraphics[width=0.09\textwidth]{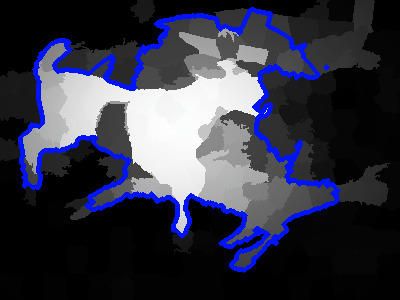} &
                 \includegraphics[width=0.09\textwidth]{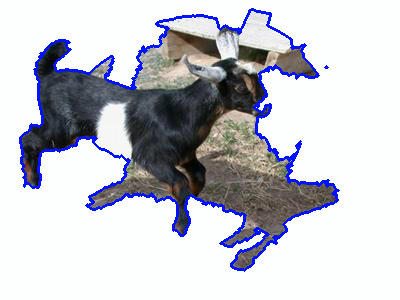} &
                 \\
                 \includegraphics[width=0.09\textwidth]{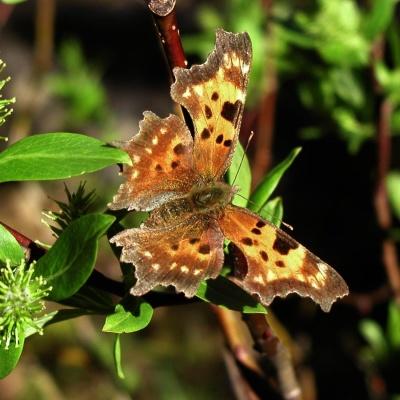} &
                 \includegraphics[width=0.09\textwidth]{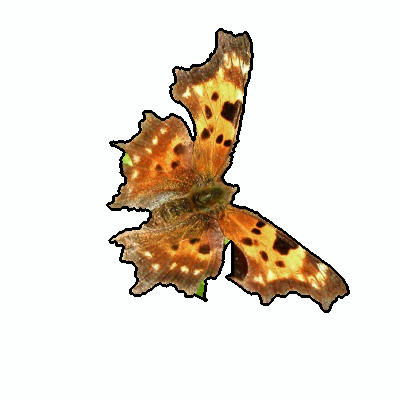} &
                 \includegraphics[width=0.09\textwidth]{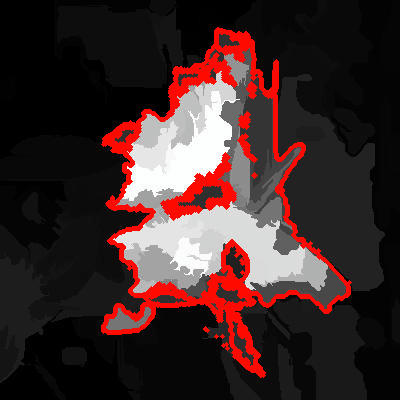} &
                 \includegraphics[width=0.09\textwidth]{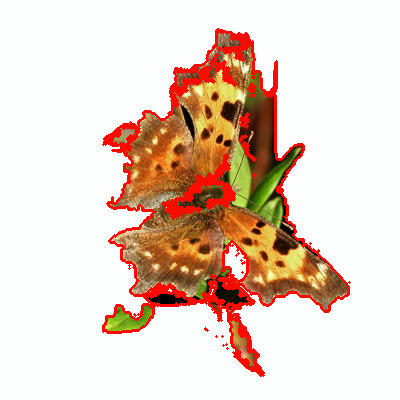} &
                 \includegraphics[width=0.09\textwidth]{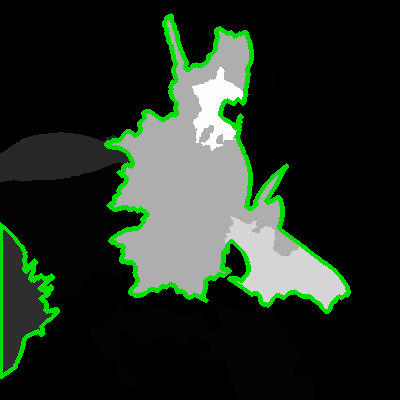} &
                 \includegraphics[width=0.09\textwidth]{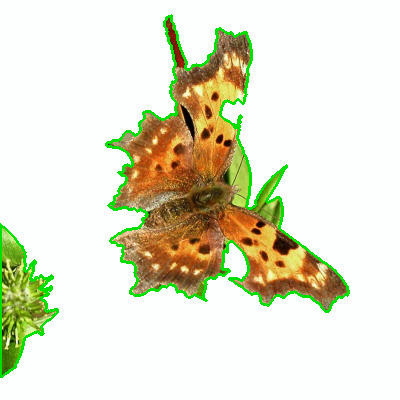} &
                 \includegraphics[width=0.09\textwidth]{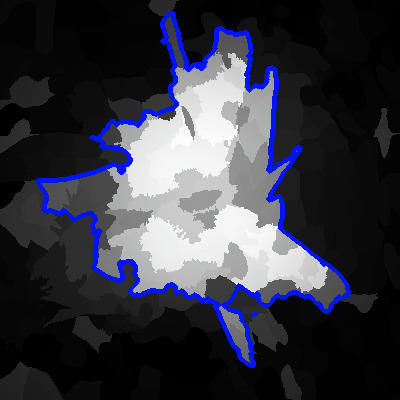} &
                 \includegraphics[width=0.09\textwidth]{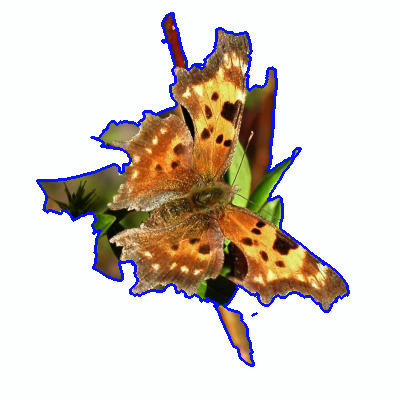} &
                 \\
                 \includegraphics[width=0.09\textwidth]{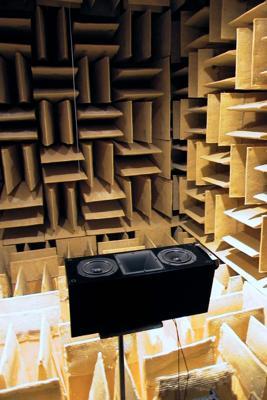} &
                 \includegraphics[width=0.09\textwidth]{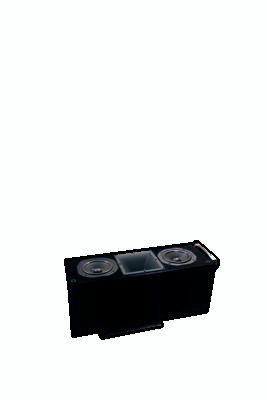} &
                 \includegraphics[width=0.09\textwidth]{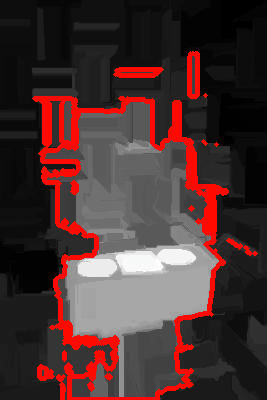} &
                 \includegraphics[width=0.09\textwidth]{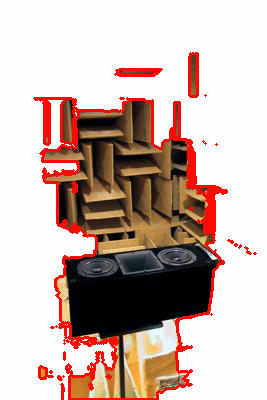} &
                 \includegraphics[width=0.09\textwidth]{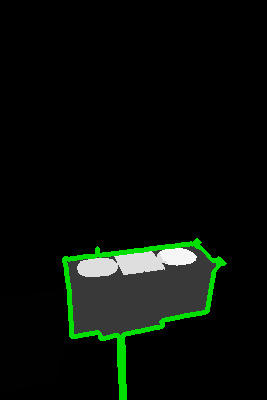} &
                 \includegraphics[width=0.09\textwidth]{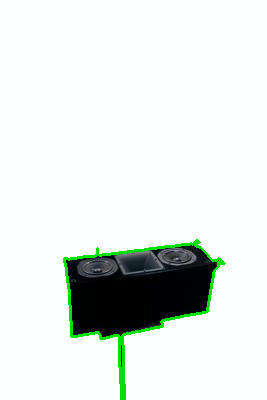} &
                 \includegraphics[width=0.09\textwidth]{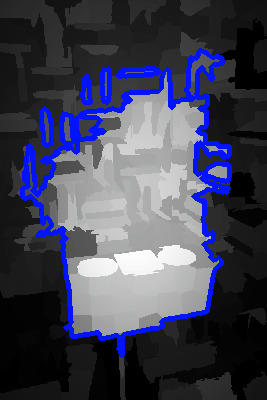} &
                 \includegraphics[width=0.09\textwidth]{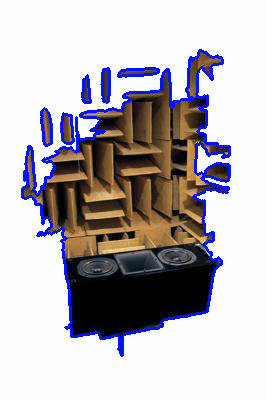} &
                 \\
                 \includegraphics[width=0.09\textwidth]{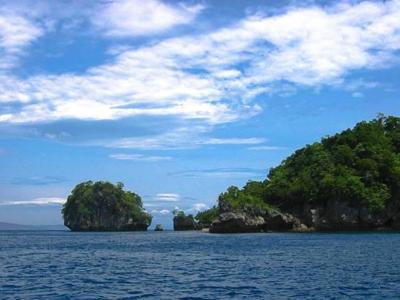} &
                 \includegraphics[width=0.09\textwidth]{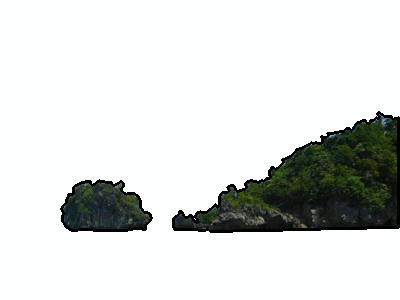} &
                 \includegraphics[width=0.09\textwidth]{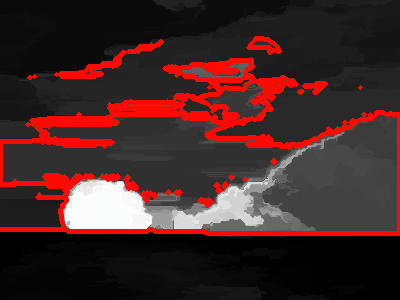} &
                 \includegraphics[width=0.09\textwidth]{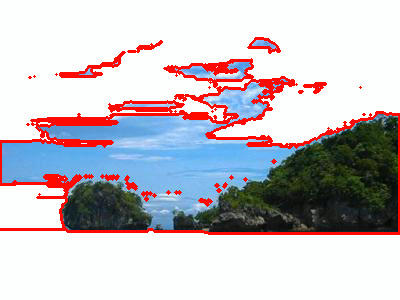} &
                 \includegraphics[width=0.09\textwidth]{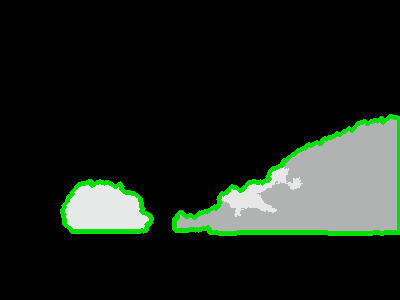} &
                 \includegraphics[width=0.09\textwidth]{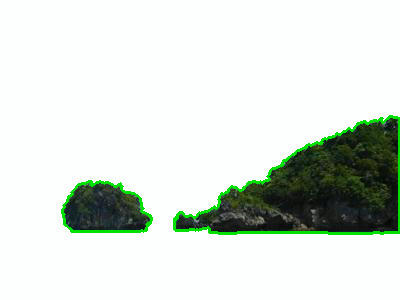} &
                 \includegraphics[width=0.09\textwidth]{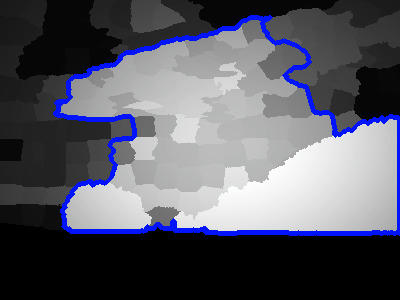} &
                 \includegraphics[width=0.09\textwidth]{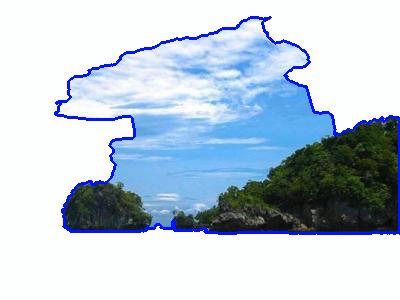} &
                 \\
                 \includegraphics[width=0.09\textwidth]{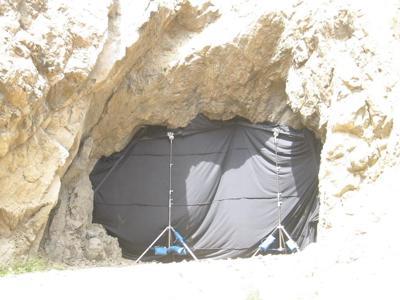} &
                 \includegraphics[width=0.09\textwidth]{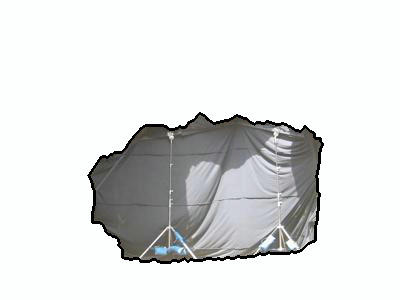} &
                 \includegraphics[width=0.09\textwidth]{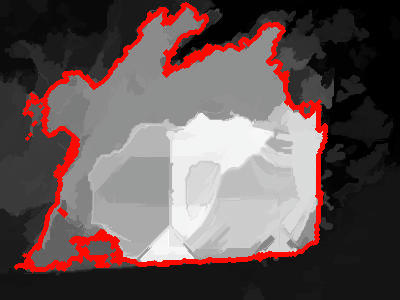} &
                 \includegraphics[width=0.09\textwidth]{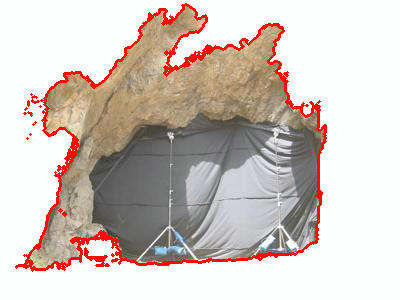} &
                 \includegraphics[width=0.09\textwidth]{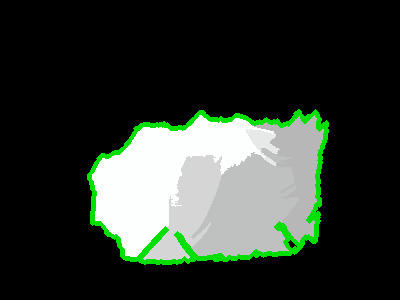} &
                 \includegraphics[width=0.09\textwidth]{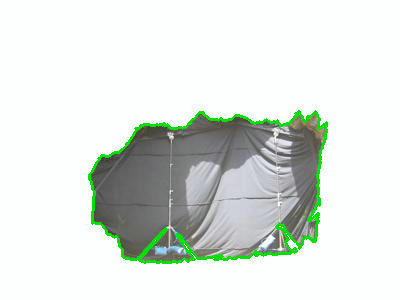} &
                 \includegraphics[width=0.09\textwidth]{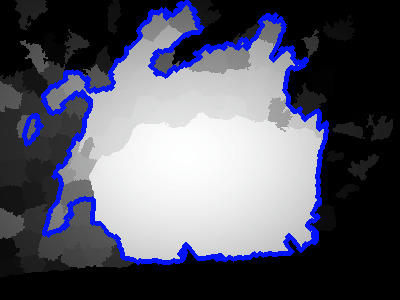} &
                 \includegraphics[width=0.09\textwidth]{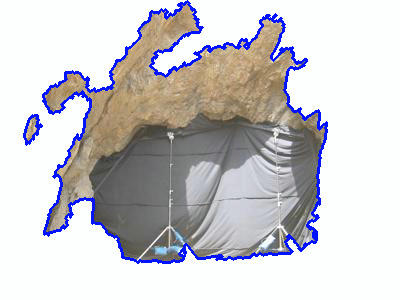} &
                 \\
                (a) & (b) & (c) & (d) & (e) & (f) & (g) & (h)
            \end{tabular}
        \caption{(a) Input image. (b) Ground-truth segmentation; (c-h) DRFI/ITSELF/SMD saliency maps with mean-saliency threshold  segmentation boundaries depicted on red/green/blue, respectively. Note how ITSELF tend to create more accentuated contrast between the object and background, adhering to the boundaries.  }
        
        \label{fig:segmentation_results}
\end{figure*}

    The saliency/superpixel loop provides a final saliency estimation with more semantic meaning than previous methods. Even though ITSELF was not completely accurate given the ground-truth in Figure \ref{fig:boundary-recall}, the wrongfully salient regions are highly different from most of the background or highly similar to the foreground. The WF-Measure of ITSELF and SMD are, respectively, $0.689$ and $0.781$. Nevertheless, SMD increases the saliency of background regions close to the horse. ITSELF does have a big non-salient region segmented on the top of the image; however, this region does not share similar characteristics to most of the background, unlike SMD's miss-estimation. More examples can be seen on Figure \ref{fig:segmentation_results}.
    
\begin{figure}[t!]
    \centering
    \begin{tabular}{c c c c}
             \includegraphics[width=0.09\textwidth]{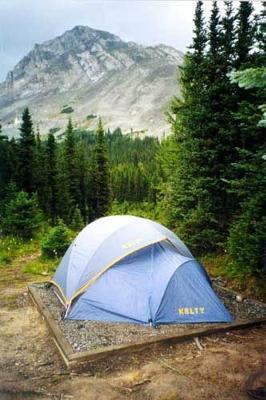} &
             \includegraphics[width=0.09\textwidth]{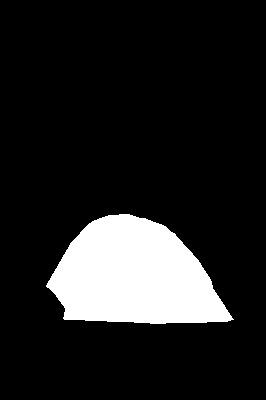} &
             \includegraphics[width=0.09\textwidth]{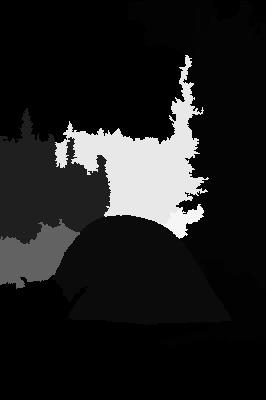} &
             \includegraphics[width=0.09\textwidth]{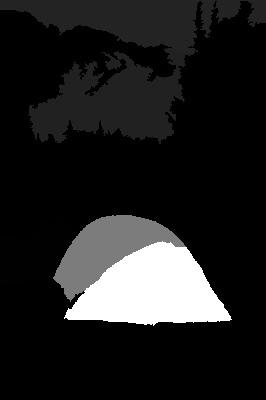} \\
            (a) & (b) & (c) & (d)
        \end{tabular}
    \caption{(a) Original image. (b) Superpixel Segmentation; (c) Reported ITSELF result; (d) Improved result by removing the center and focus priors.}
    
    \label{fig:bad-priors-example}
    \end{figure}

    These datasets are composed of profoundly different objects, and using priors assumptions does make the model prone to error on images that these assumptions fail to describe important salient characteristics (Figure \ref{fig:bad-priors-example}).
    
\begin{figure}[t!]
    \centering
    \begin{tabular}{c c c}
             \includegraphics[width=0.13\textwidth]{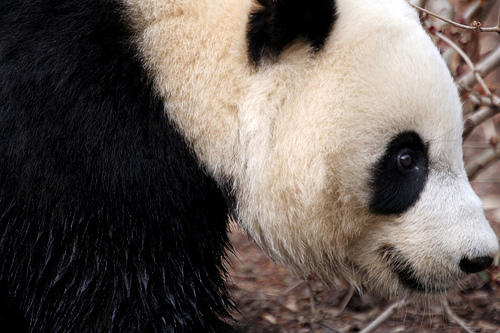} &
             \includegraphics[width=0.13\textwidth]{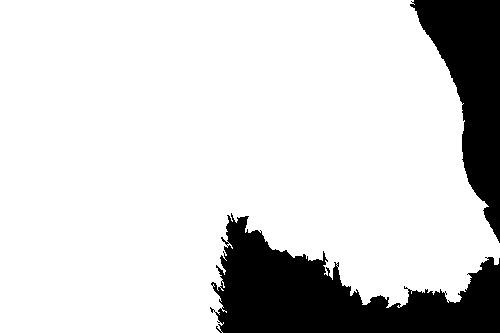} &
             \includegraphics[width=0.13\textwidth]{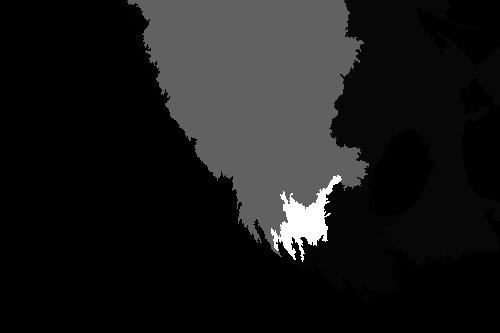}
        \end{tabular}
        \begin{tabular}{c c c}
             \includegraphics[width=0.13\textwidth]{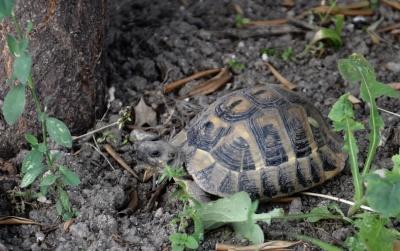} &
             \includegraphics[width=0.13\textwidth]{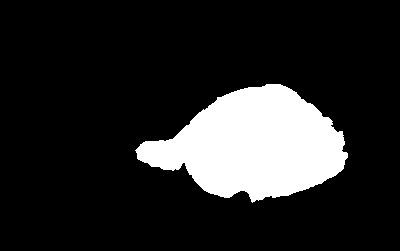} &
             \includegraphics[width=0.13\textwidth]{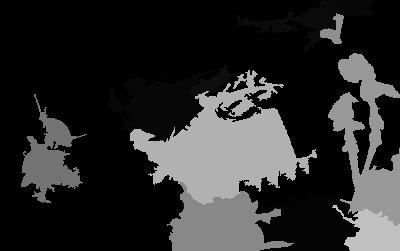} \\
            (a) & (b) & (c)
        \end{tabular}
    \caption{(a) Original image. (b) Ground-truth; (c) ITSELF saliency map. On the second image, note how there are contrasting green parts on the image background.}
    
    \label{fig:itself-fail}
    \end{figure}

    Another fail case scenario is when the salient object is too large or when they are too similar to most of the background (Figure \ref{fig:itself-fail}).
    
\begin{table}[ht]
\resizebox{\columnwidth}{!}{%
\renewcommand{\arraystretch}{2}
\begin{tabular}{|c|c|c|c|c|c|c|}
\hline
\multirow{4}{*}{\begin{turn}{90}ECSSD\end{turn}} 
& Methods & WF-Measure & BR & MAE & PRE$^\omega$ & REC$^\omega$ \\ \cline{2-7}
& \textbf{ITSELF} & 0.509 & {\color[HTML]{0000FE} 0.473} &  {\color[HTML]{0000FE}0.177} & {\color[HTML]{0000FE}0.601} & 0.491 \\ \cline{2-7}
& SMD &  {\color[HTML]{0000FE}0.517} &  0.425 &  0.227 & 0.543 & {\color[HTML]{FE0000}0.660} \\ \cline{2-7}
& DRFI & {\color[HTML]{FE0000} 0.547} & {\color[HTML]{FE0000} 0.556} & {\color[HTML]{FE0000}0.159} & {\color[HTML]{FE0000}0.606} & {\color[HTML]{0000FE}0.564} \\ \hline\hline

\multirow{4}{*}{\begin{turn}{90}DUT OMRON\end{turn}} 
& Methods & WF-Measure & BR & MAE & PRE$^\omega$ & REC$^\omega$  \\ \cline{2-7}
& \textbf{ITSELF} & {\color[HTML]{0000FE}0.406} & {\color[HTML]{FE0000}0.436} & {\color[HTML]{0000FE}0.144} & {\color[HTML]{FE0000}0.416} & 0.540 \\ \cline{2-7}
& SMD & {\color[HTML]{FE0000} 0.424} & 0.353 & {\color[HTML]{FE0000} 0.136} & {\color[HTML]{0000FE}0.387} & {\color[HTML]{0000FE}0.659}\\ \cline{2-7}
& DRFI & 0.357 & {\color[HTML]{0000FE} 0.356} & 0.193 & 0.290 & {\color[HTML]{FE0000}0.679}\\ \hline\hline

\multirow{4}{*}{\begin{turn}{90}ICOSEG\end{turn}} 
& Methods & WF-Measure & BR & MAE & PRE$^\omega$ & REC$^\omega$ \\ \cline{2-7}
& \textbf{ITSELF} & {\color[HTML]{0000FE}0.580} & {\color[HTML]{0000FE}0.571} & {\color[HTML]{0000FE}0.149} & {\color[HTML]{0000FE}0.676} & 0.618\\ \cline{2-7}
& SMD    & {\color[HTML]{FE0000} 0.611} & 0.527 & {\color[HTML]{FE0000} 0.138} & {\color[HTML]{FE0000}0.696} & {\color[HTML]{FE0000}0.656} \\ \cline{2-7}
& DRFI  & {\color[HTML]{333333} 0.547} & {\color[HTML]{FE0000} 0.591} & 0.152 & 0.657 & {\color[HTML]{0000FE}0.635} \\ \hline\hline

\multirow{4}{*}{\begin{turn}{90}MSRA10K\end{turn}} 
& Methods & WF-Measure & BR & MAE & PRE$^\omega$ & REC$^\omega$  \\ \cline{2-7}
& \textbf{ITSELF} & {\color[HTML]{0000FE}0.675} & {\color[HTML]{FE0000}0.634} & {\color[HTML]{0000FE}0.116} & {\color[HTML]{0000FE}0.724} & 0.680\\ \cline{2-7}
& SMD    & {\color[HTML]{FE0000} 0.704} & 0.594 & {\color[HTML]{FE0000} 0.104} & {\color[HTML]{FE0000}0.730} & {\color[HTML]{FE0000}0.733} \\ \cline{2-7}
& DRFI   &  0.583 & {\color[HTML]{0000FE} 0.435} & 0.149 & 0.525 & {\color[HTML]{0000FE}0.724} \\ \hline \hline

\multirow{4}{*}{\begin{turn}{90}Lungs\end{turn}} 
& Methods & WF-Measure & BR & MAE & PRE$^\omega$ & REC$^\omega$  \\ \cline{2-7}
& \textbf{ITSELF} & {\color[HTML]{FE0000} 0.621}&  {\color[HTML]{0000FE}0.208} & {\color[HTML]{FE0000} 0.141} & {\color[HTML]{FE0000}0.857} & 0.506 \\ \cline{2-7}
& SMD    & {\color[HTML]{0000FE}0.404} & 0.095 & {\color[HTML]{0000FE}0.325} & {\color[HTML]{0000FE}0.294} & {\color[HTML]{FE0000}0.724}  \\ \cline{2-7}
& DRFI   & 0.325 & {\color[HTML]{FE0000}0.255} & 0.412 & 0.224 & {\color[HTML]{0000FE}0.664} \\ \hline \hline

\multirow{4}{*}{\begin{turn}{90}Parasites\end{turn}} 
& Methods & WF-Measure & BR & MAE & PRE$^\omega$ & REC$^\omega$ \\ \cline{2-7}
& \textbf{ITSELF} & {\color[HTML]{FE0000} 0.538} & {\color[HTML]{FE0000}0.382} & {\color[HTML]{FE0000} 0.013} & {\color[HTML]{FE0000}0.490} & {\color[HTML]{FE0000}0.683} \\ \cline{2-7}
& SMD    & {\color[HTML]{0000FE}0.121} & 0.192 & {\color[HTML]{0000FE}0.155} & {\color[HTML]{0000FE}0.078} & {\color[HTML]{0000FE}0.662} \\ \cline{2-7}
& DRFI   & 0.041 & {\color[HTML]{0000FE}0.320} & 0.164 & 0.022 & 0.433 \\ \hline
\end{tabular}%
}
\vspace{0.2cm}
\caption{The best scores are colored in red and blue, respectively.}
\label{tab:quantitative-results}
\end{table}

\subsection{Non-natural-image dataset comparisons}
\begin{figure*}[t!]
        \centering
        \begin{tabular}{c c c c c c c c c}
                 \includegraphics[width=0.09\textwidth]{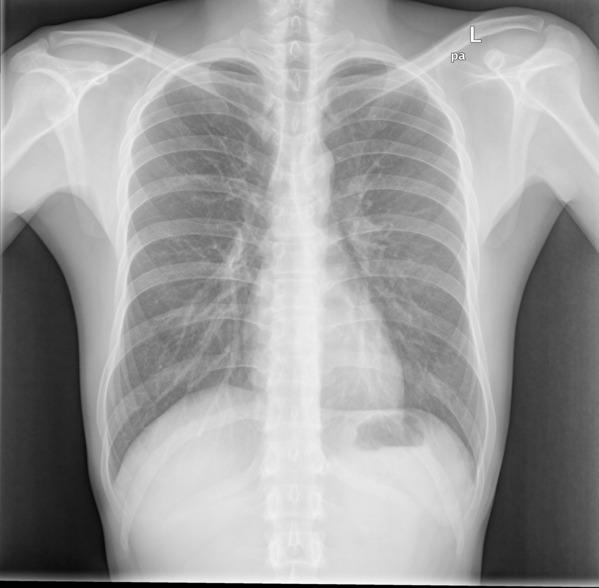} &
                 \includegraphics[width=0.09\textwidth]{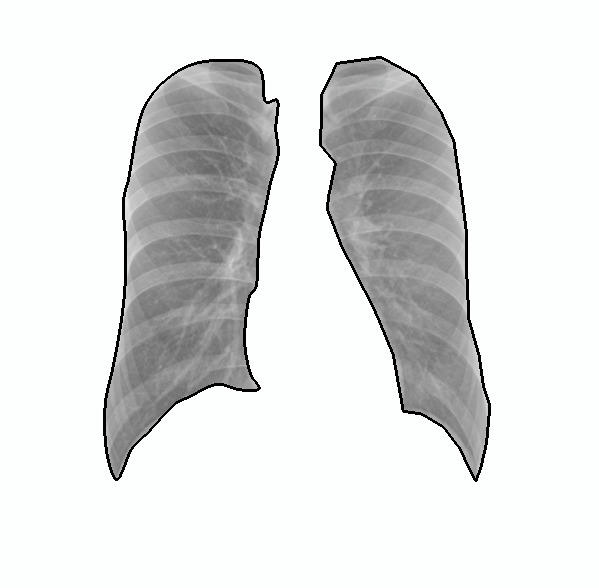} &
                 \includegraphics[width=0.09\textwidth]{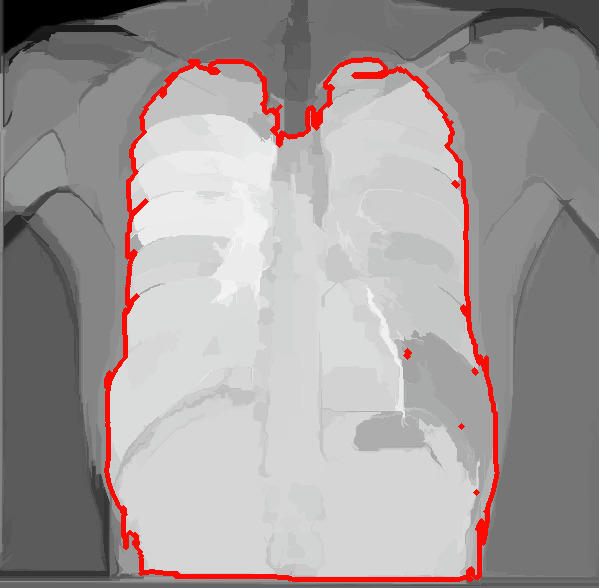} &
                 \includegraphics[width=0.09\textwidth]{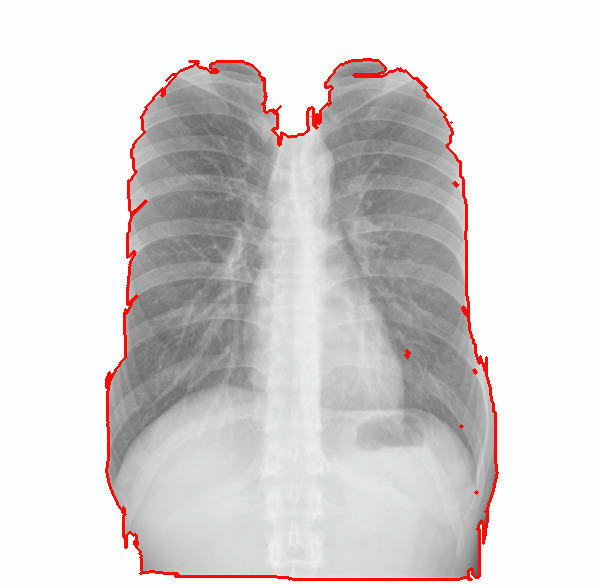} &
                 \includegraphics[width=0.09\textwidth]{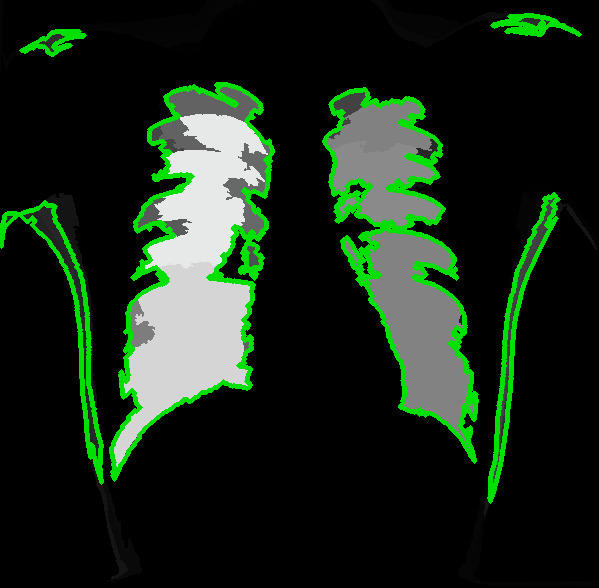} &
                 \includegraphics[width=0.09\textwidth]{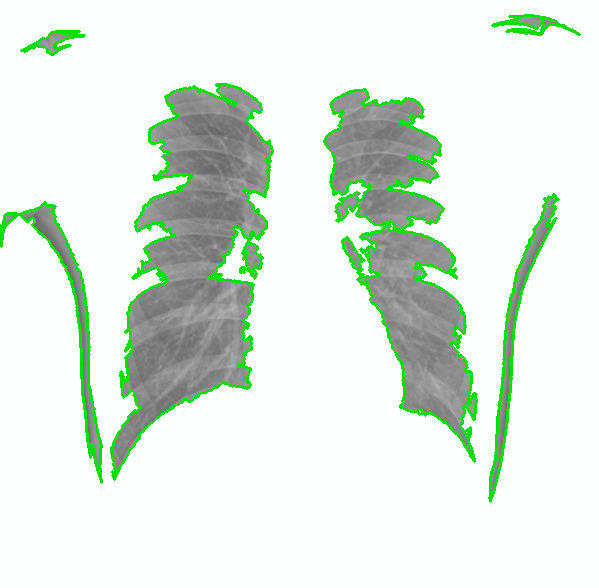} &
                 \includegraphics[width=0.09\textwidth]{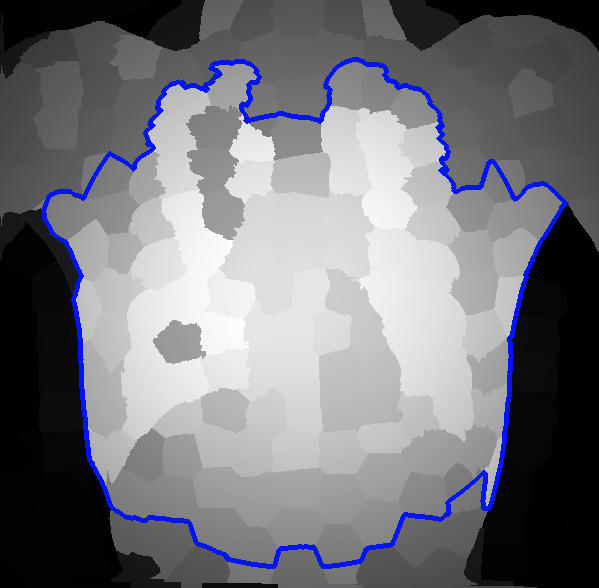} &
                 \includegraphics[width=0.09\textwidth]{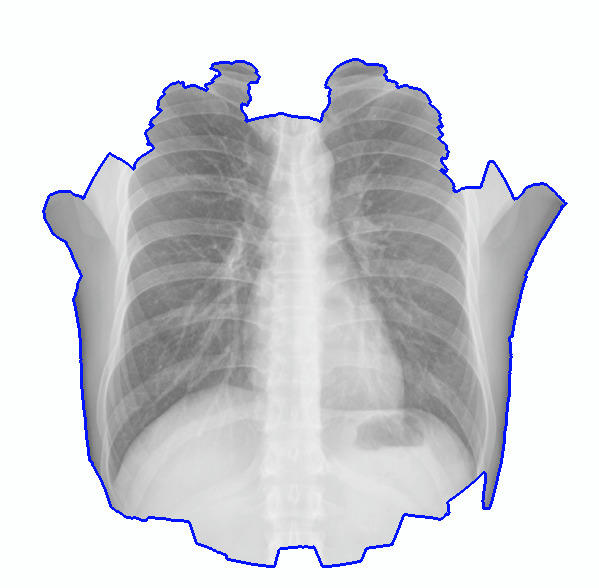} &
                 \\
                 \includegraphics[width=0.09\textwidth]{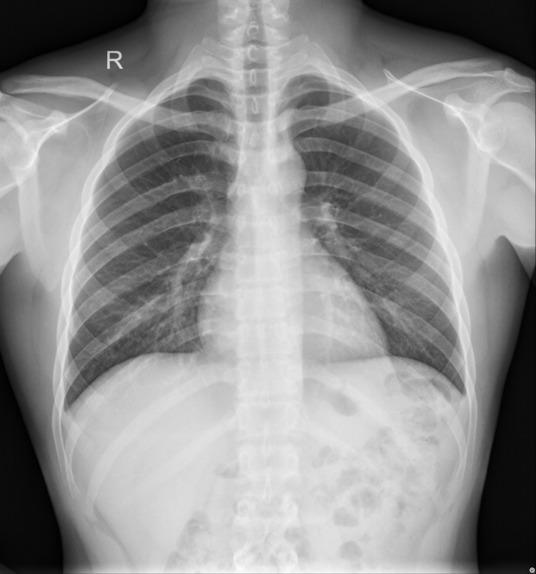} &
                 \includegraphics[width=0.09\textwidth]{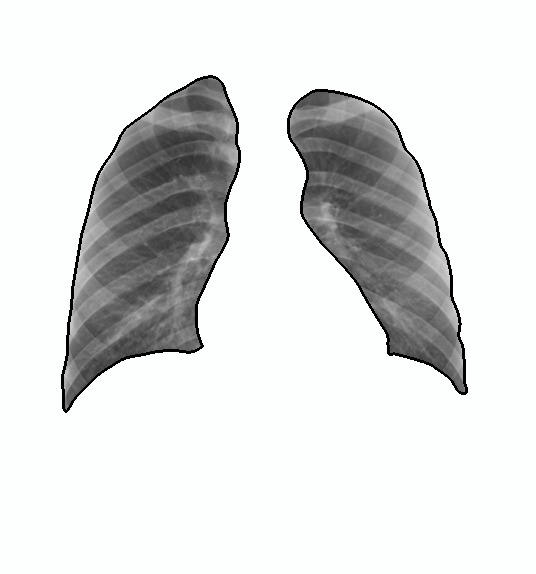} &
                 \includegraphics[width=0.09\textwidth]{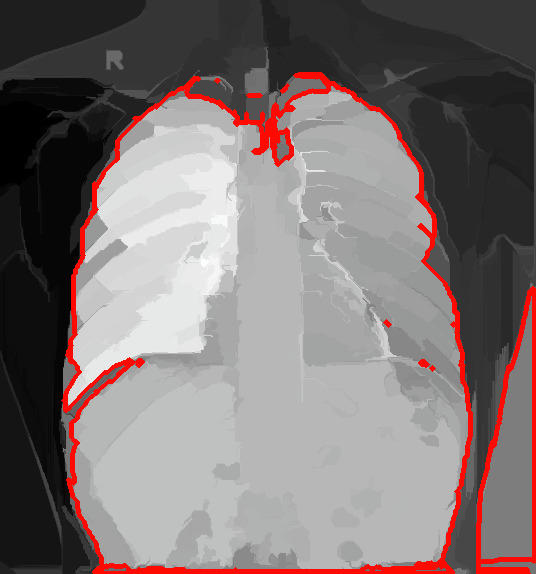} &
                 \includegraphics[width=0.09\textwidth]{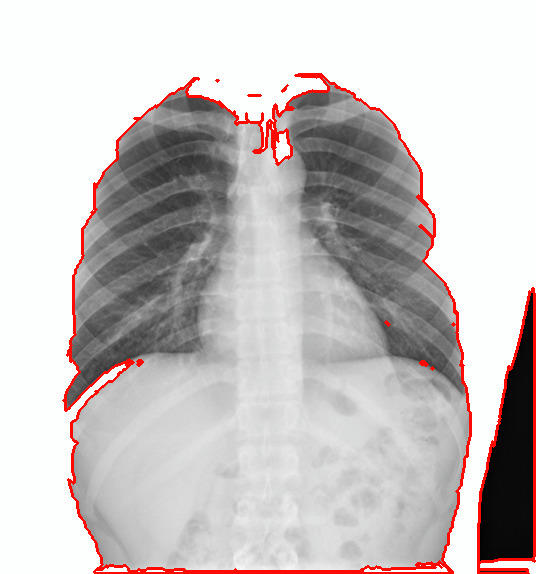} &
                 \includegraphics[width=0.09\textwidth]{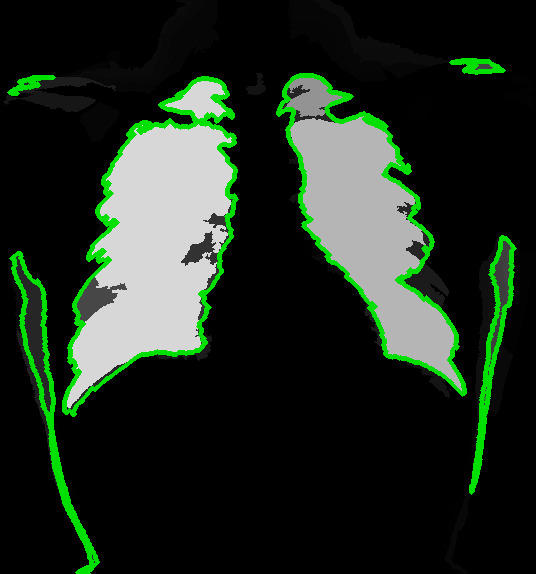} &
                 \includegraphics[width=0.09\textwidth]{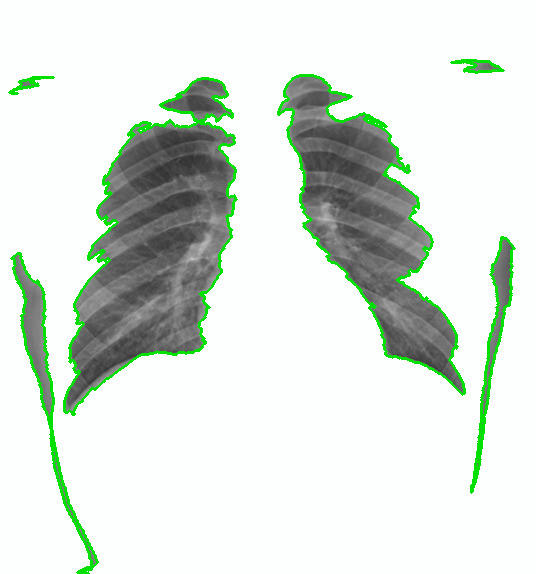} &
                 \includegraphics[width=0.09\textwidth]{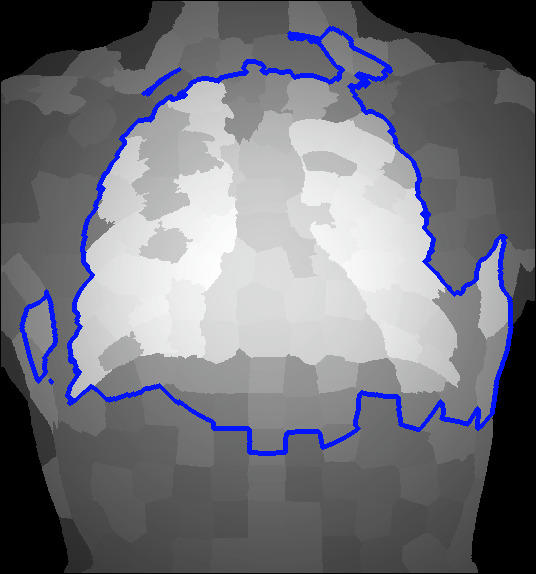} &
                 \includegraphics[width=0.09\textwidth]{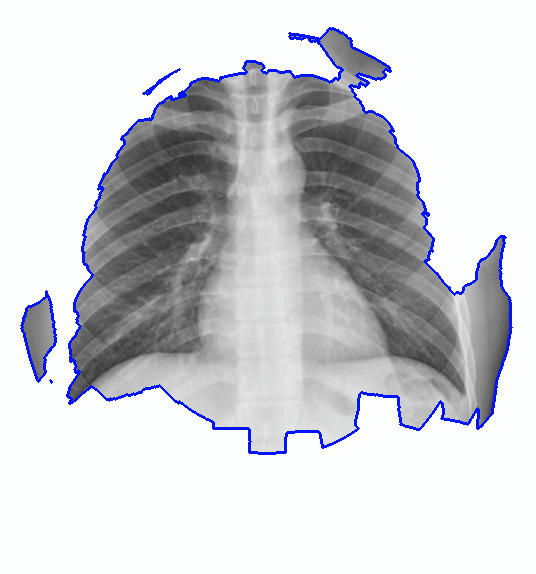} &
                 \\
                 \includegraphics[width=0.09\textwidth]{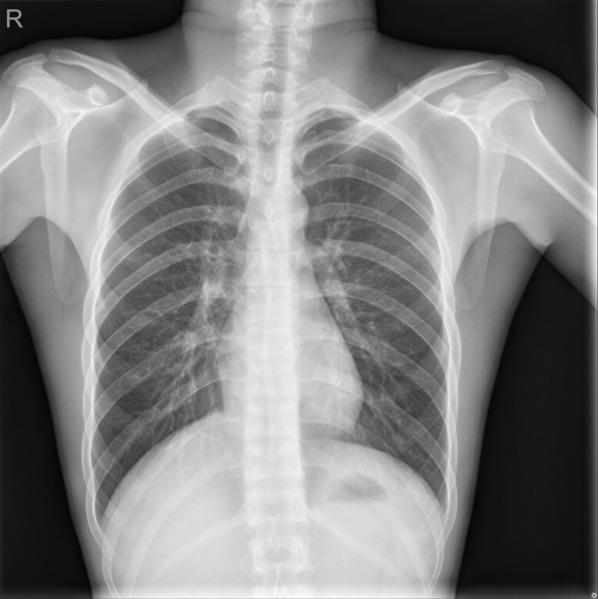} &
                 \includegraphics[width=0.09\textwidth]{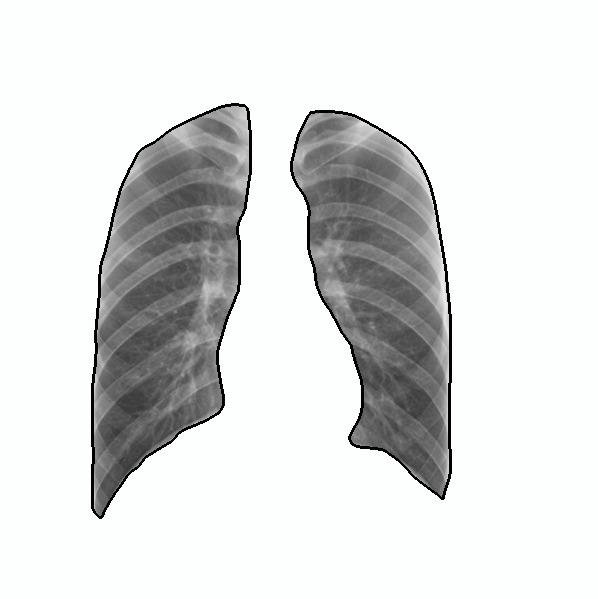} &
                 \includegraphics[width=0.09\textwidth]{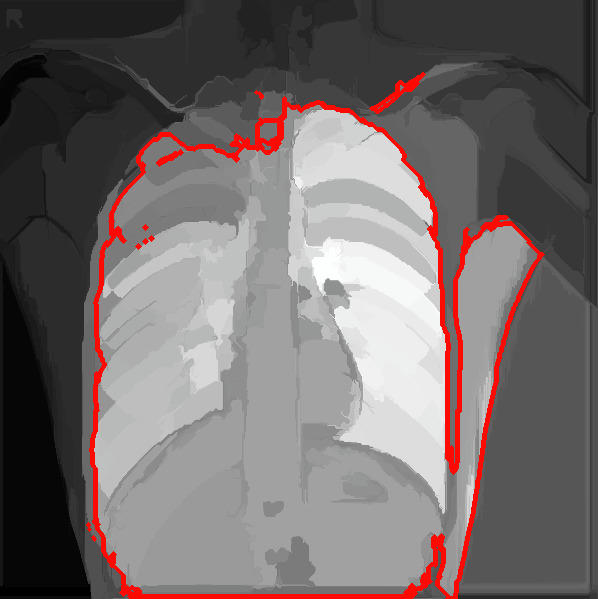} &
                 \includegraphics[width=0.09\textwidth]{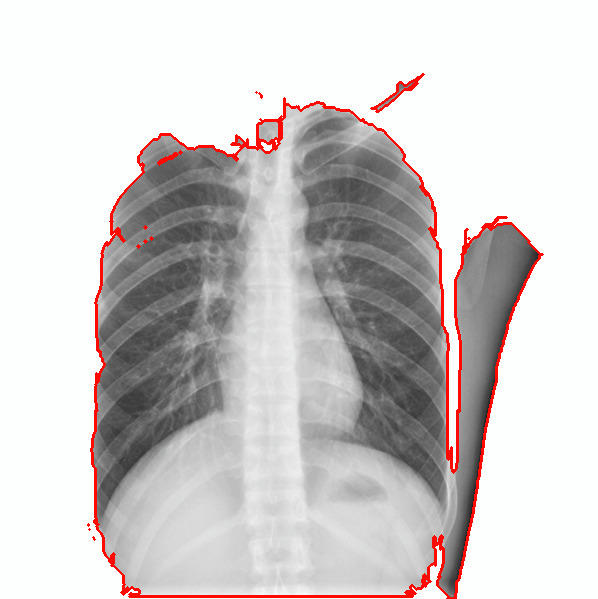} &
                 \includegraphics[width=0.09\textwidth]{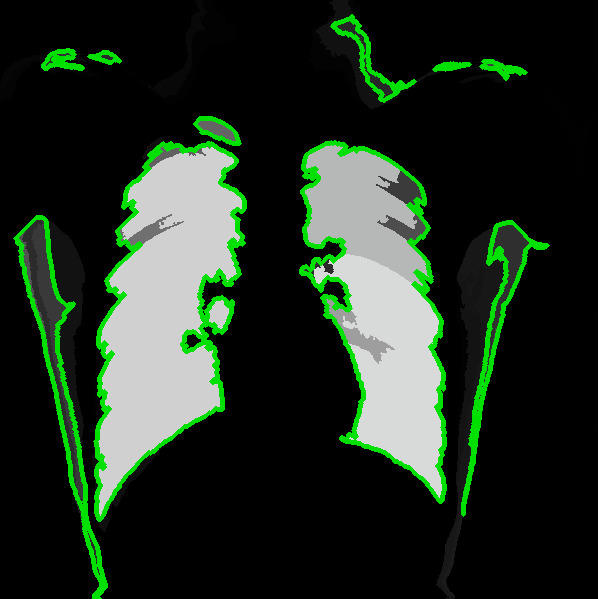} &
                 \includegraphics[width=0.09\textwidth]{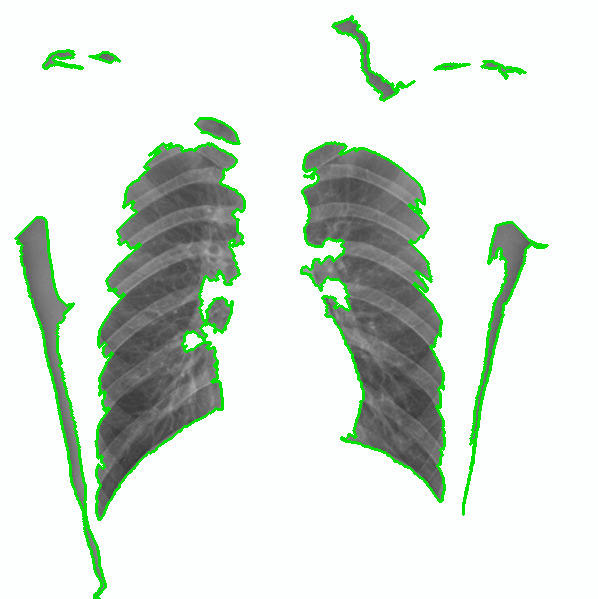} &
                 \includegraphics[width=0.09\textwidth]{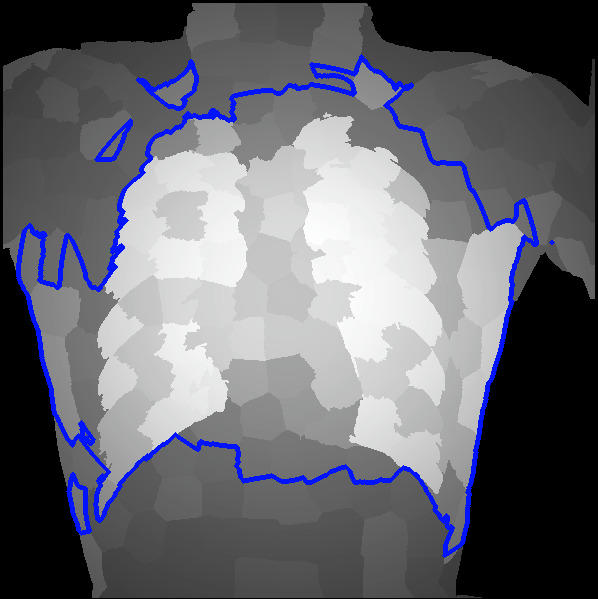} &
                 \includegraphics[width=0.09\textwidth]{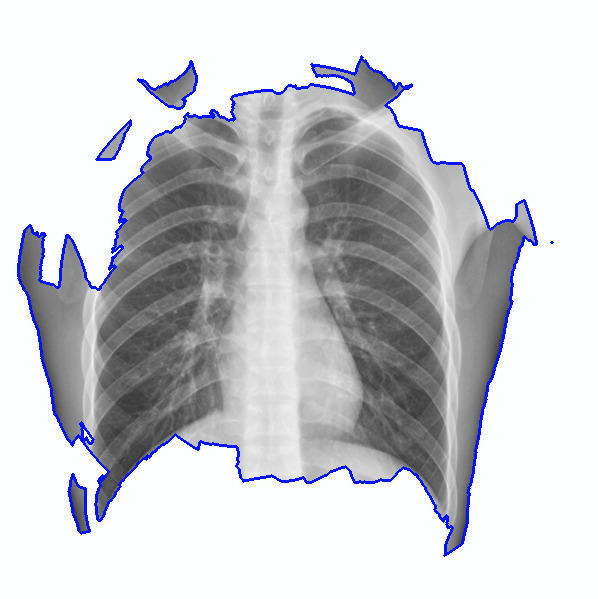} &
                 \\
                 \includegraphics[width=0.09\textwidth]{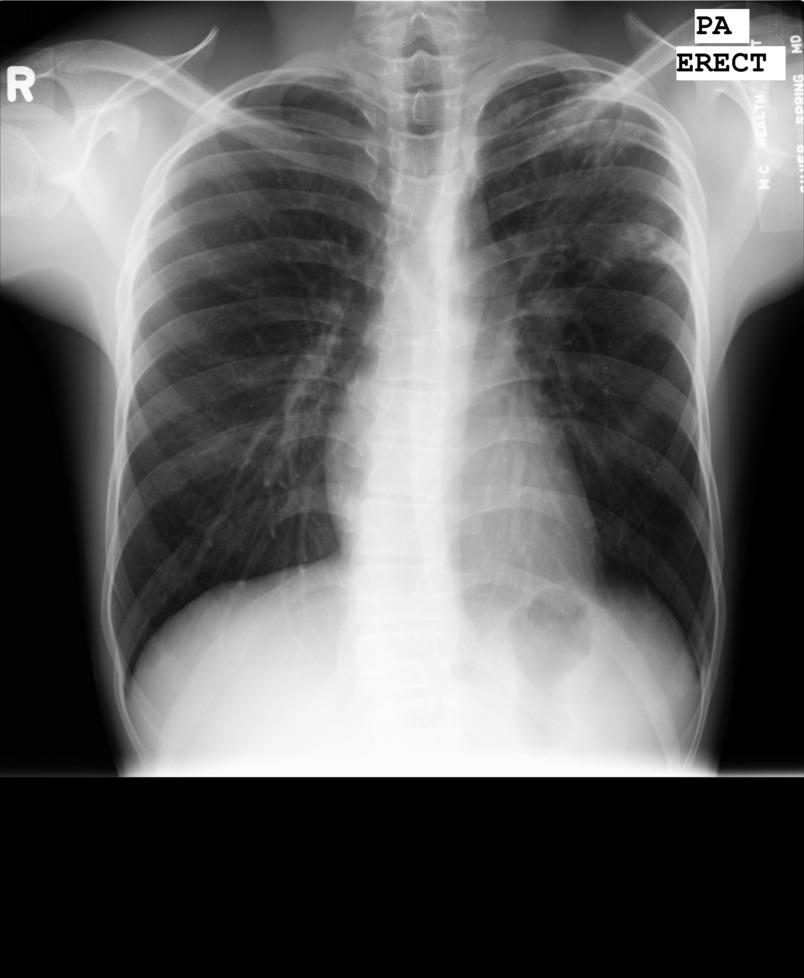} &
                 \includegraphics[width=0.09\textwidth]{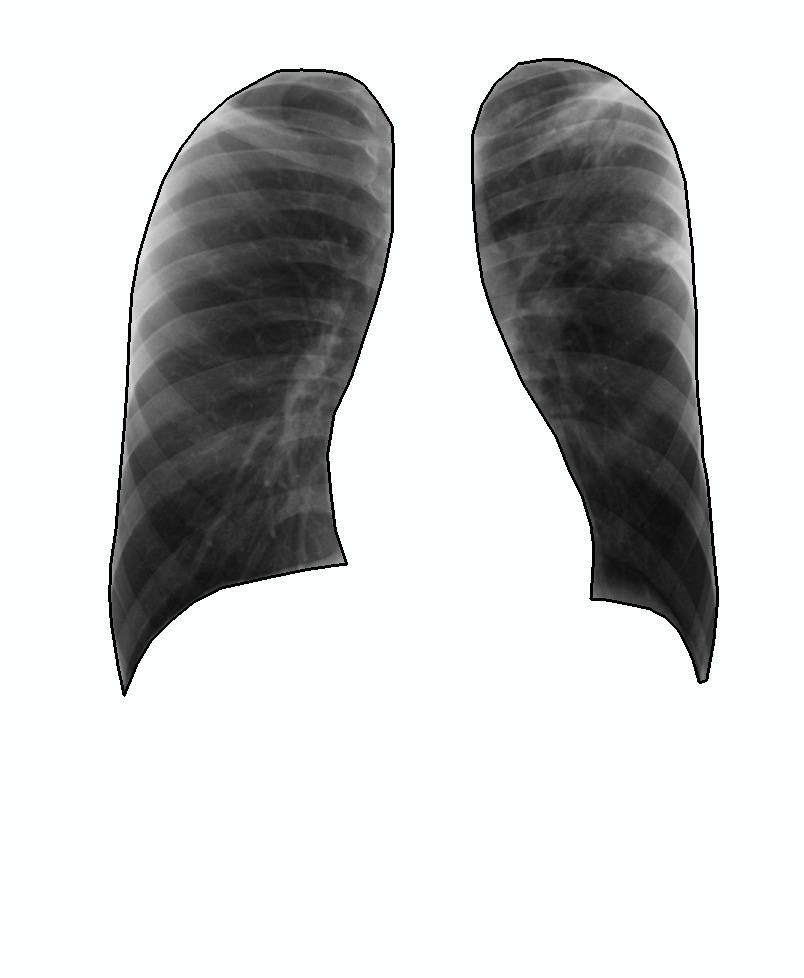} &
                 \includegraphics[width=0.09\textwidth]{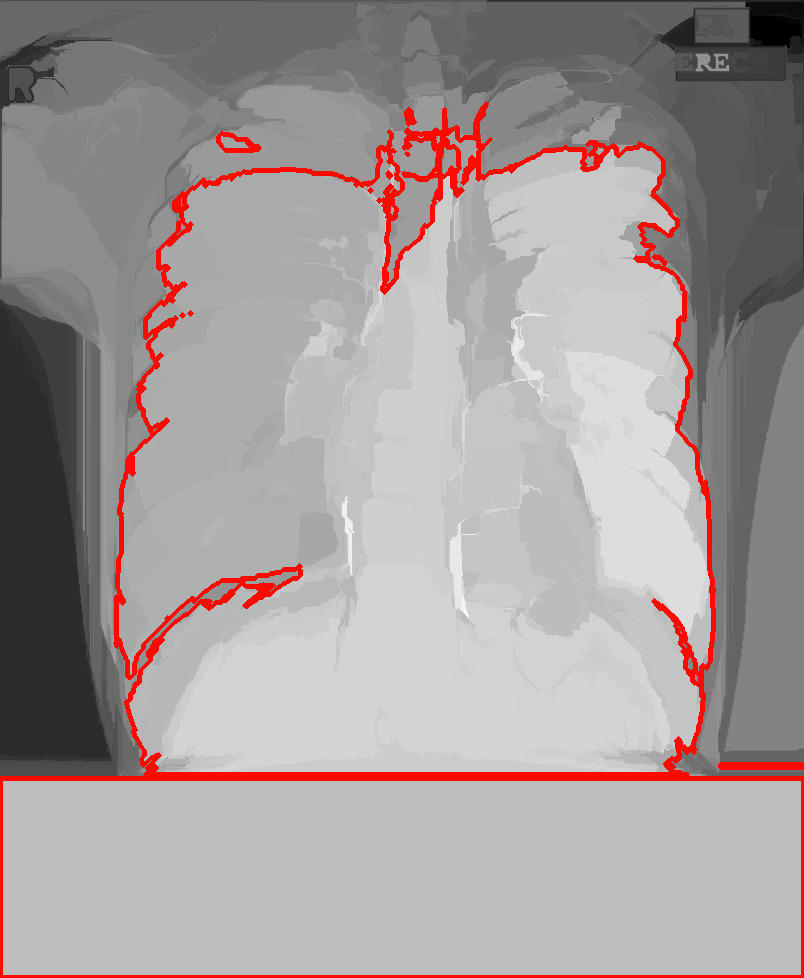} &
                 \includegraphics[width=0.09\textwidth]{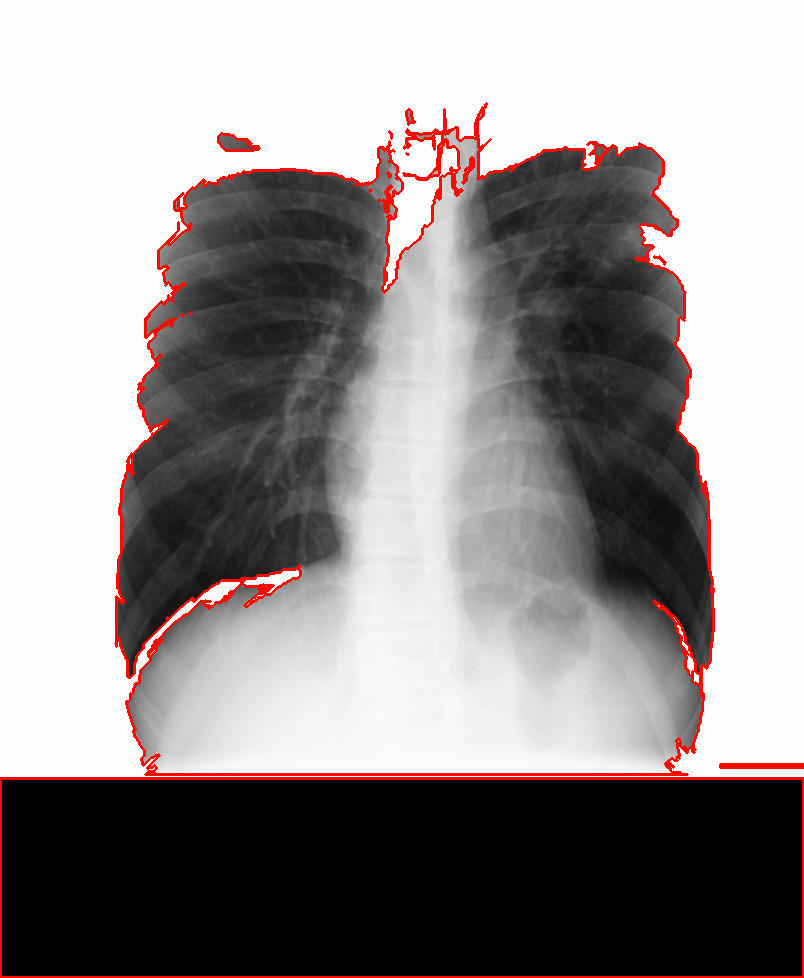} &
                 \includegraphics[width=0.09\textwidth]{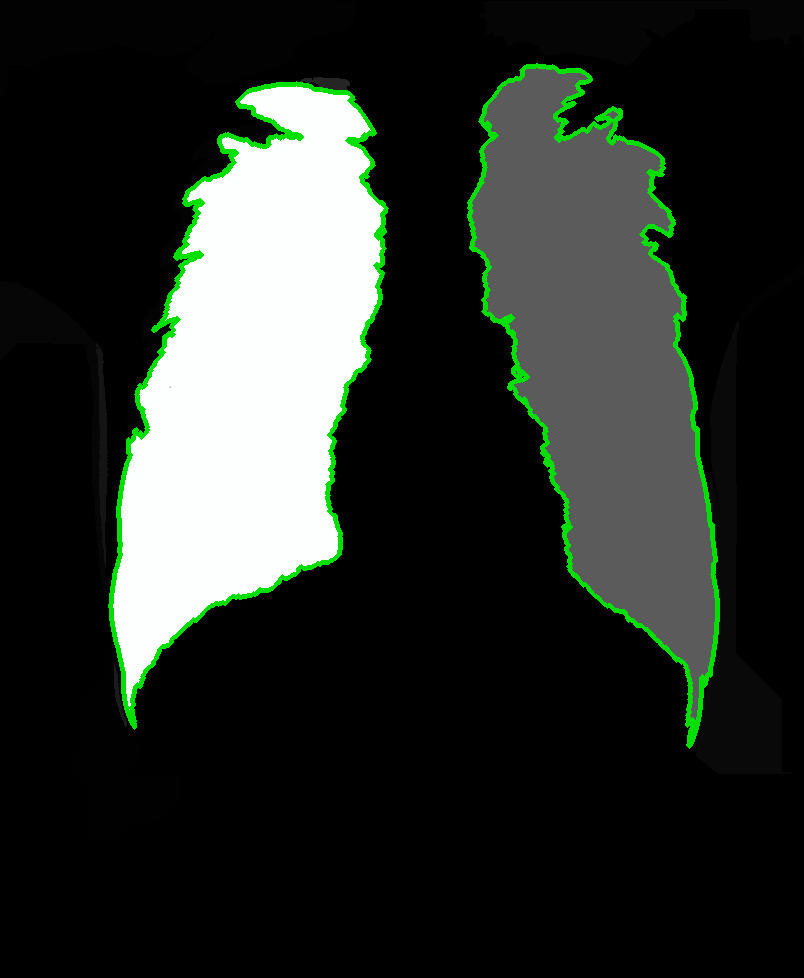} &
                 \includegraphics[width=0.09\textwidth]{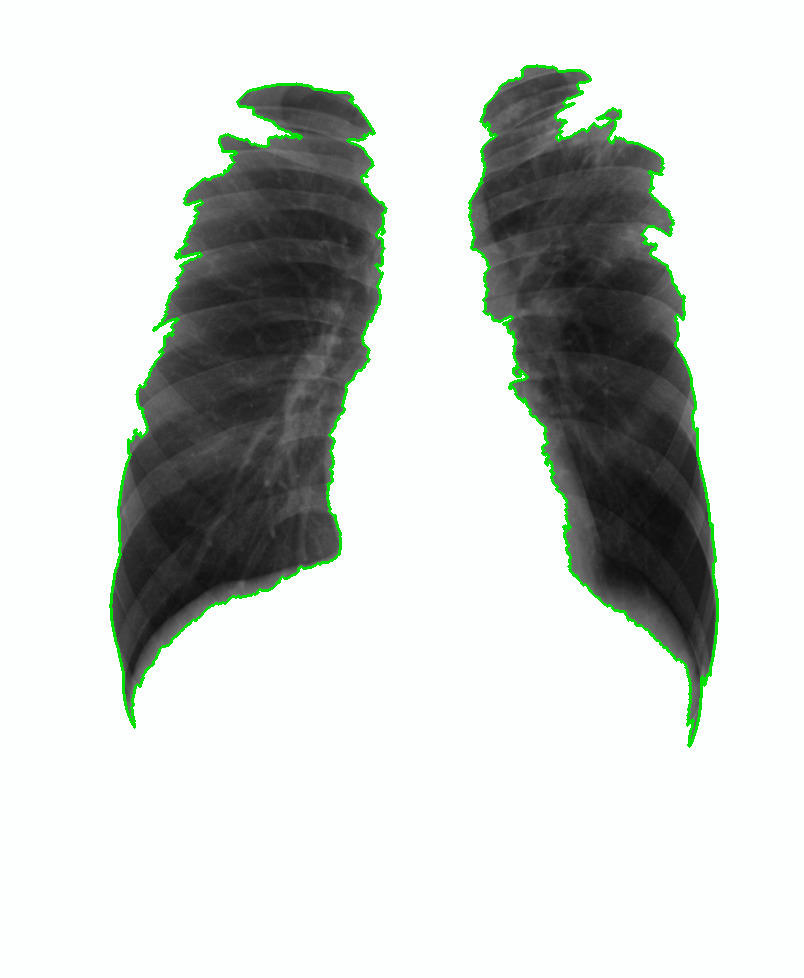} &
                 \includegraphics[width=0.09\textwidth]{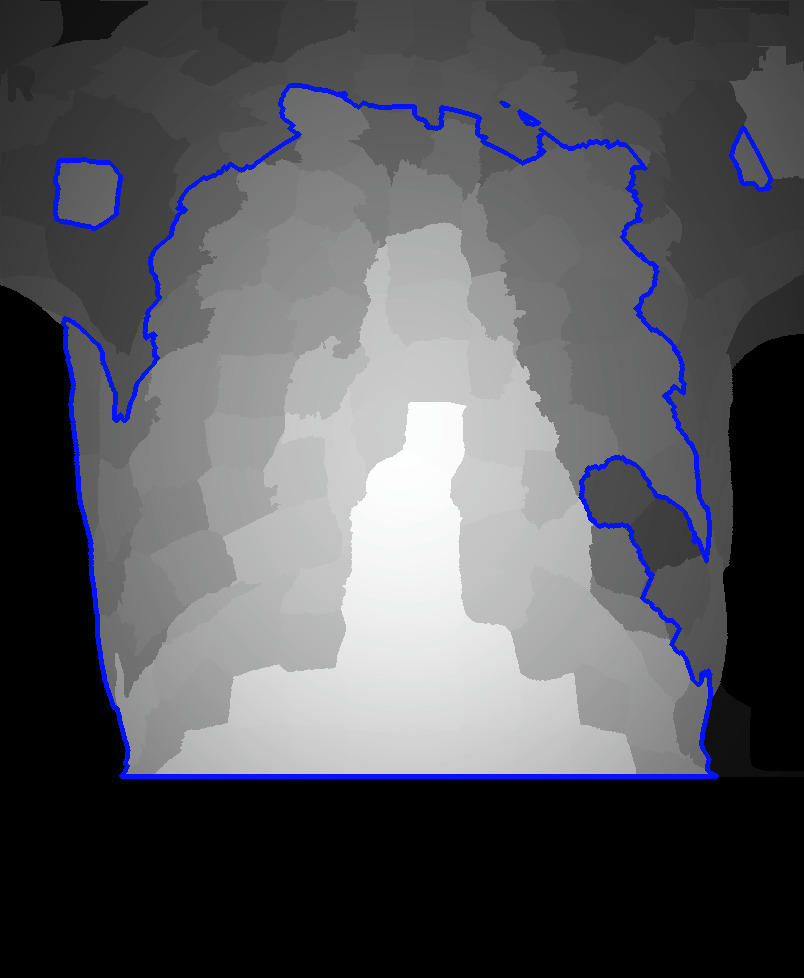} &
                 \includegraphics[width=0.09\textwidth]{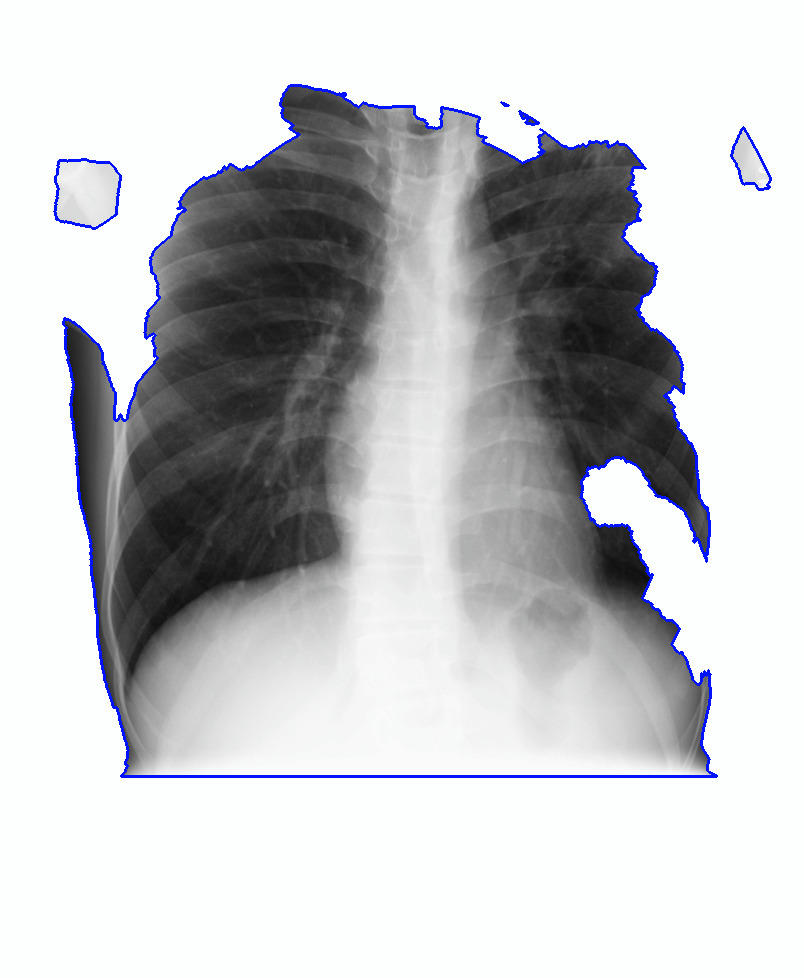} &
                 \\
                 \includegraphics[width=0.09\textwidth]{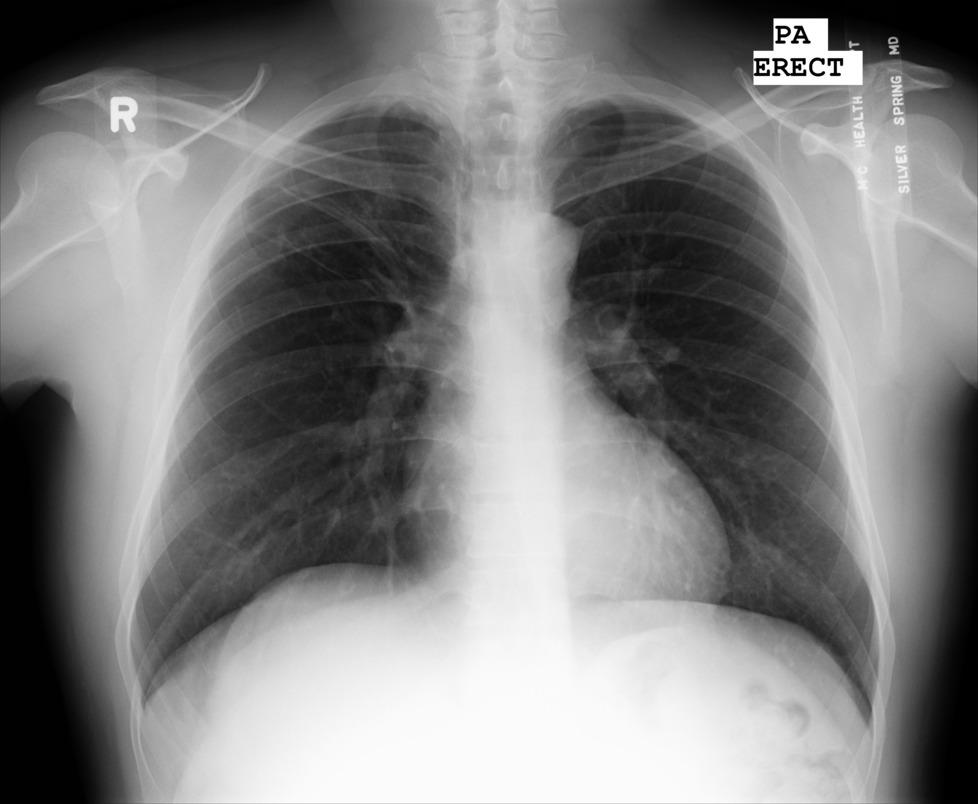} &
                 \includegraphics[width=0.09\textwidth]{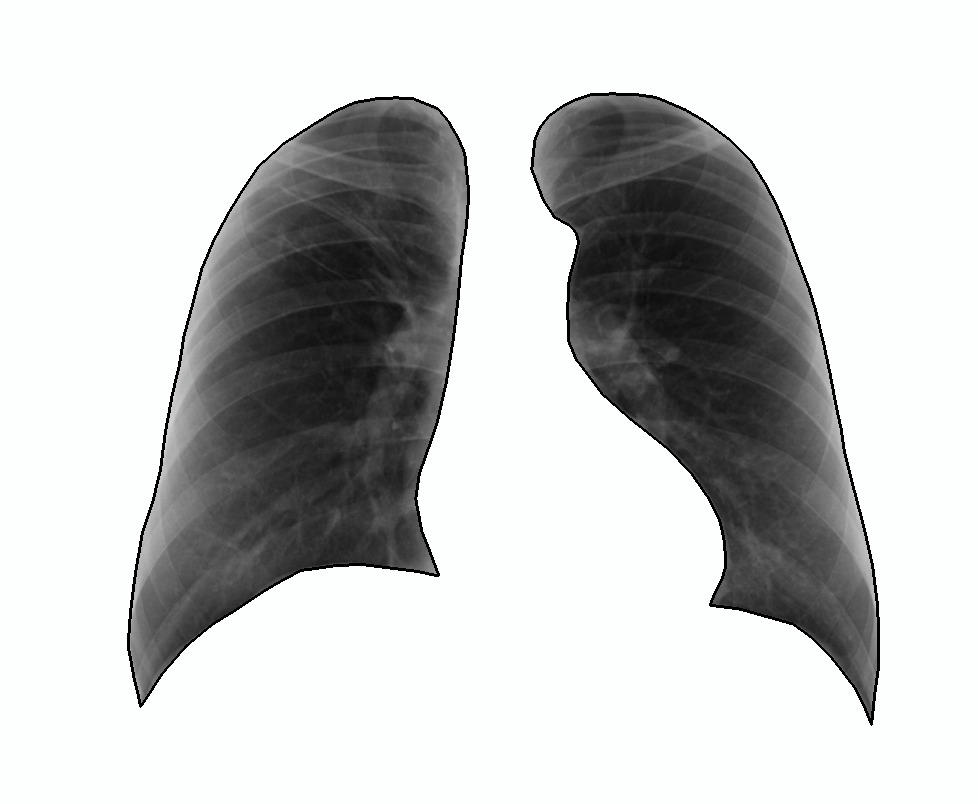} &
                 \includegraphics[width=0.09\textwidth]{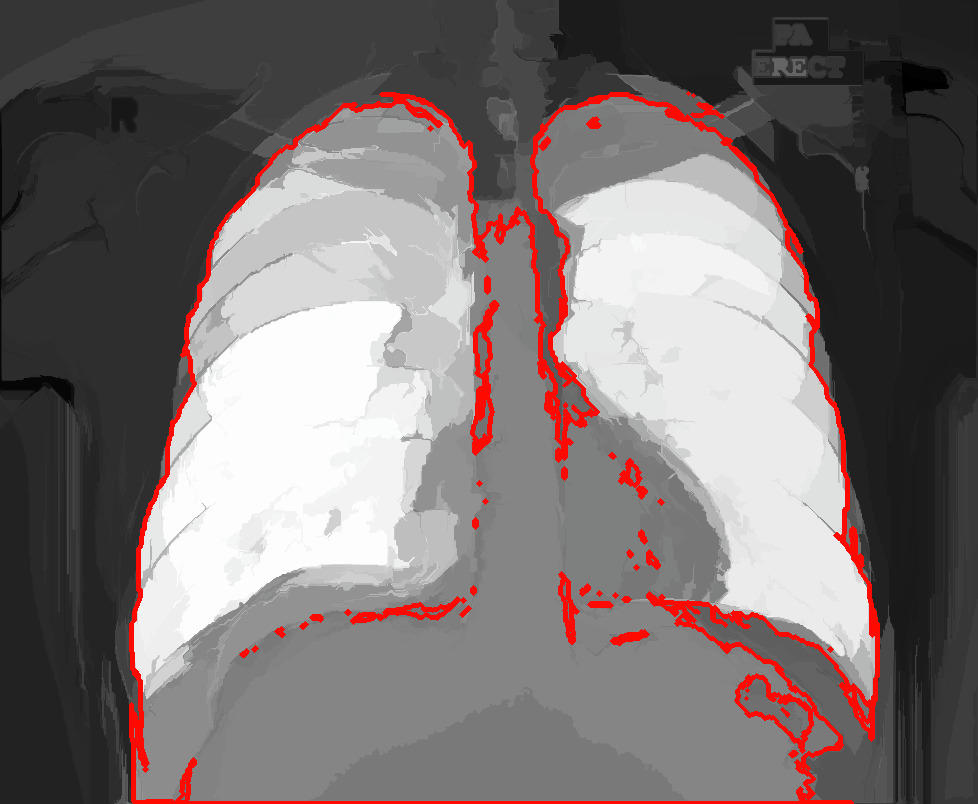} &
                 \includegraphics[width=0.09\textwidth]{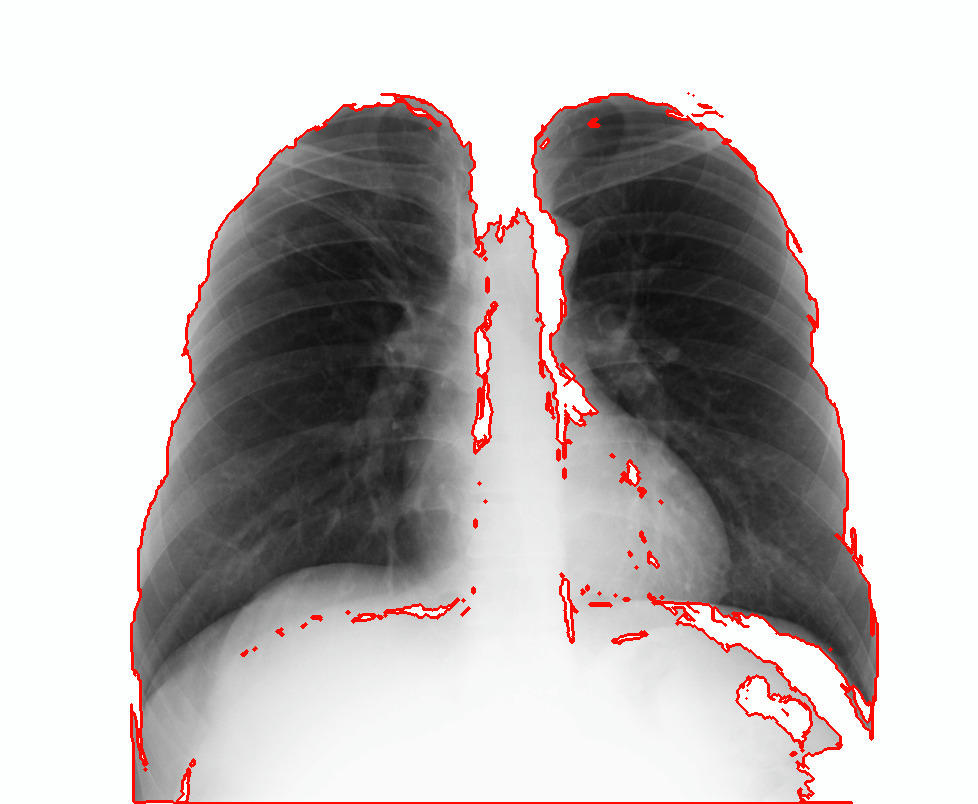} &
                 \includegraphics[width=0.09\textwidth]{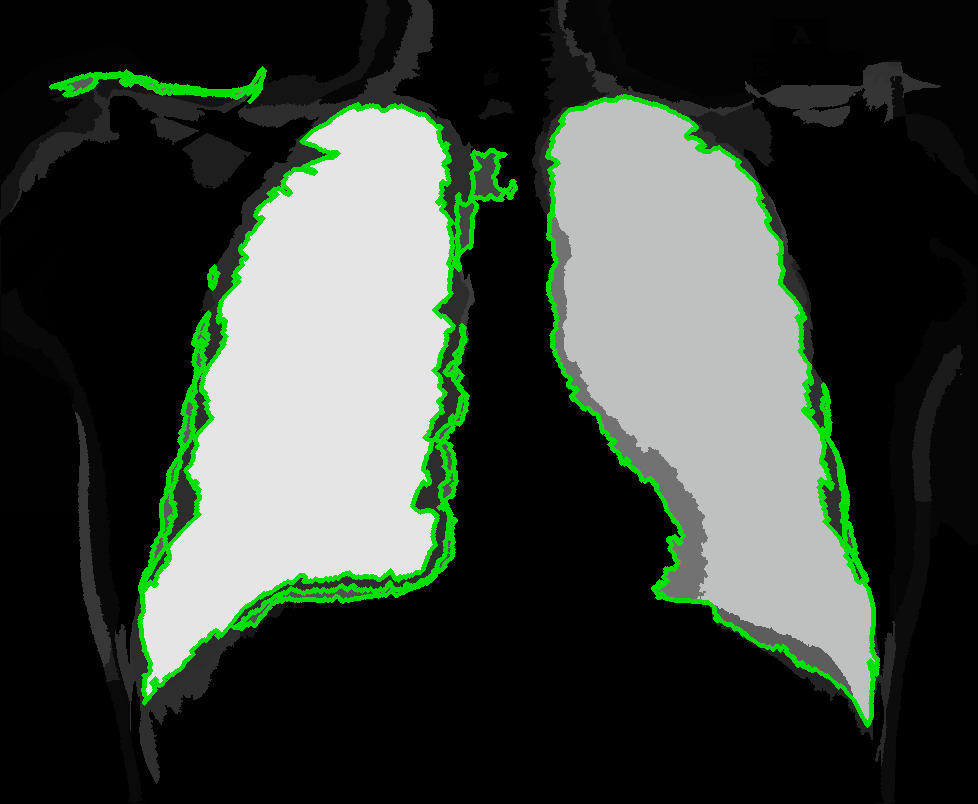} &
                 \includegraphics[width=0.09\textwidth]{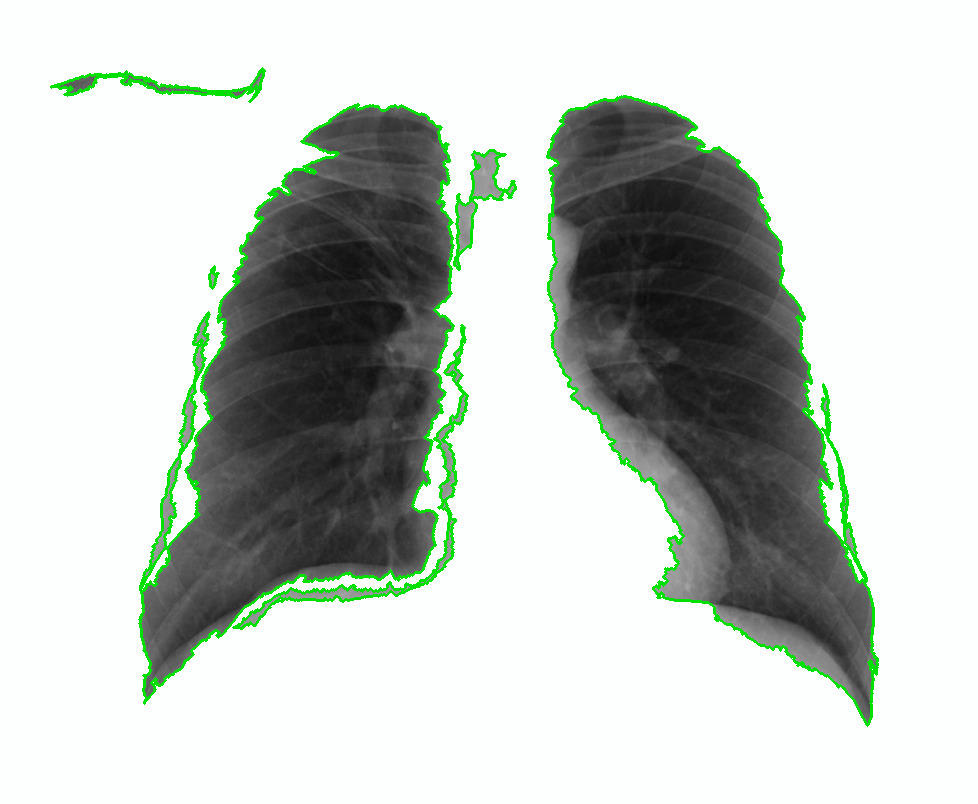} &
                 \includegraphics[width=0.09\textwidth]{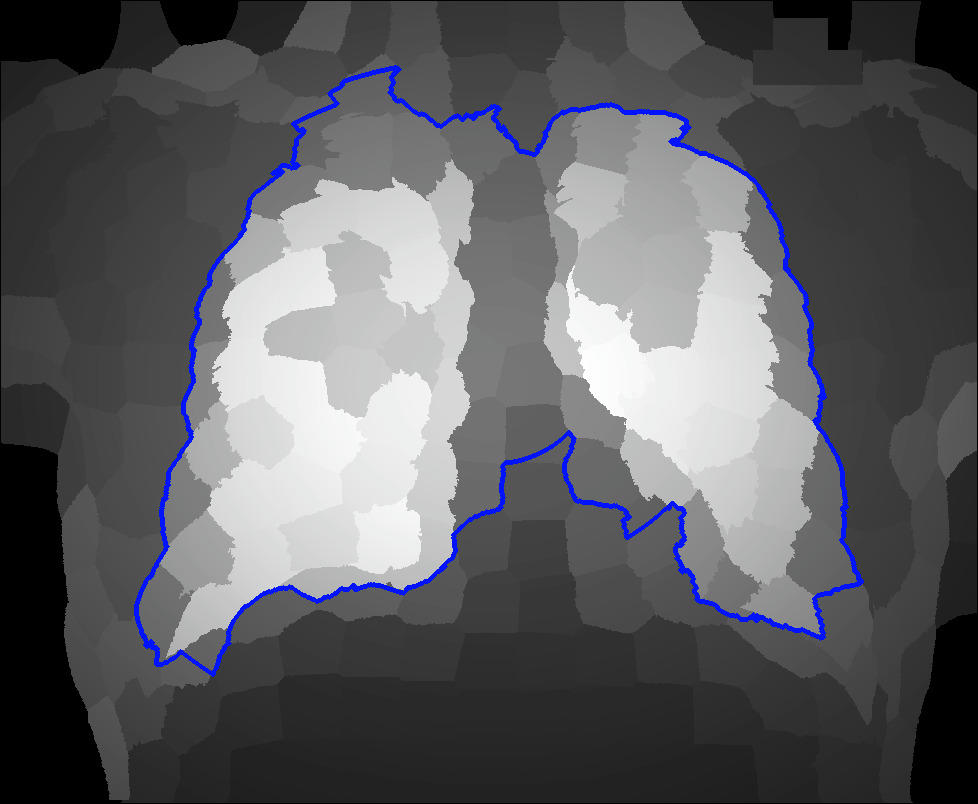} &
                 \includegraphics[width=0.09\textwidth]{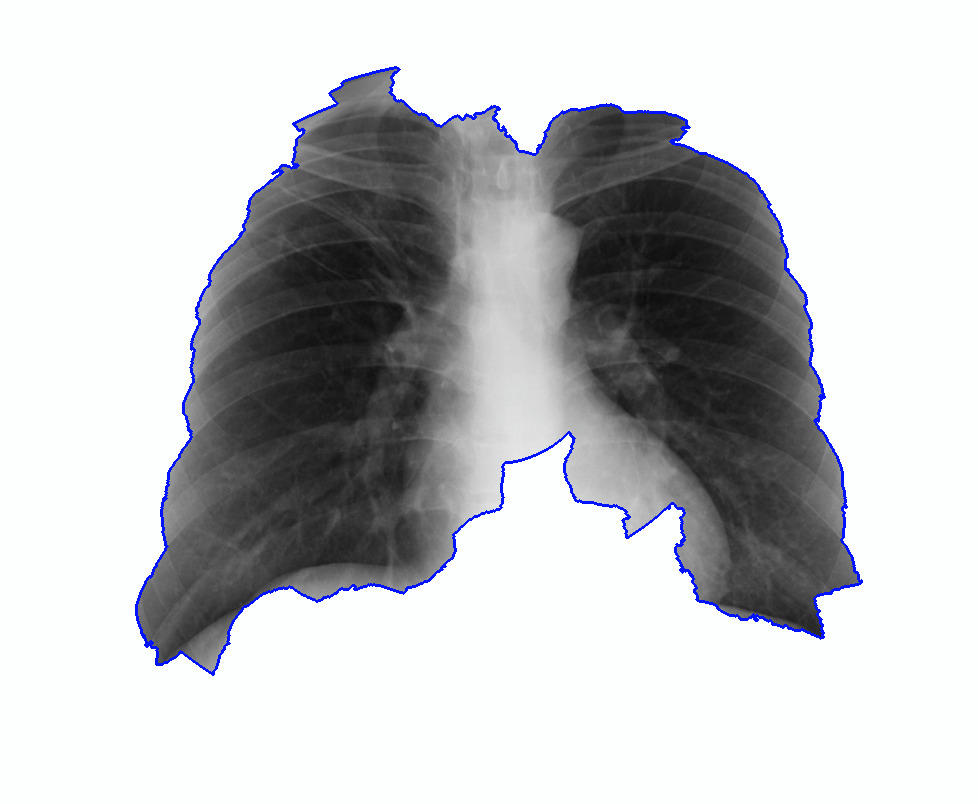} &
                 \\
                 \includegraphics[width=0.09\textwidth]{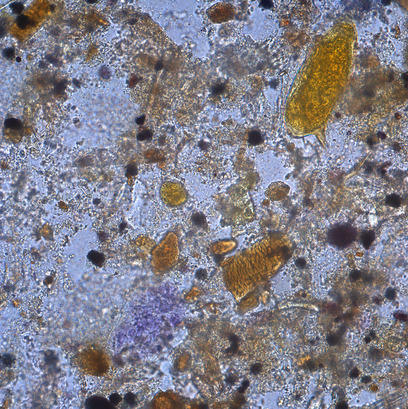} &
                 \includegraphics[width=0.09\textwidth]{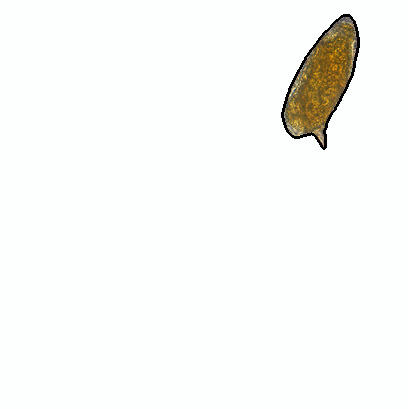} &
                 \includegraphics[width=0.09\textwidth]{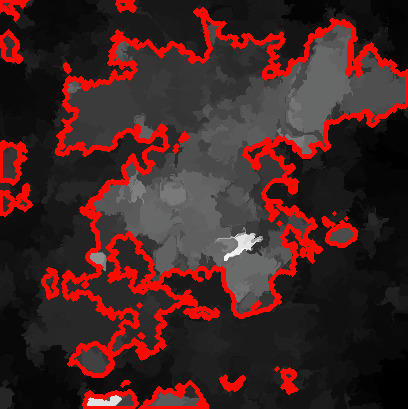} &
                 \includegraphics[width=0.09\textwidth]{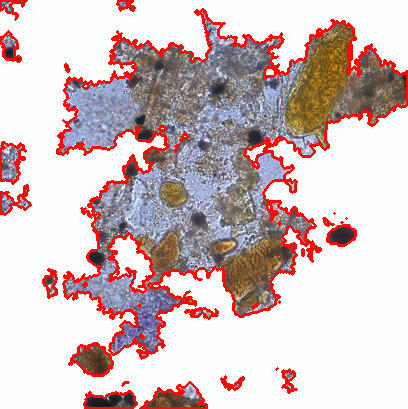} &
                 \includegraphics[width=0.09\textwidth]{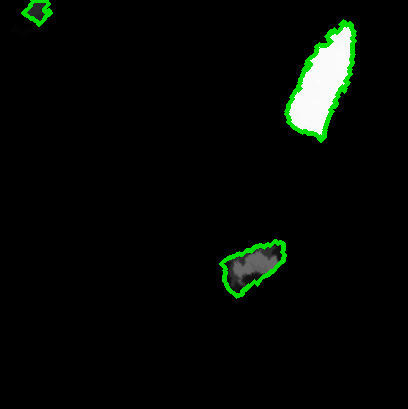} &
                 \includegraphics[width=0.09\textwidth]{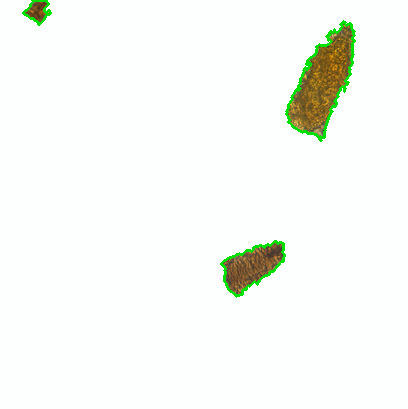} &
                 \includegraphics[width=0.09\textwidth]{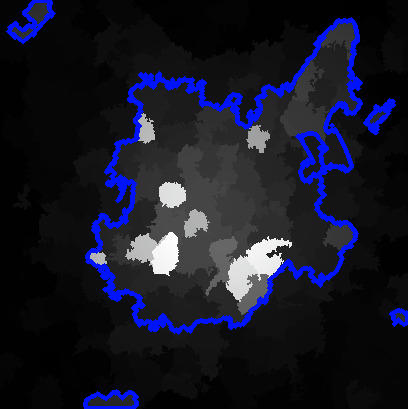} &
                 \includegraphics[width=0.09\textwidth]{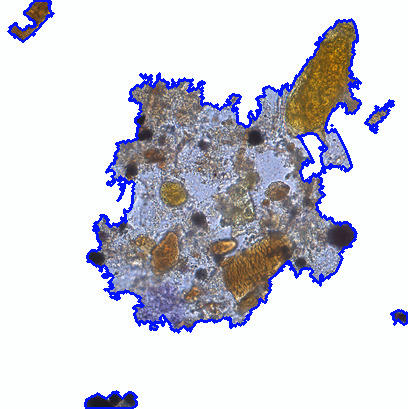} &
                 \\
                 \includegraphics[width=0.09\textwidth]{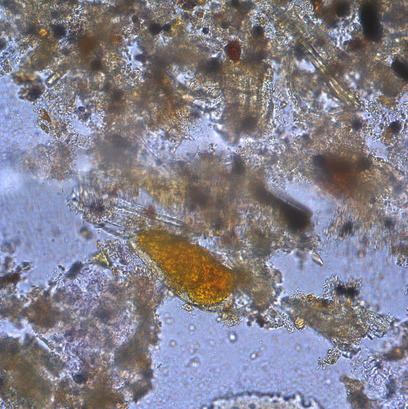} &
                 \includegraphics[width=0.09\textwidth]{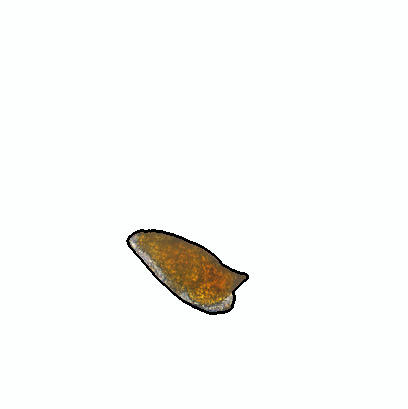} &
                 \includegraphics[width=0.09\textwidth]{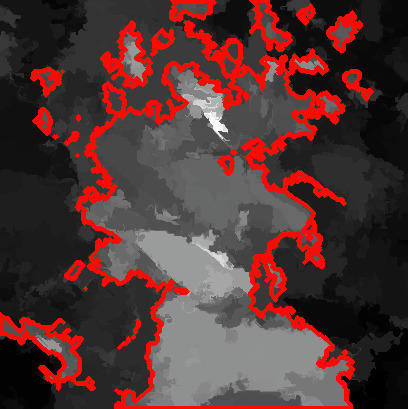} &
                 \includegraphics[width=0.09\textwidth]{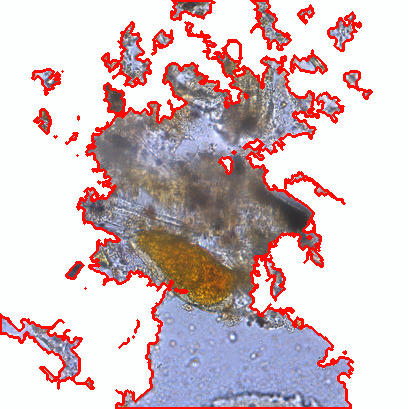} &
                 \includegraphics[width=0.09\textwidth]{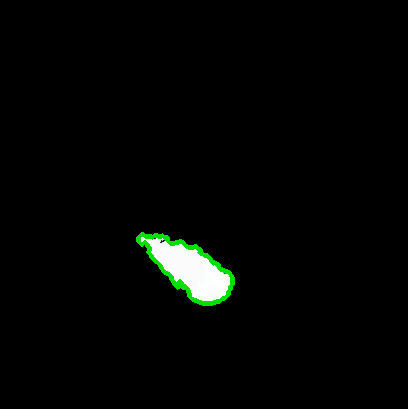} &
                 \includegraphics[width=0.09\textwidth]{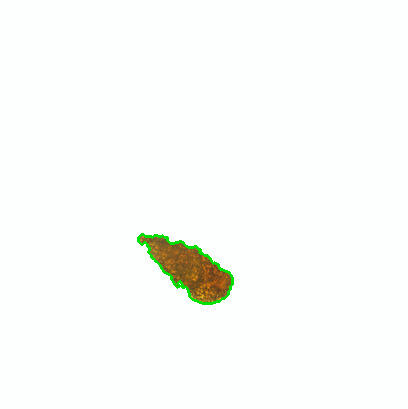} &
                 \includegraphics[width=0.09\textwidth]{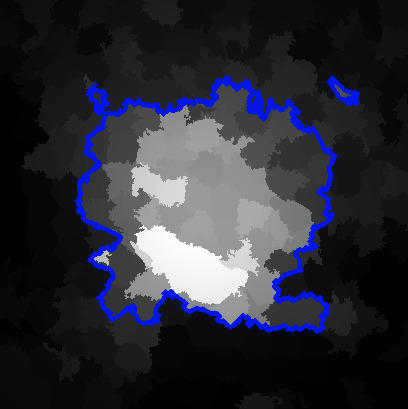} &
                 \includegraphics[width=0.09\textwidth]{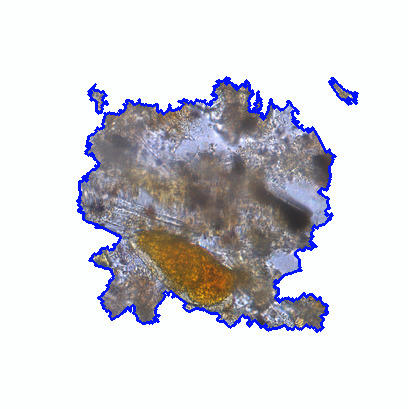} &
                 \\
                 \includegraphics[width=0.09\textwidth]{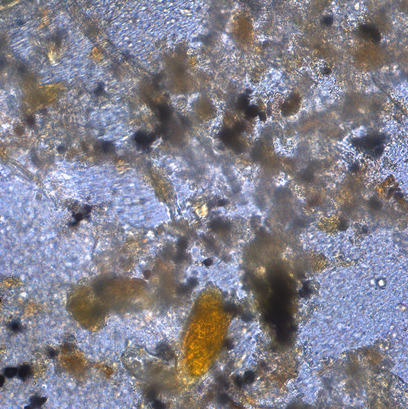} &
                 \includegraphics[width=0.09\textwidth]{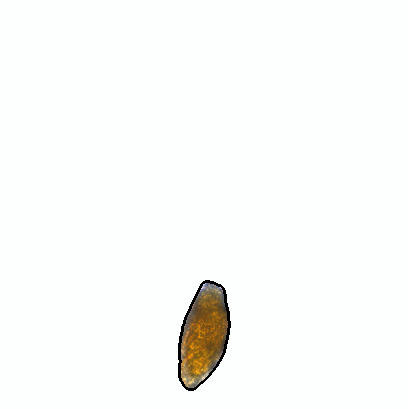} &
                 \includegraphics[width=0.09\textwidth]{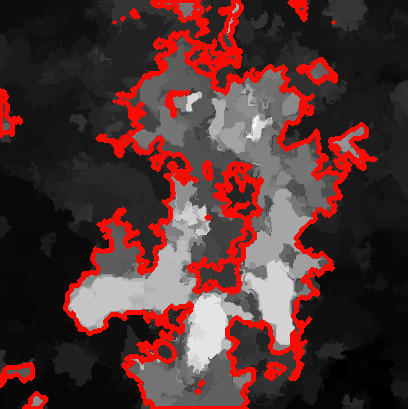} & 
                 \includegraphics[width=0.09\textwidth]{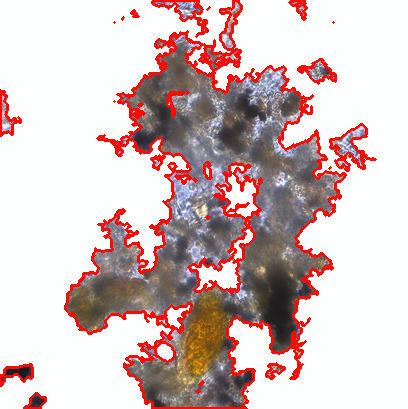} &
                 \includegraphics[width=0.09\textwidth]{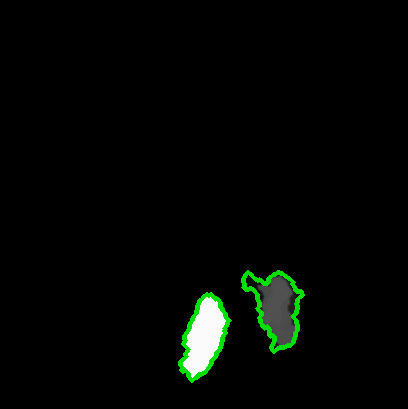} &
                 \includegraphics[width=0.09\textwidth]{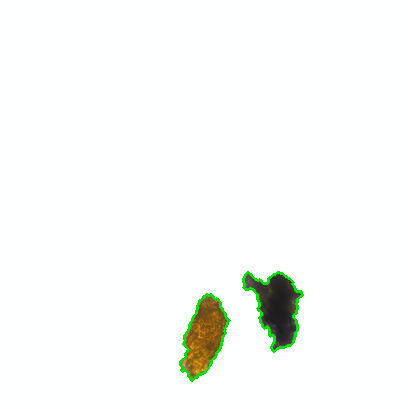} &
                 \includegraphics[width=0.09\textwidth]{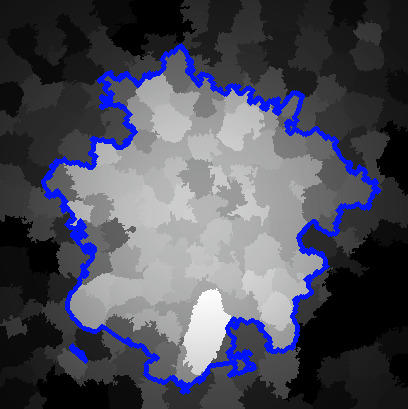} &
                 \includegraphics[width=0.09\textwidth]{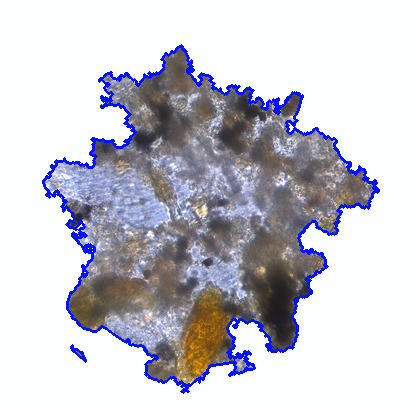} &
                 \\
                 \includegraphics[width=0.09\textwidth]{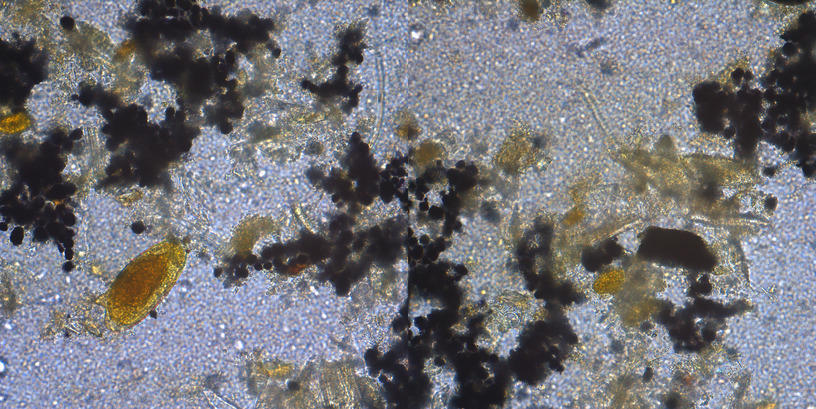} &
                 \includegraphics[width=0.09\textwidth]{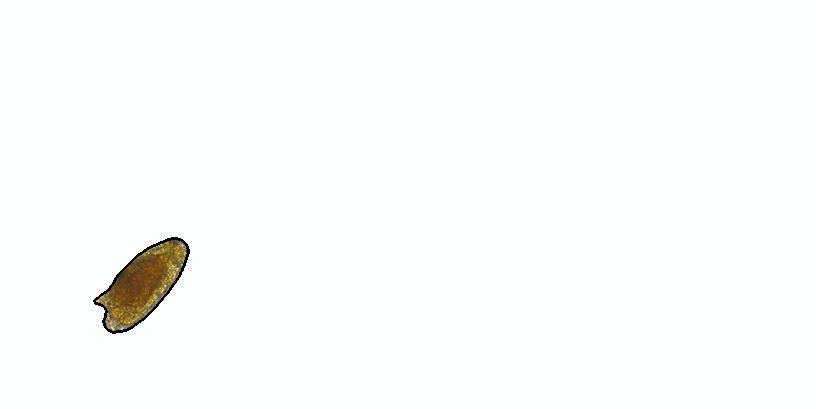} &
                 \includegraphics[width=0.09\textwidth]{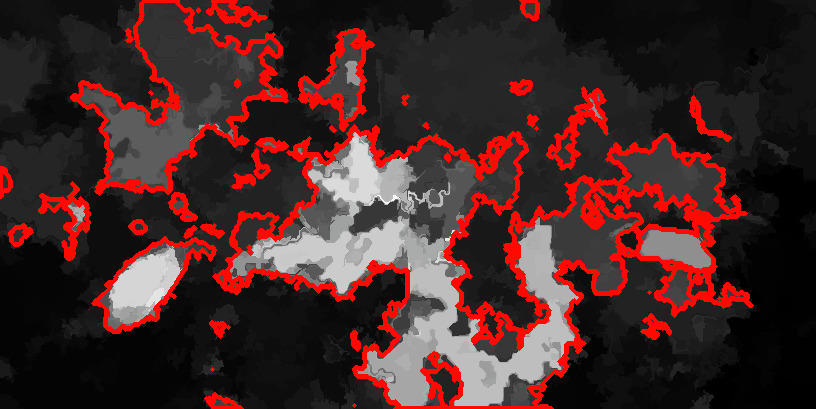} &
                 \includegraphics[width=0.09\textwidth]{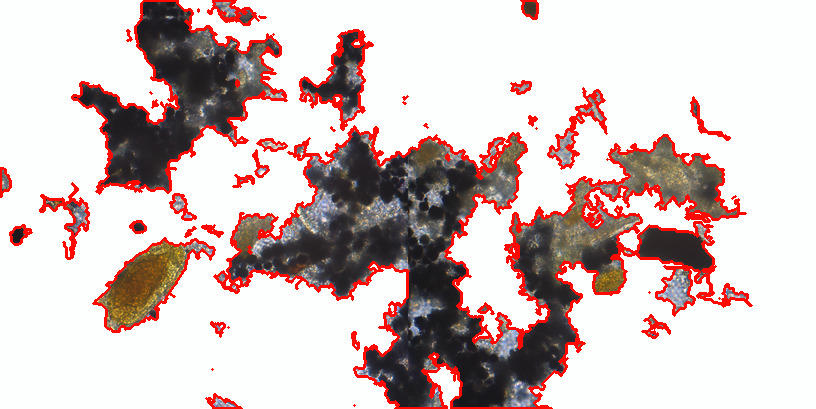} &
                 \includegraphics[width=0.09\textwidth]{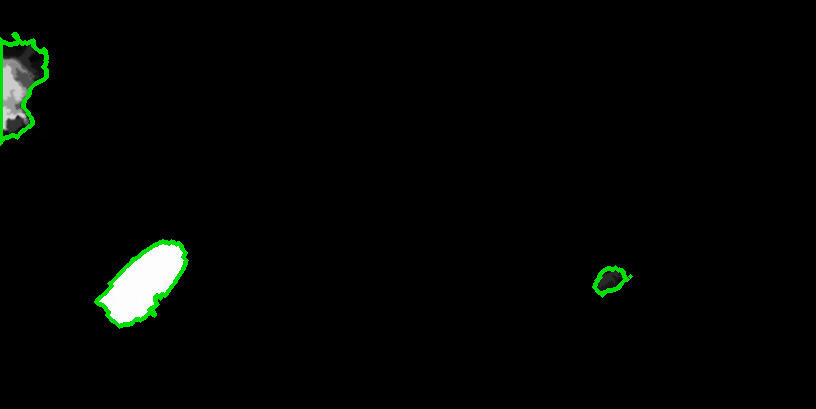} &
                 \includegraphics[width=0.09\textwidth]{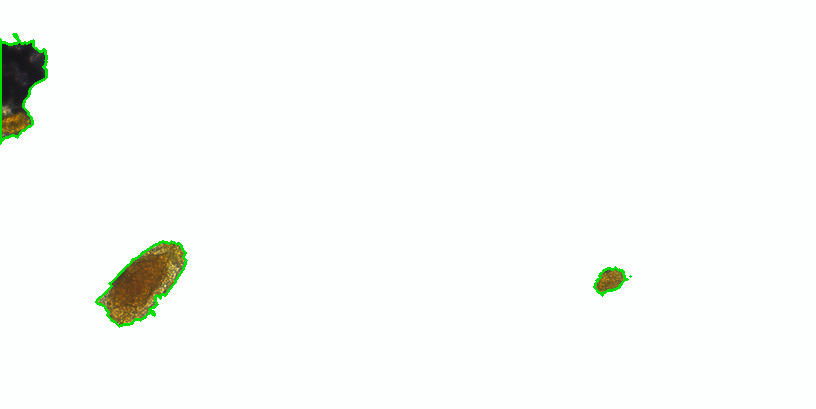} &
                 \includegraphics[width=0.09\textwidth]{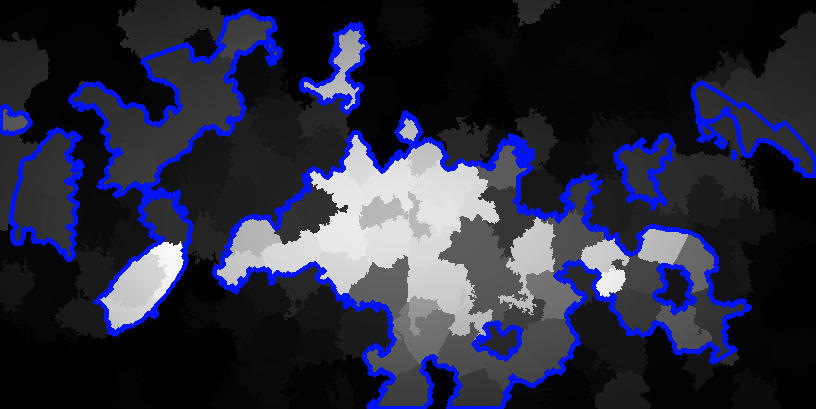} &
                 \includegraphics[width=0.09\textwidth]{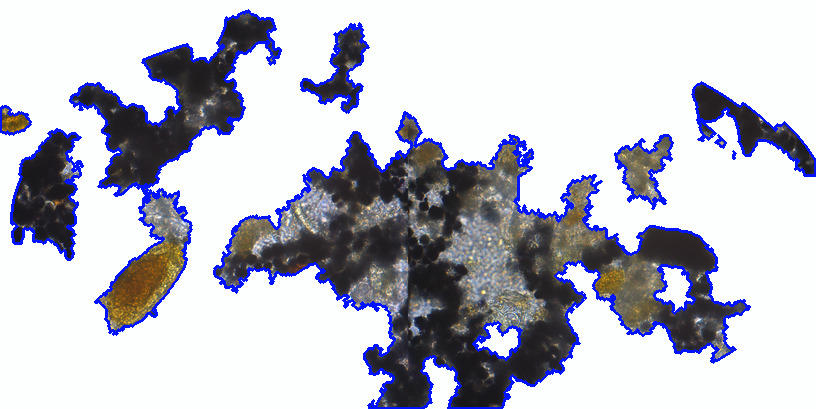} &
                 \\
                 \includegraphics[width=0.09\textwidth]{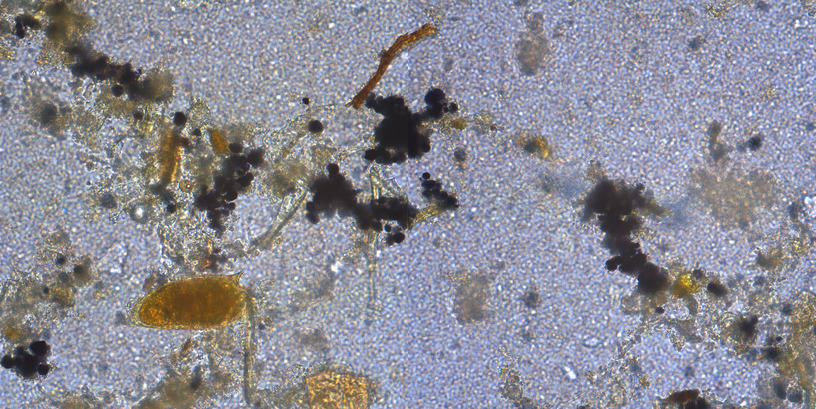} &
                 \includegraphics[width=0.09\textwidth]{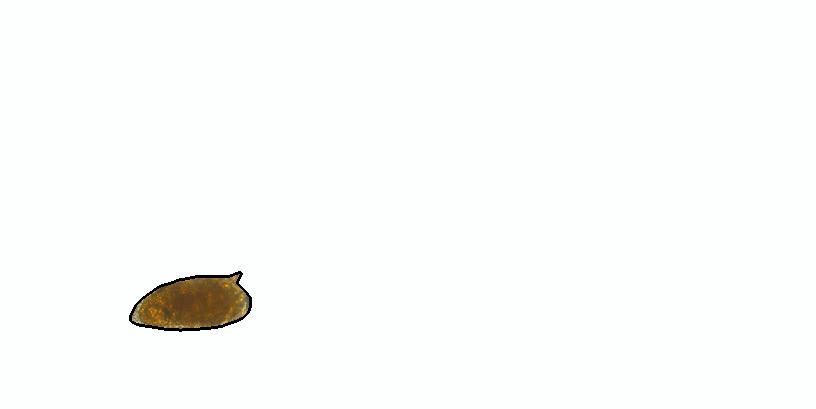} &
                 \includegraphics[width=0.09\textwidth]{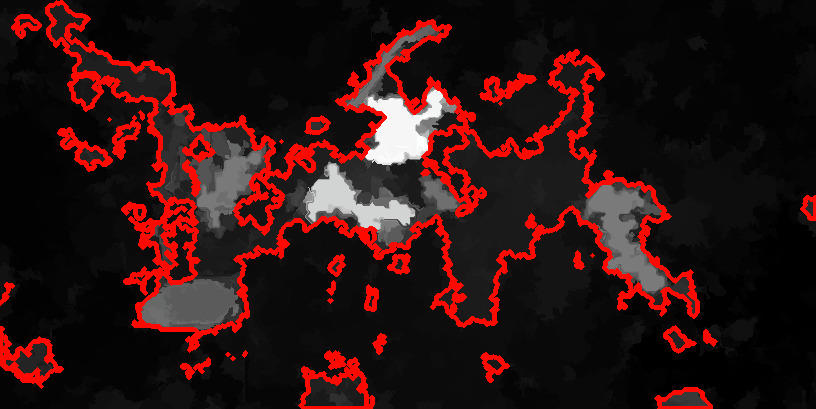} &
                 \includegraphics[width=0.09\textwidth]{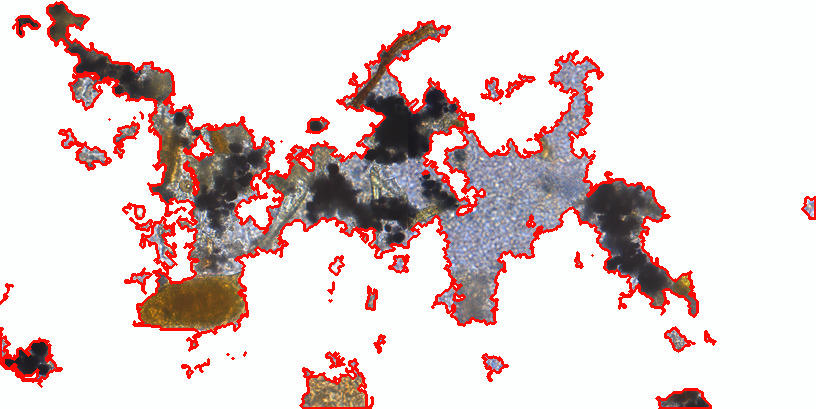} &
                 \includegraphics[width=0.09\textwidth]{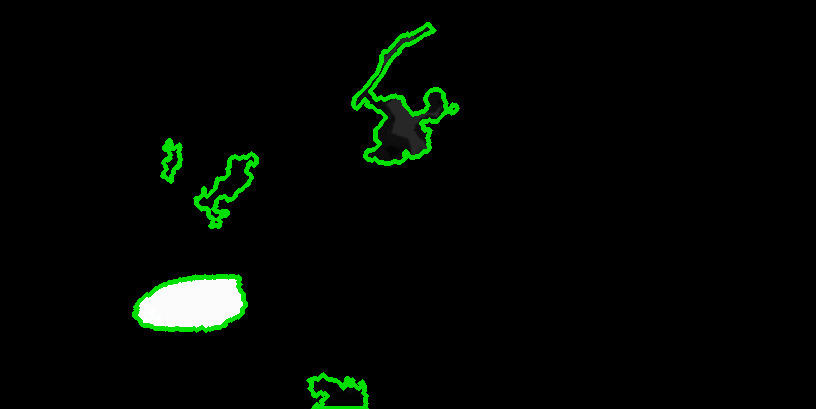} &
                 \includegraphics[width=0.09\textwidth]{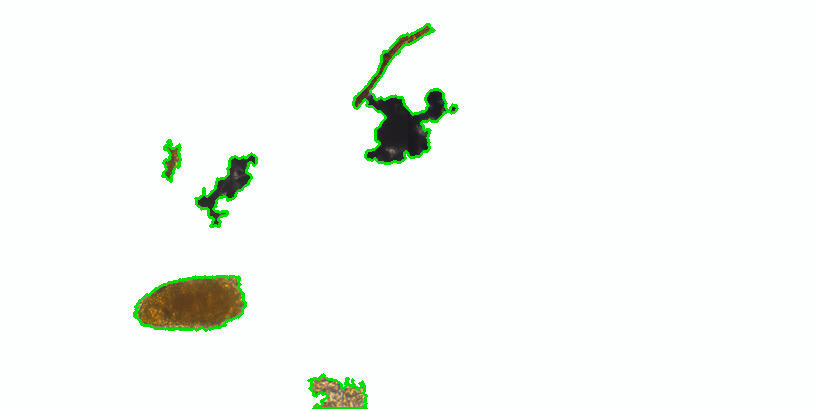} &
                 \includegraphics[width=0.09\textwidth]{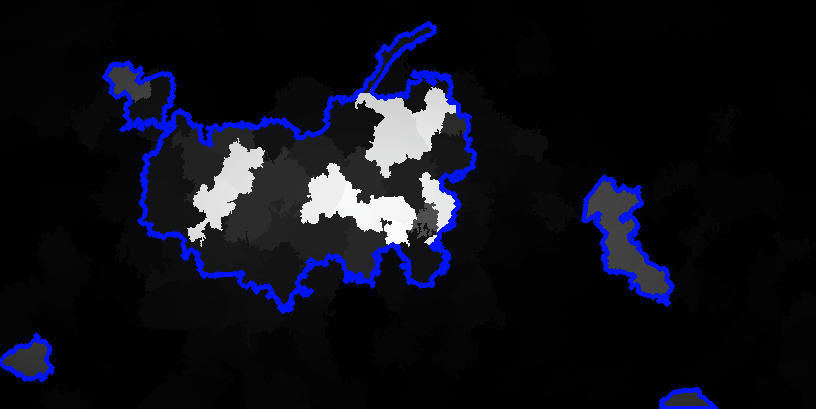} &
                 \includegraphics[width=0.09\textwidth]{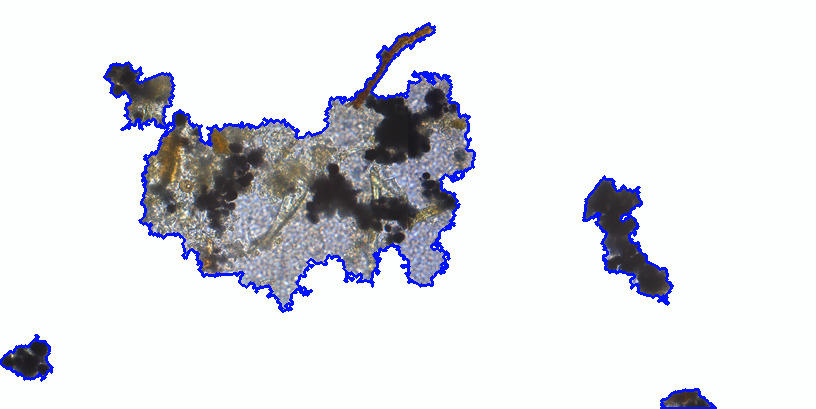} &
                 \\
                (a) & (b) & (c) & (d) & (e) & (f) & (g) & (h)
            \end{tabular}
        \caption{(a) Input image. (b) Ground-truth segmentation; (c-h) DRFI/ITSELF/SMD saliency maps with mean-saliency threshold  segmentation boundaries depicted on red/green/blue, respectively Note how ITSELF tend to create more accentuated contrast between the object and background, adhering to the boundaries. }
        
        \label{fig:segmentation_results-non-natural}
\end{figure*}

    Concerning the two evaluated non-natural-image datasets, ITSELF outperformed both SMD and DRFI by a big margin, especially on the parasite dataset (Table \ref{tab:quantitative-results}). Figure \ref{fig:segmentation_results-non-natural} shows side-by-side example results of the three methods. While DRFI and SMD create over-salient regions, ITSELF provides a better definition of the objects. 
    
    On the x-ray images, the inside of the lungs has different characteristics compared to the ribs. Because most of the patients' lung boundaries overlap with their ribs, ITSELF's results cannot achieve a high BR score. However, ITSELF obtained substantially higher precision when compared to the other methods.
    
\begin{figure}[t!]
    \centering
    \begin{tabular}{c c c}
             \includegraphics[width=0.13\textwidth]{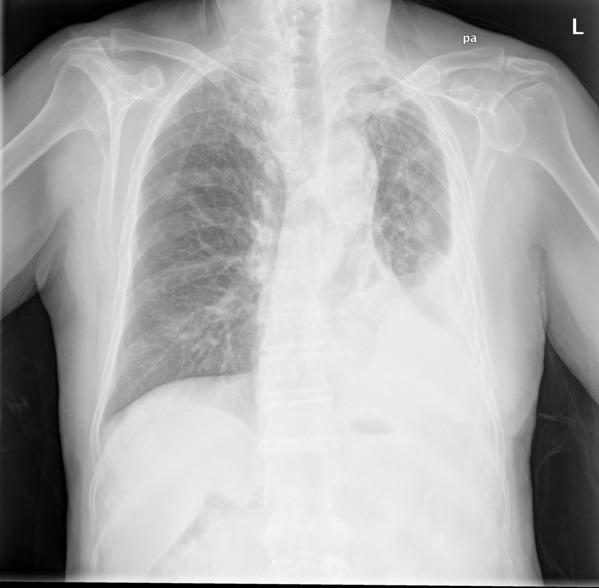} &
             \includegraphics[width=0.13\textwidth]{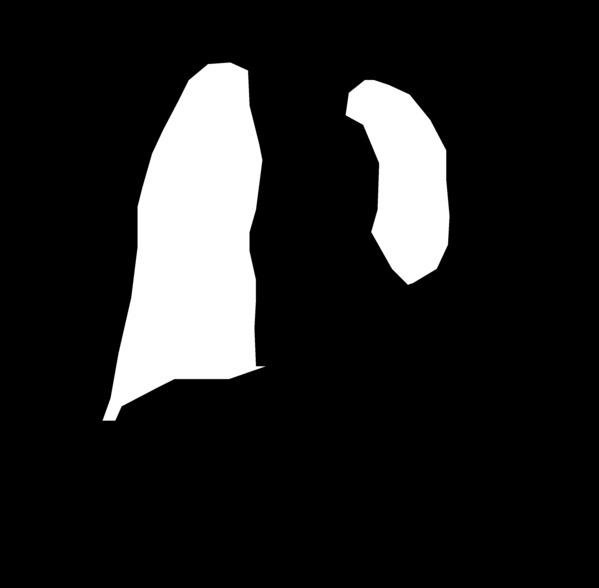} &
             \includegraphics[width=0.13\textwidth]{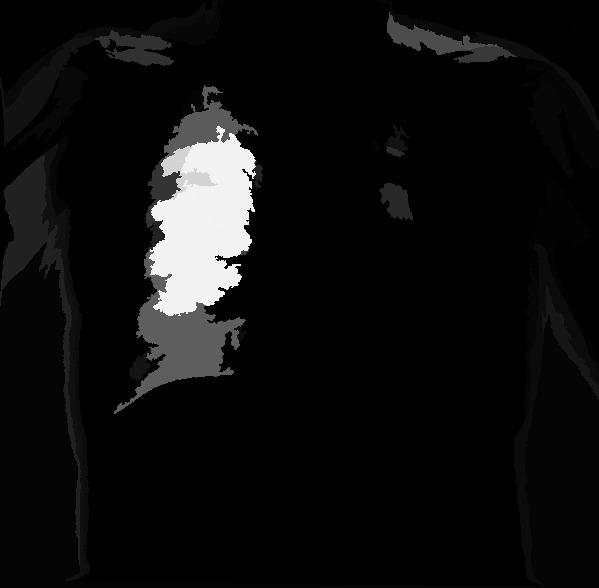} \\
            (a) & (b) & (c)
        \end{tabular}
    \caption{(a) Original image. (b) Ground-truth; (c) ITSELF saliency map. ITSELF completely lost the smaller and brighter lung.}
    
    \label{fig:fail-xray-unhealthy}
    \end{figure}

    ITSELF fails to correctly detect both lungs on images where one is too small or has higher intensities than the other (Figure \ref{fig:fail-xray-unhealthy}). These are characteristics of not healthy images, so future works might use ITSELF segmentation error to indicate unhealthy patients.
    
\begin{figure}[t!]
    \centering
    \begin{tabular}{c c}
             \includegraphics[width=0.18\textwidth]{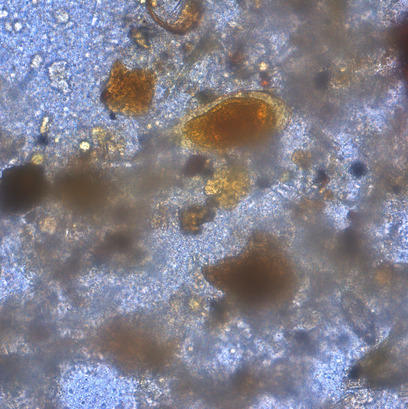} &
             \includegraphics[width=0.18\textwidth]{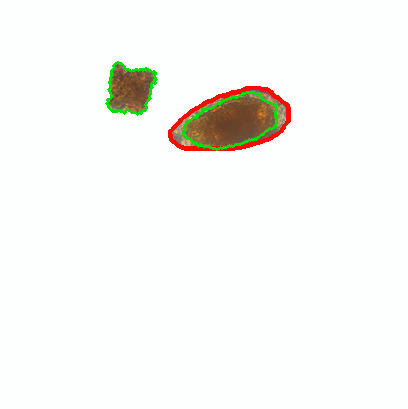} \\
            (a) & (b)
        \end{tabular}
    \caption{(a) Original image. (b) Ground-truth (red) and ITSELF (green) segmentations overlayed. Note the lighter yellow membrane segmented on the ground-truth that was lost by ITSELF.}
    
    \label{fig:parasite-membrane}
    \end{figure}

    A similar issue to the ribs happens on the parasite-eggs dataset. The parasite-eggs are enclosed by a membrane that often gets less colored than the egg's core (Figure \ref{fig:parasite-membrane}).  
    
\begin{figure}[t!]
    \begin{tabular}{c c c}
             \includegraphics[width=0.13\textwidth]{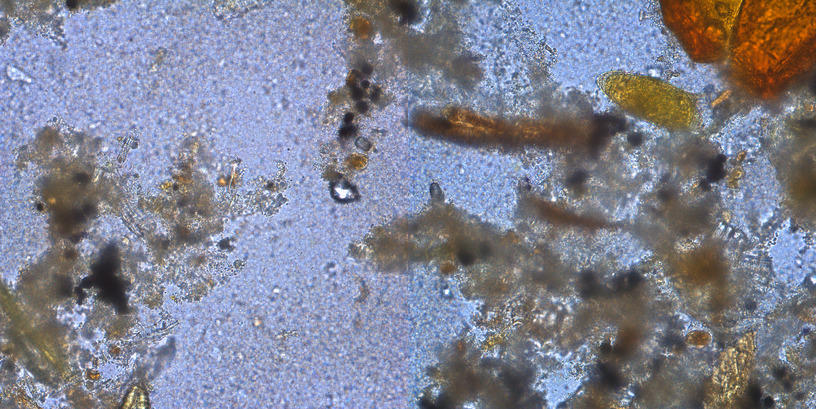} &
             \includegraphics[width=0.13\textwidth]{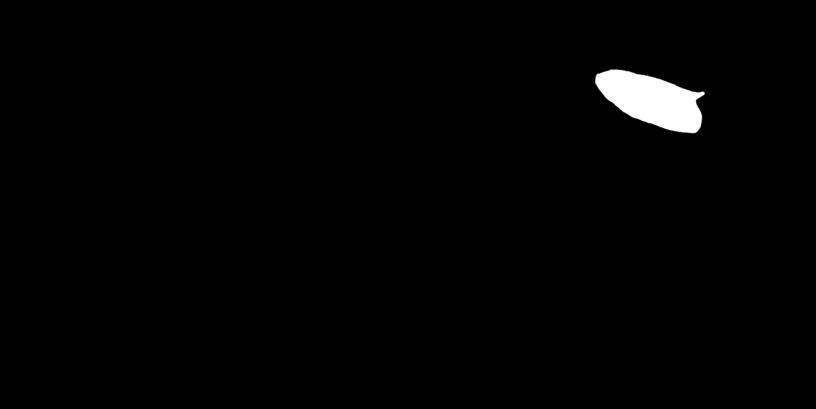} &
             \includegraphics[width=0.13\textwidth]{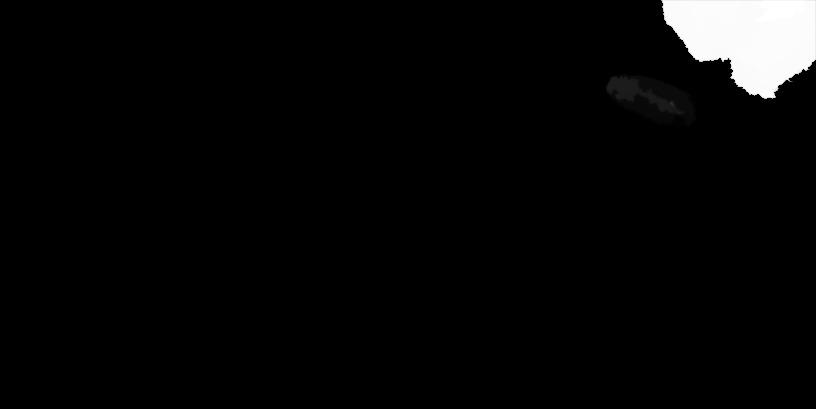} \\
            (a) & (b) & (c)
        \end{tabular}
    \caption{(a) Original image. (b) Ground-truth; (c) ITSELF saliency map. ITSELF highlights the top right impurity instead of the parasite-egg.}
    
    \label{fig:fail-parasite-membrane}
    \end{figure}

    ITSELF fails to accurately detect the salient parasite eggs on a few images where the impurities are elliptical, share similar colors, and are within the size range (Figure \ref{fig:fail-parasite-membrane}). However, ITSELF's BR, and $REC^\omega$ scores are mostly affected by the miss-estimation of the membrane saliency.

\section{Conclusion}\label{sec:conclusion}
    We have presented ITSELF, a saliency estimation framework that is flexible for multiple image domains and allows the user to tailor salient characteristics. By using object-based superpixels, we proposed a novel loop interaction between saliency estimation and superpixel segmentation that iteratively improves both results. Thanks to that interaction, our method creates more semantically explainable maps and segmentation. 
    
    We compared implementations of our framework to two state-of-the-art saliency methods on six datasets, four of which are composed of natural images and two non-natural ones. We achieve competitive results on the natural images and outperformed by a significant margin on non-natural images. Note that we do not claim these are optimal implementations of our framework; instead, they demonstrate the framework's flexibility to different scenarios.
    
    For future work, we want to explore two ideas presented in this paper. One is to combine ITSELF to other saliency estimators, including deep-neural-networks, and evaluate its performance. Another lane is to further explore the usage of user-provided scribbles to model more priors and queries, allowing ITSELF to be used in weakly-supervised segmentation and interactive segmentation.
\FloatBarrier


%





\ifCLASSOPTIONcaptionsoff
  \newpage
\fi

\bibliographystyle{IEEEtran}
\bibliography{IEEEabrv,main}

\begin{IEEEbiographynophoto}{Leonardo de Melo Joao}
Leonardo de Melo Joao received the B.Sc. degree in computer science from the Pontifical Catholic University of Minas Gerais, Pocos de Caldas, Brazil, in 2018. He is currently pursuing the master’s degree in computer science from the University of Campinas, Campinas, Brazil. His research interests include machine learning, image processing, computer vision and medical image segmentation.
\end{IEEEbiographynophoto}

\begin{IEEEbiographynophoto}{Felipe de Castro Belem}
Felipe de Castro Belem received a M.Sc. degree in computer science from the University of Campinas, in 2020, and a B.Sc. degree in computer science from the Pontifical Catholic University of Minas Gerais, in 2017, both in Brazil. He is currently pursuing a Ph.D. degree in computer science from the University of Campinas, and his research interests include image processing, computer graphics, and machine learning.
\end{IEEEbiographynophoto}

\begin{IEEEbiographynophoto}{Alexandre Xavier Falcao}
Alexandre  Xavier  Falcao  is  full  professor at the Institute of Computing, University of Campinas, Campinas, SP, Brazil.  He received a B.Sc. in Electrical Engineering from the Federal University of Pernambuco, Recife, PE, Brazil, in 1988. He has worked in biomedical image processing, visualization and analysis since 1991. In 1993, he received a M.Sc.in Electrical Engineering from the University of Campinas, Campinas, SP,Brazil. During 1994-1996, he worked with the Medical Image Processing Group at the Department of Radiology, University of Pennsylvania, PA, USA,on interactive image segmentation for his doctorate. He got his doctorate in Electrical Engineering from the University of Campinas in 1996. In1997, he worked in a project for Globo TV at a research center, CPqD-TELEBRAS in Campinas, developing methods for video quality assessment.His experience as professor of Computer Science and Engineering started in 1998 at the University of Campinas. His main research interests include image/video processing, visualization, and analysis; graph algorithms and dynamic programming; image annotation, organization, and retrieval; machine learning and pattern recognition; and image analysis applications in Biology,Medicine, Biometrics, Geology, and Agriculture
\end{IEEEbiographynophoto}

\end{document}